\PassOptionsToPackage{table}{xcolor}
\documentclass[]{paper}

\RequirePackage{tocloft}

\usepackage[utf8]{inputenc}
\usepackage[T1]{fontenc}
\usepackage{hyperref}
\usepackage{url}
\usepackage{booktabs}
\usepackage{amsfonts}
\usepackage{nicefrac}
\usepackage{microtype}
\usepackage{caption}
\usepackage{multirow} 
\usepackage{makecell}
\usepackage{siunitx}

\usepackage{graphicx}
\usepackage{amssymb}

\usepackage{times}
\usepackage{latexsym}
\usepackage{inconsolata}
\usepackage{amsmath}
\hypersetup{colorlinks=true, linkcolor=purple} 
\usepackage{adjustbox}

\usepackage{fontawesome5}
\usepackage{array}
\usepackage{longtable}

\usepackage{enumitem}

\usepackage{color}
\usepackage{tikz}
\usepackage[edges]{forest}
\usepackage{xcolor}
\usepackage{wrapfig}
\usetikzlibrary{shapes.geometric, arrows.meta, shadows, positioning, decorations.pathmorphing}

\definecolor{agentblue}{RGB}{65, 105, 225}
\definecolor{toolorange}{RGB}{255, 140, 0}
\definecolor{signalgreen}{RGB}{46, 139, 87}
\definecolor{signalpurple}{RGB}{147, 112, 219}

\definecolor{majorelleblue}{rgb}{0.38, 0.31, 0.86}
\definecolor{boxcolor}{rgb}{0.38, 0.31, 0.86}
\definecolor{hidden-draw}{RGB}{205, 44, 36}
\definecolor{ao}{rgb}{0.0, 0.0, 1.0}
\definecolor{ferrarired}{rgb}{1.0, 0.11, 0.0}
\definecolor{bgColor}{RGB}{250,250,250}        
\definecolor{agentColor}{RGB}{255, 248, 227}    
\definecolor{toolColor}{RGB}{251, 255, 242}      
\definecolor{compareColor}{RGB}{244, 242, 255}   
\definecolor{appColor}{RGB}{255, 242, 242}  
\definecolor{evalColor}{rgb}{0.96, 1.0, 0.98}
\definecolor{oppColor}{RGB}{230, 246, 250}       
\definecolor{rootColor}{RGB}{252, 252, 252}      
\definecolor{darkblue}{RGB}{0, 51, 102}
\definecolor{cardinal}{rgb}{0.77, 0.12, 0.23}

\hypersetup{citecolor=ao}
\hypersetup{linkcolor=ferrarired}

\newcommand{\huggingface}{%
  \smash{%
    \raisebox{-0.3\height}{%
      \includegraphics[height=1.5em]{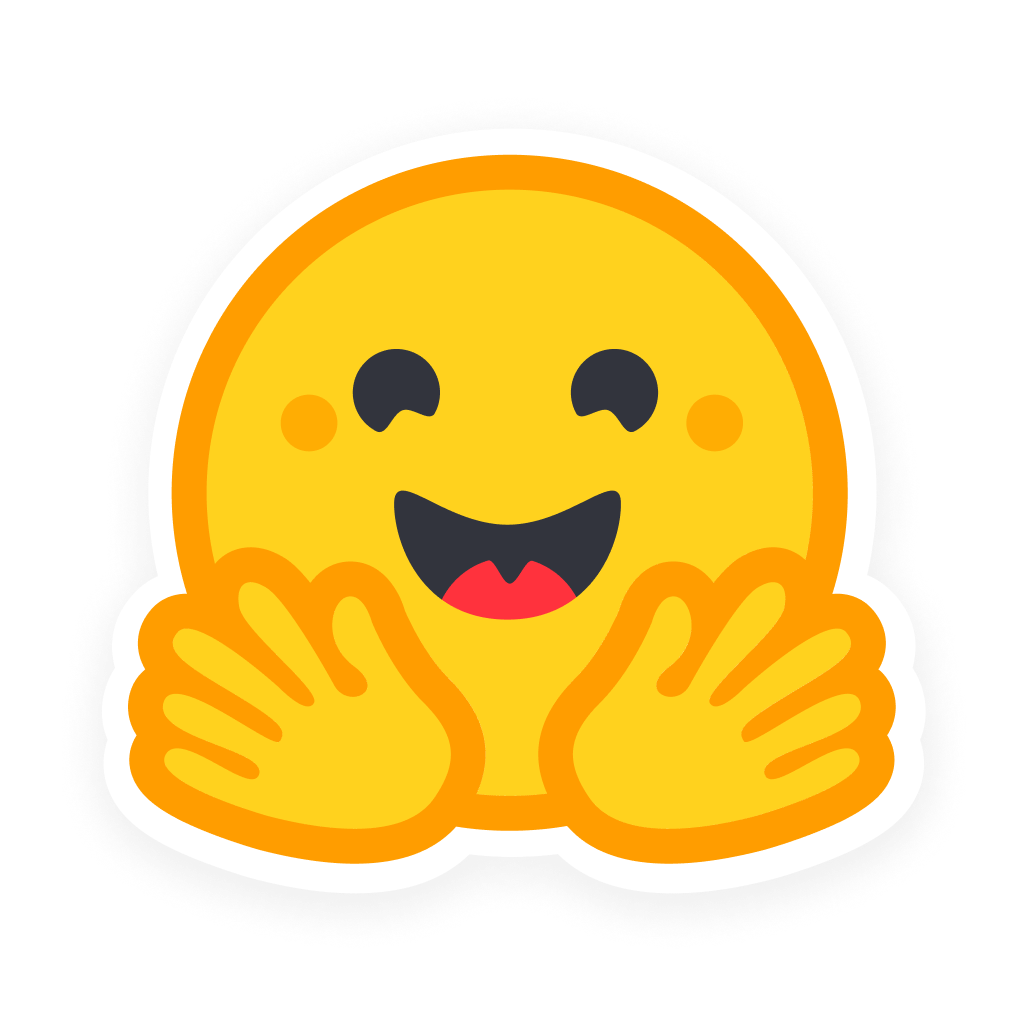}%
    }%
  }%
}

\NewDocumentCommand{\heng}
{ mO{} }{\textcolor{red}{\textsuperscript{\textit{Heng}}\textsf{\textbf{\small[#1]}}}}

\tikzstyle{my-box}=[
    rectangle,
    draw=gray!40,
    rounded corners,
    text opacity=1,
    minimum height=1.5em,
    minimum width=5em,
    inner sep=2pt,
    align=center,
]

\usepackage{subcaption}
\usepackage{listings}
\usepackage{csquotes}

\title{Adaptation of Agentic AI: A Survey of Post-Training, Memory, and Skills}

\author{Pengcheng Jiang$^{1*}$, Jiacheng Lin$^{1*}$, Zhiyi Shi$^{1,4*}$, Zifeng Wang$^{13}$, Luxi He$^{3}$, Yichen Wu$^{4}$, Ming Zhong$^{1}$,
Peiyang Song$^{5,6}$, Qizheng Zhang$^{2}$,
Heng Wang$^{1}$, Xueqiang Xu$^{1}$, Hanwen Xu$^{7}$, Pengrui Han$^{1}$, Dylan Zhang$^{1}$, Jiashuo Sun$^{1}$, Chaoqi Yang$^{1}$,
Kun Qian$^{14}$, Tian Wang$^{14}$, Changran Hu$^{5}$, 
Manling Li$^{10}$, Quanzheng Li$^{4,12}$, \\ Hao Peng$^{1}$, Sheng Wang$^{7}$, Jingbo Shang$^{8}$, Chao Zhang$^{9}$, Jiaxuan You$^{1}$, Liyuan Liu$^{1}$, Pan Lu$^{2}$, Yu Zhang$^{11}$, \\ Heng Ji$^{1}$, Yejin Choi$^{2}$, Dawn Song$^{5}$, Jimeng Sun$^{1, 13}$, Jiawei Han$^{1\dagger}$ \\
\vspace{2em}
$^1$\raisebox{-0.5ex}{\includegraphics[height=2.9ex]{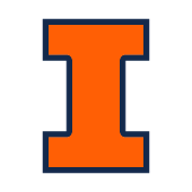}}
UIUC
\quad $^2$\raisebox{-0.5ex}{\includegraphics[height=2.8ex]{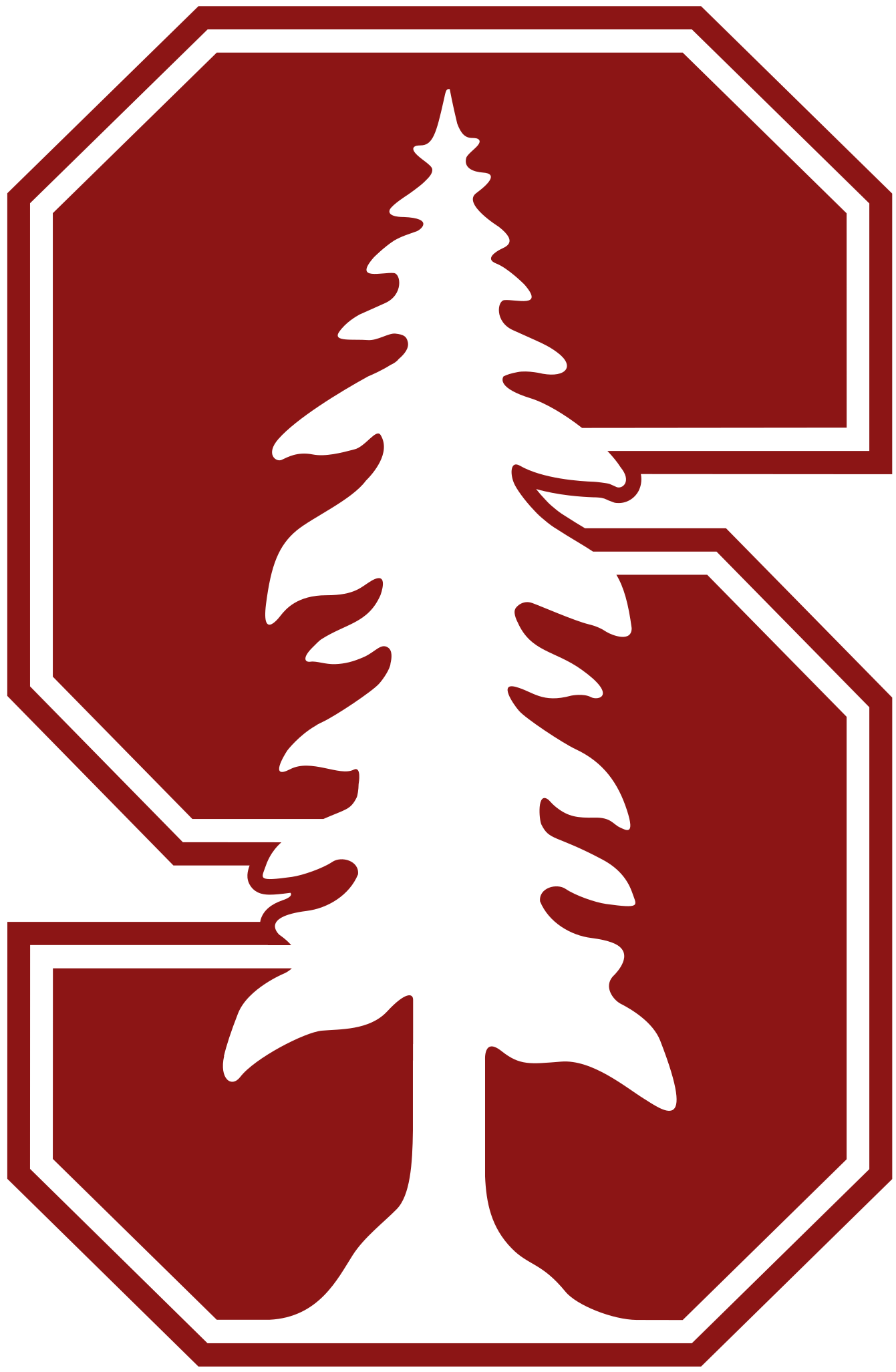}}
Stanford
\quad $^3$\raisebox{-0.5ex}{\includegraphics[height=2.8ex]{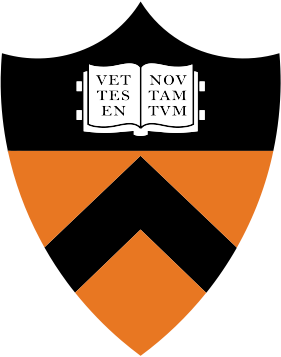}}
Princeton
\quad $^4$\raisebox{-0.3ex}{\includegraphics[height=2.4ex]{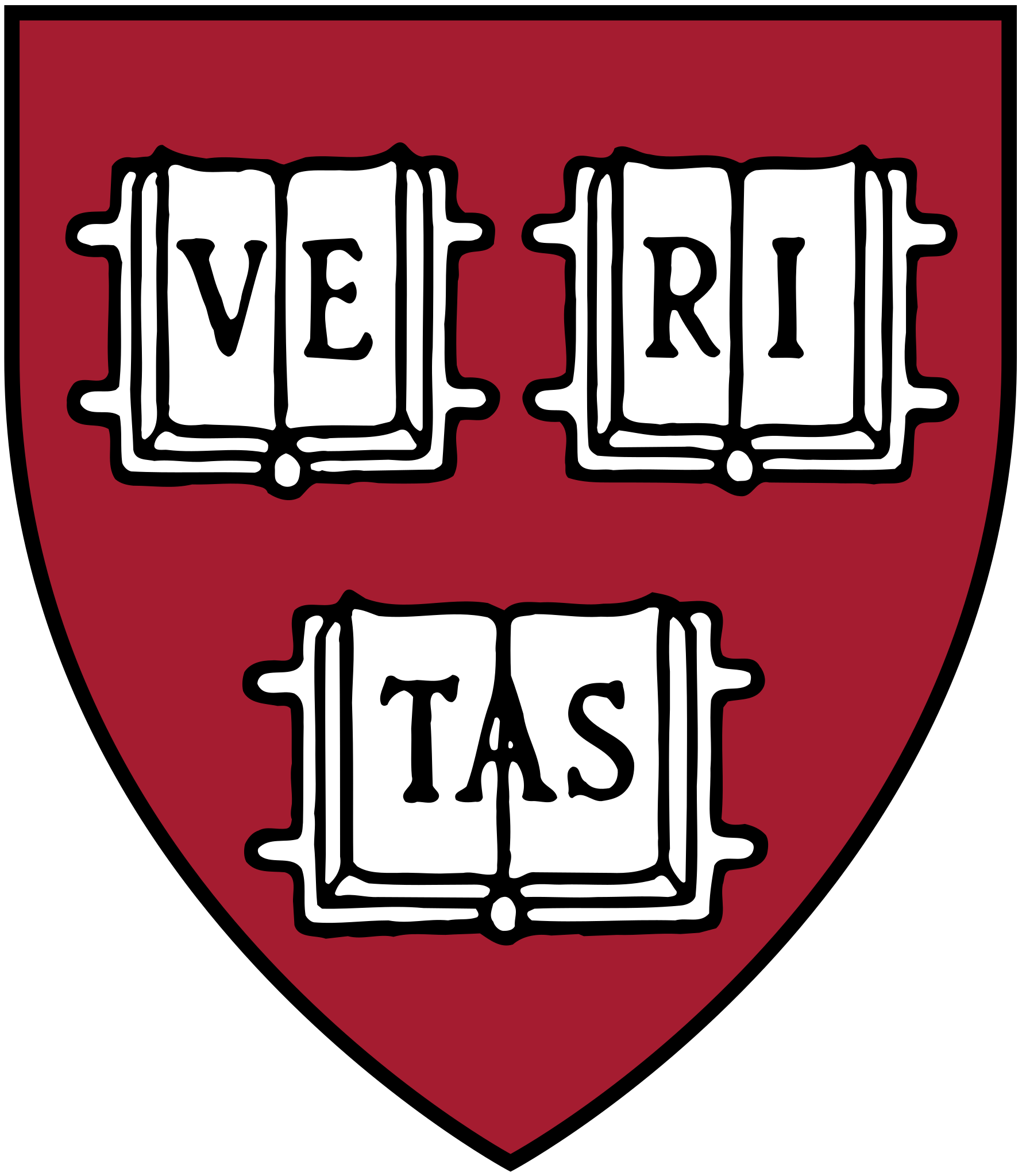}}
Harvard
\quad $^5$\raisebox{-0.3ex}{\includegraphics[height=2.7ex]{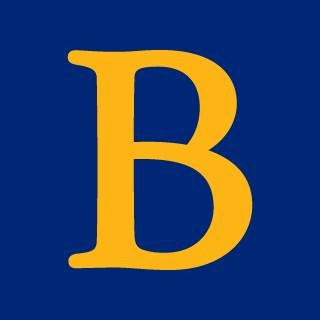}}
UC Berkeley
\quad $^6$\raisebox{-0.4ex}{\includegraphics[height=2.6ex]{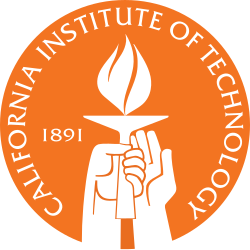}}
Caltech
\quad $^7$\raisebox{-0.2ex}{\includegraphics[height=2.2ex]{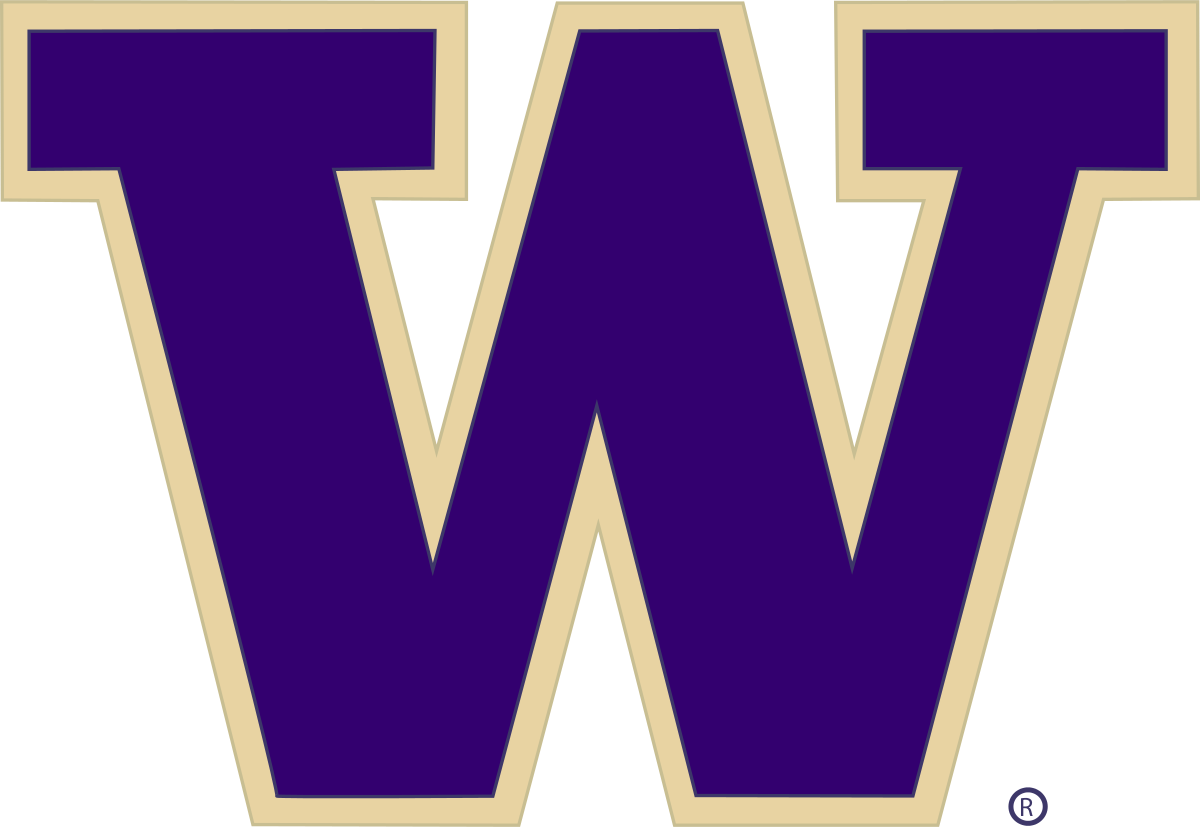}}
UW
\\
\vspace{1em}
\quad $^8$\raisebox{-0.3ex}{\includegraphics[height=2.5ex]{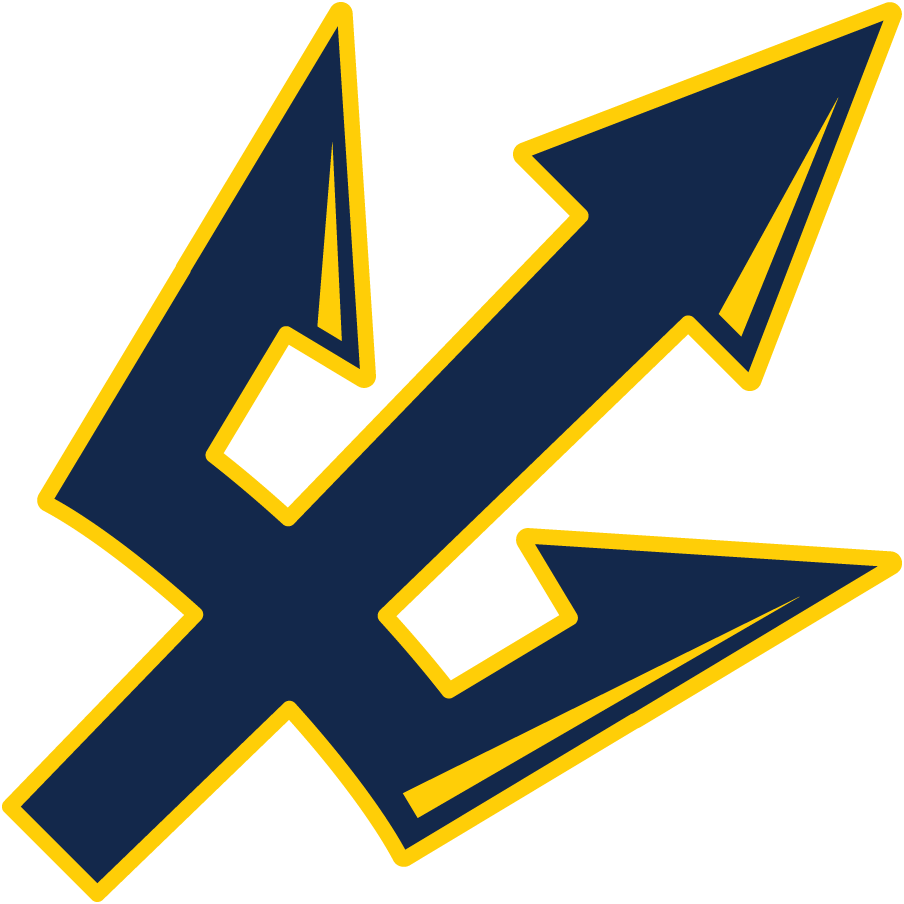}}
UCSD
\quad $^9$\raisebox{-0.3ex}{\includegraphics[height=2.4ex]{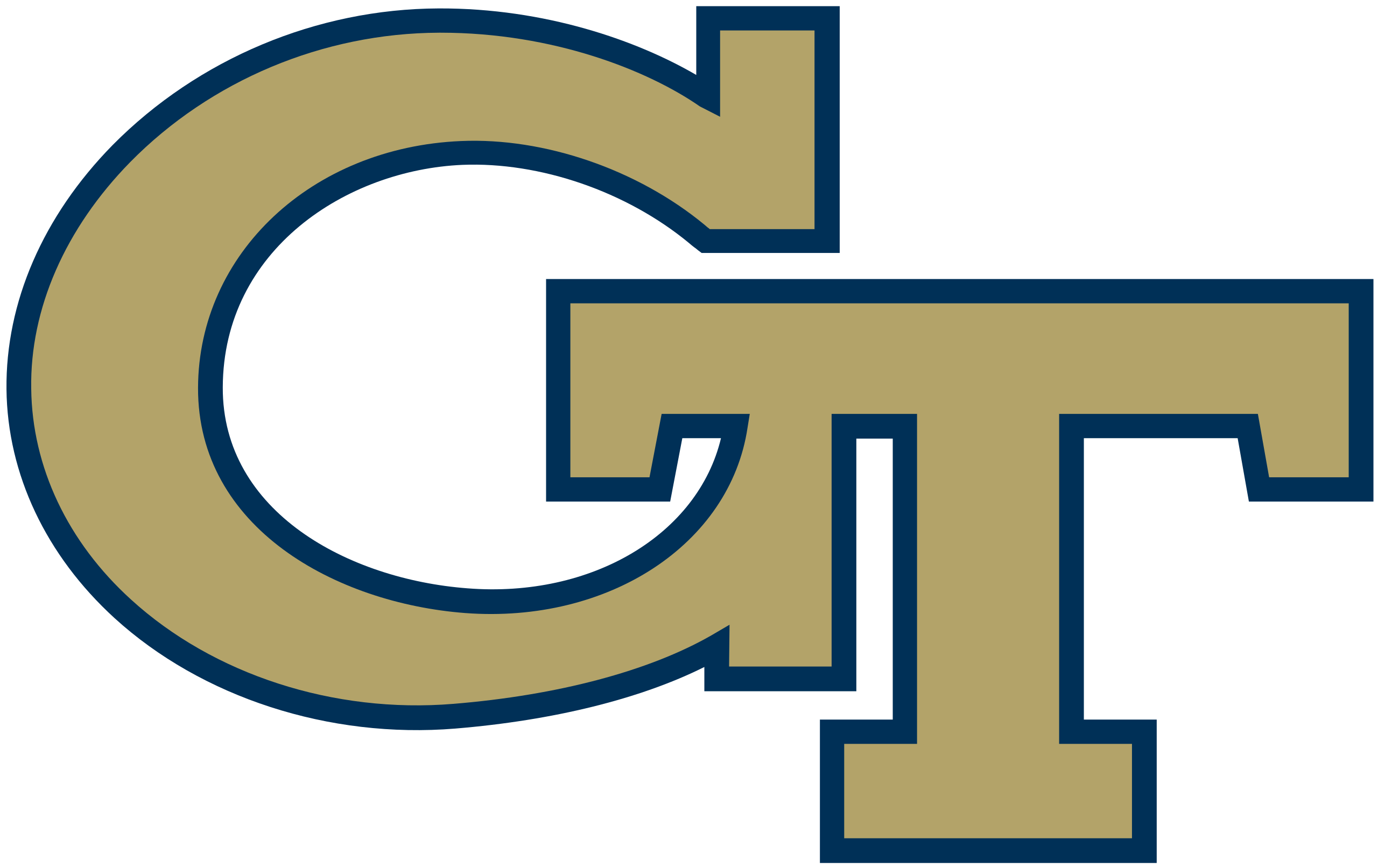}}
Georgia Tech
\quad $^{10}$\raisebox{-0.3ex}{\includegraphics[height=2.5ex]{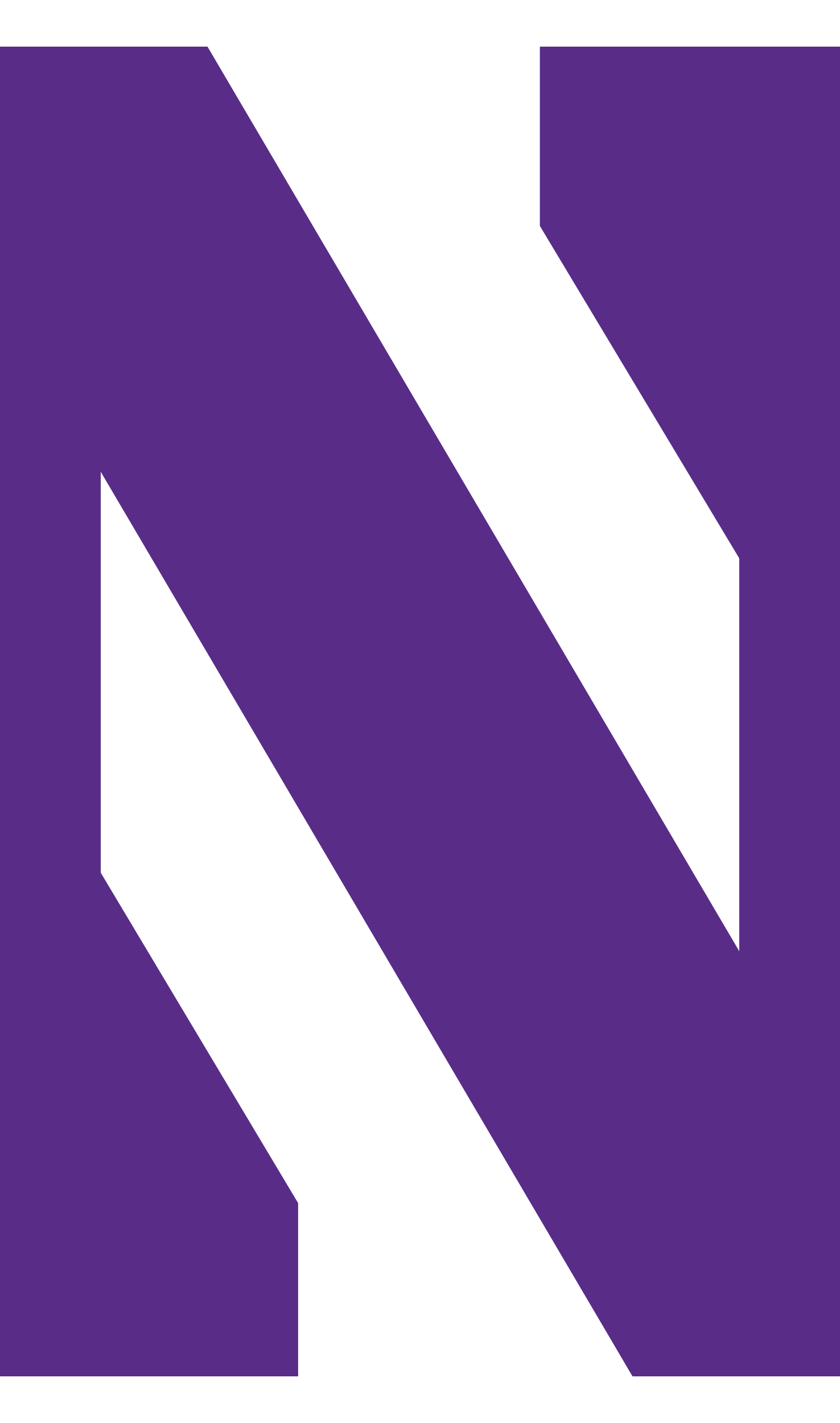}}
Northwestern
\quad $^{11}$\raisebox{-0.3ex}{\includegraphics[height=2.4ex]{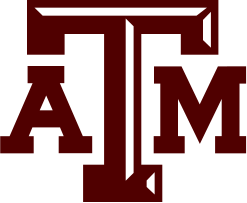}}
TAMU
\quad $^{12}$MGH
\quad $^{13}$Keiji AI
\quad $^{14}$Unity
} 

\abstract{
Large language model (LLM) agents are moving beyond prompting alone. ChatGPT marked the rise of general-purpose LLM assistants, DeepSeek showed that on-policy reinforcement learning with verifiable rewards can improve reasoning and tool use, and OpenClaw highlights a newer direction in which agents accumulate persistent memory and reusable skills. Yet the research landscape remains fragmented across post-training, retrieval, memory, and skill systems. This survey studies these developments under a single notion of \emph{adaptation}: improving an agent, its tools, or their interaction after pretraining. We organize the field with a four-paradigm framework spanning agent adaptation and tool adaptation. On the agent side, A1 (tool-execution-signaled) and A2 (agent-output-signaled) improve the agent itself through supervised fine-tuning, preference optimization, and reinforcement learning with verifiable rewards. On the tool side, T1 (agent-agnostic) provides reusable pre-trained modules any agent can call, while T2 (agent-supervised) uses the agent's outputs to train memory systems, skill libraries, or lightweight subagents. Using this framework, we review post-training methods, adaptive memory architectures, and agent skills; compare their trade-offs in cost, flexibility, and generalization; and summarize evaluation practices across deep research, software development, computer use, and drug discovery. We conclude by outlining open problems in agent-tool co-adaptation, continual learning, safety, and efficient deployment.
}

\repo{\href{https://github.com/pat-jj/Awesome-Adaptation-of-Agentic-AI}{https://github.com/pat-jj/Awesome-Adaptation-of-Agentic-AI}}

\begin{document}
\maketitle

\begin{figure*}[h]
    \centering
    \includegraphics[width=0.95\textwidth]{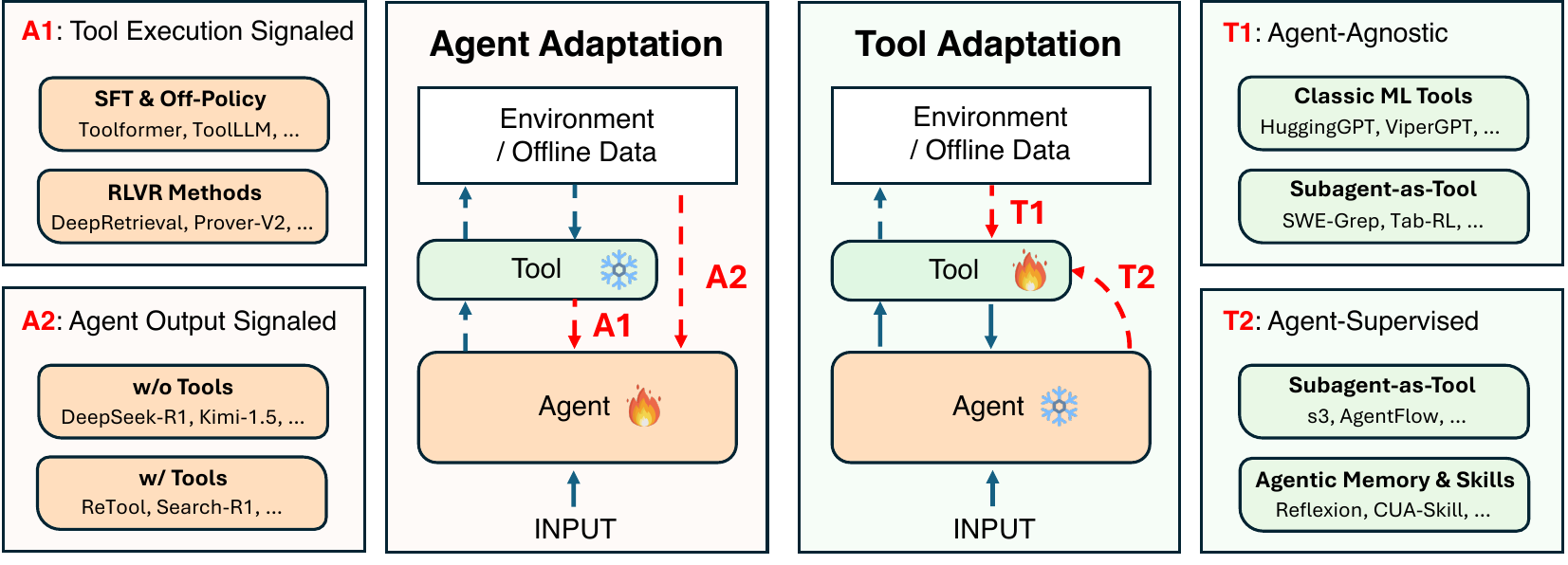}
    \caption{
    \textbf{Overview of adaptations in agentic AI.} \textit{Agent}: the foundation models serving as orchestration and reasoning modules; \textit{Tool}: callable components other than the agent model that operate independently, e.g., APIs, ML models, subagents, or memory. 
    We categorize these adaptations into two: agent adaptation (\textbf{A1} \& \textbf{A2}): adapting agent models, and tool adaptation (\textbf{T1} \& \textbf{T2}): adapting tools for agents. See more details in \S\ref{sec:overview}.
    }
    \label{fig:agent_and_tool}
\end{figure*}
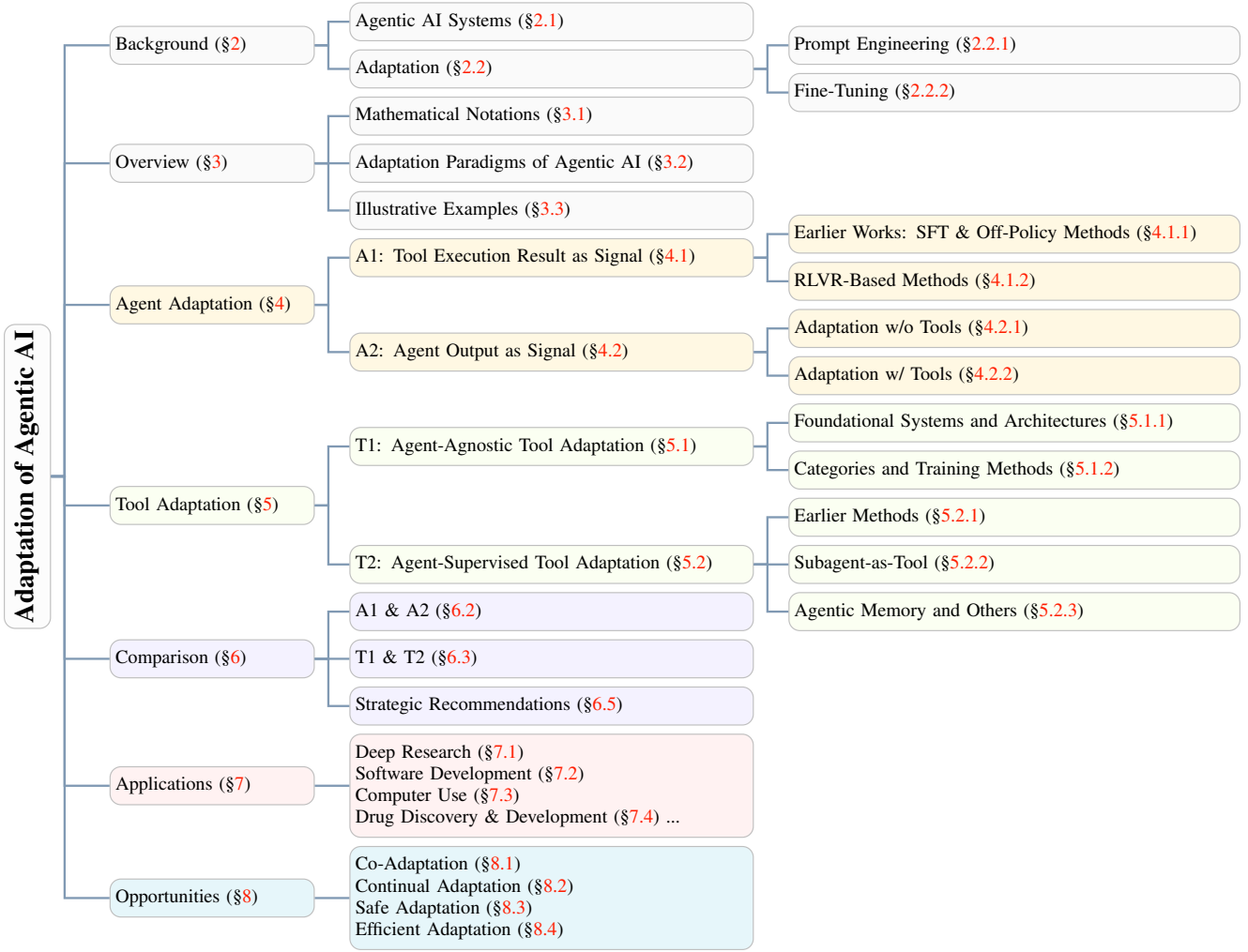
\begin{figure*}[htp]
    \centering
    \resizebox{\linewidth}{!}{
        \begin{forest}
            forked edges,
            for tree={
                grow=east,
                reversed=true,
                anchor=base west,
                parent anchor=east,
                child anchor=west,
                base=left,
                font=\small,
                rectangle,
                draw=gray!50,
                rounded corners,
                align=left,
                minimum width=5em,
                edge+={darkblue!50, line width=0.8pt},
                s sep=3pt,
                inner xsep=2pt,
                inner ysep=3pt,
                ver/.style={rotate=90, child anchor=north, parent anchor=south, anchor=center},
            },
            where level=1{text width=6.5em,font=\scriptsize,}{},
            where level=2{text width=13.2em,font=\scriptsize,}{},
            where level=3{text width=14.8em,font=\scriptsize,}{},
            [
                \textbf{Adaptation of Agentic AI}, ver, fill=rootColor
                [
                    Background (\S \ref{sec:background}), fill=bgColor
                    [
                        Agentic AI Systems (\S \ref{subsec:background_1_agentic_system}), fill=bgColor
                    ]
                    [
                        Adaptation (\S \ref{subsec:background_2_adaptation}), fill=bgColor
                        [
                            Prompt Engineering (\S \ref{subsubsec:background_2_1}), fill=bgColor
                        ]
                        [
                            Fine-Tuning (\S \ref{subsubsec:background_2_2}), fill=bgColor
                        ]
                    ]
                ]
                [
                    Overview (\S\ref{sec:overview}), fill=bgColor
                    [
                        Mathematical Notations (\S \ref{subsec:math_notations}), fill=bgColor
                    ]
                    [
                        Adaptation Paradigms of Agentic AI (\S \ref{subsec:formalization}), fill=bgColor
                    ]
                    [
                        Illustrative Examples (\S \ref{subsec:examples}),
                        fill=bgColor
                    ]
                ]
                [
                    Agent Adaptation (\S\ref{sec:agent_adaptation}), fill=agentColor
                    [
                        A1: Tool Execution Result as Signal (\S\ref{subsec:tool_execution_signal}), fill=agentColor
                        [
                            Earlier Works: SFT \& Off-Policy Methods (\S\ref{subsubsec:3.1.1}), fill=agentColor
                        ]
                        [
                            RLVR-Based Methods (\S\ref{subsubsec:3.1.2}), fill=agentColor
                        ]
                    ]
                    [
                        A2: Agent Output as Signal (\S\ref{subsec:agent_output_as_signal_for_agent}), fill=agentColor
                        [
                            Adaptation w/o Tools (\S \ref{subsec:agent_adaptation_no_tools}), fill=agentColor
                        ]
                        [
                            Adaptation w/ Tools (\S \ref{subsec:agent_adaptation_with_tools}), fill=agentColor
                        ]
                    ]
                ]
                [
                    Tool Adaptation (\S\ref{sec:tool_adaptation}), fill=toolColor
                    [
                        T1: Agent-Agnostic Tool Adaptation (\S\ref{subsec:agent_agnostic_tool_training}), fill=toolColor
                        [
                            Foundational Systems and Architectures (\S\ref{subsubsec:4.1.1}), fill=toolColor
                        ]
                        [
                            Categories and Training Methods (\S\ref{subsubsec:4.1.2}), fill=toolColor
                        ]
                    ]
                    [
                        T2: Agent-Supervised Tool Adaptation (\S\ref{subsec:agent_output_as_signal_for_tool}), fill=toolColor
                        [
                            Earlier Methods (\S\ref{subsubsec:4.2.1}), fill=toolColor
                        ]
                        [
                            Subagent-as-Tool (\S\ref{subsubsec:4.2.2}), fill=toolColor
                        ]
                        [
                            Agentic Memory and Skills (\S\ref{subsubsec:4.2.3}), fill=toolColor
                        ]
                    ]
                ]
                [
                    Comparison (\S \ref{sec:comparison}), fill=compareColor
                    [
                        A1 \& A2 (\S\ref{sec:comparison_agent}), fill=compareColor
                    ]
                    [
                        T1 \& T2 (\S\ref{sec:comparison_tool}), fill=compareColor
                    ]
                    [
                        Strategic Recommendations (\S\ref{sec:comparison_summary}), fill=compareColor
                    ]
                ]
                [
                    Evaluation (\S\ref{sec:eval}), fill=evalColor
                    [
                        Benchmark Landscape (\S\ref{subsec:benchmark_landscape}), fill=evalColor
                    ]
                    [
                        Adaptation Signal (\S\ref{subsec:eval_signal}), fill=evalColor
                    ]
                    [
                        Adaptation Dynamics (\S\ref{subsec:adapt_dynamics}), fill=evalColor
                    ]
                    [
                        Systemic Evaluation (\S\ref{subsec:systemic_eval}), fill=evalColor
                    ]
                    [
                        Discussion (\S\ref{subsec:eval_discussion}), fill=evalColor
                    ]
                ]
                [
                    Applications (\S\ref{sec:applications_and_existing_works}), fill=appColor
                    [
                        Deep Research (\S\ref{subsec:app_deep_research}) \\ Software Development (\S\ref{subsec:app_software}) \\ Computer Use (\S\ref{subsec:app_computer_use}) \\ Drug Discovery \& Development (\S\ref{subsec:app_drug}) ...,  fill=appColor
                    ]
                ]
                [
                    Opportunities (\S\ref{sec:opportunities}), fill=oppColor
                    [
                        Co-Adaptation (\S\ref{subsec:co-adapt}) \\
                        Continual Adaptation (\S\ref{subsec:continual_adapt}) \\
                        Safe Adaptation (\S\ref{subsec:safe_adapt}) \\
                        Efficient Adaptation (\S\ref{subsec:efficient_adapt}), fill=oppColor
                    ]
                ]
            ]
        \end{forest}
    }
    \caption{The structure of this paper.}
    \label{fig:taxo}
\end{figure*}

\section{Introduction}

The rapid progress of foundation models, large language models (LLMs) in particular, has enabled a new class of agentic AI systems that perceive their environment, invoke external tools, manage memory, and execute multi-step plans to complete complex tasks~\citep{fang2025comprehensive,luo2025large,xu2025llm,schick2023toolformer}. These systems show strong potential across scientific discovery~\cite{gao2024empowering,xu2025comprehensive}, software development, and clinical research~\cite{wang2025accelerating,wang2025perspective,gu2025challengespathsaisoftware}.
Yet current agentic systems remain limited by unreliable tool use, shallow long-horizon planning, domain-specific reasoning gaps, and poor generalization to environments where the agent lacks prior interaction experience~\citep{qin2024toolllm,shi2024replug,jin2024matching,song2025survey}. Even the most capable foundation models require additional \textit{adaptation} to specialize for particular tasks or deployment scenarios.

Three mechanisms drive this adaptation process. First, \textbf{post-training} (supervised fine-tuning, reinforcement learning with verifiable rewards, preference optimization) modifies the agent's parameters to improve reasoning and tool use. Second, \textbf{memory} systems (episodic buffers, reflective databases, structured knowledge graphs) allow agents to retain and build upon past experience without retraining the core model. Third, \textbf{skills}---reusable units of procedural knowledge encoding \textit{how} to perform tasks~\cite{wu2025agentskills}---accumulate through diverse mechanisms: agents internalize tool-use procedures via post-training, while external skill libraries support discovery, invocation, and refinement across sessions. These three mechanisms interact within a shared design space that this survey maps.

Unlike existing surveys on AI agents~\citep{fang2025comprehensive,gao2025survey,plaat2025agentic,tao2024survey,wang2024survey,belcak2025small}, we focus specifically on adaptation. We introduce a unified framework that organizes agentic adaptation into four paradigms spanning both agent and tool adaptation (Figure~\ref{fig:agent_and_tool}), exposing trade-offs that guide paradigm selection based on supervision signals, task requirements, and system-level constraints.

Our framework organizes adaptation along two dimensions according to which component is optimized (\S \ref{sec:overview}). \textbf{Agent Adaptation} modifies the agent's internal parameters, representations, or behavioral policies to meet task requirements, through fine-tuning~\cite{hu2022lora} or reinforcement learning with environment feedback~\cite{gehring2025rlefgroundingcodellms,jiang2025deepretrieval}. \textbf{Tool Adaptation} shifts the optimization target to the agent's external tools (retrievers, planners, memory modules, specialized models), enabling frozen agents to benefit from an adaptive operational environment~\cite{zhou2025memento,shi2024replug,wang2024learning}. Within these two dimensions, we identify four distinct adaptation strategies:

\begin{itemize}
    \item \textbf{A1: Tool Execution Signaled Agent Adaptation} (\S \ref{subsub:a1_math}, \S \ref{subsec:tool_execution_signal}):
    The agent is optimized using verifiable outcomes produced by external tools it invokes.
    This paradigm captures settings where correctness signals arise directly from tool execution, such as code sandbox results, retrieval relevance scores, or API call outcomes.
    
    \item \textbf{A2: Agent Output Signaled Agent Adaptation} (\S\ref{subsub:a2_math}, \S\ref{subsec:agent_output_as_signal_for_agent}): 
    The agent is optimized using evaluations of its own outputs, e.g., final answers, plans, or reasoning traces, possibly after incorporating tool results.  
    This paradigm includes both tool-free outcome-based learning and tool-augmented adaptation driven by answer correctness or preference scores.

    \item \textbf{T1: Agent-Agnostic Tool Adaptation} (\S\ref{subsub:t1_math}, \S\ref{subsec:agent_agnostic_tool_training}): 
    Tools are trained independently of the frozen agent.  
    These tools include retrievers, domain-specific models, and other pre-trained components that can be used as plug-and-play modules orchestrated by the frozen agent.

    \item \textbf{T2: Agent-Supervised Tool Adaptation} (\S\ref{subsub:t2_math}, \S\ref{subsec:agent_output_as_signal_for_tool}):
    The agent remains fixed while its tools are adapted using signals derived from the agent’s outputs.  
    This paradigm includes reward-driven retriever tuning, adaptive rerankers, search subagents, and memory-update modules trained to better support the frozen agent.

\end{itemize}

\noindent
\textbf{These four strategies overlap in practice; we treat them as a working classification.} When a method simultaneously modifies both the agent and a tool, we assign it to the paradigm corresponding to its \emph{dominant locus of optimization} and note the secondary component explicitly. Current leading systems combine multiple paradigms. A deep research system may employ T1-style retrieval tools (pre-trained dense retrievers), T2-style adaptive search agents (trained via frozen LLM feedback), and A1-style reasoning agents (fine-tuned with execution feedback) in a cascaded architecture~\cite{guo2025deepseek,zhang2025nemotron,mei2025ai,xu2025comprehensive}.

The choice among these paradigms involves trade-offs along several dimensions (\S \ref{sec:comparison}). (1) \textbf{Cost and flexibility}: Agent adaptation (A1/A2) requires substantial computational resources for training billion-parameter models but offers broad flexibility, while tool adaptation (T1/T2) optimizes external components at lower cost but is constrained by the frozen agent's capabilities~\cite{jiang2025s3,zhou2025memento}. (2) \textbf{Generalization}: T1 tools trained on broad data distributions generalize well across agents and tasks~\cite{wang2024learning,yu2023augmentation}, whereas A1 methods may overfit to specific environments unless carefully regularized~\cite{gehring2025rlefgroundingcodellms}. (3) \textbf{Modularity}: T2 approaches enable independent tool upgrades without agent retraining~\cite{hao2023toolkengpt,zhou2025memento}, while A1/A2 methods risk catastrophic forgetting when adapted to new tasks.

\begin{figure*}[t]
\vspace{1em}
    \centering
    \includegraphics[width=\textwidth]{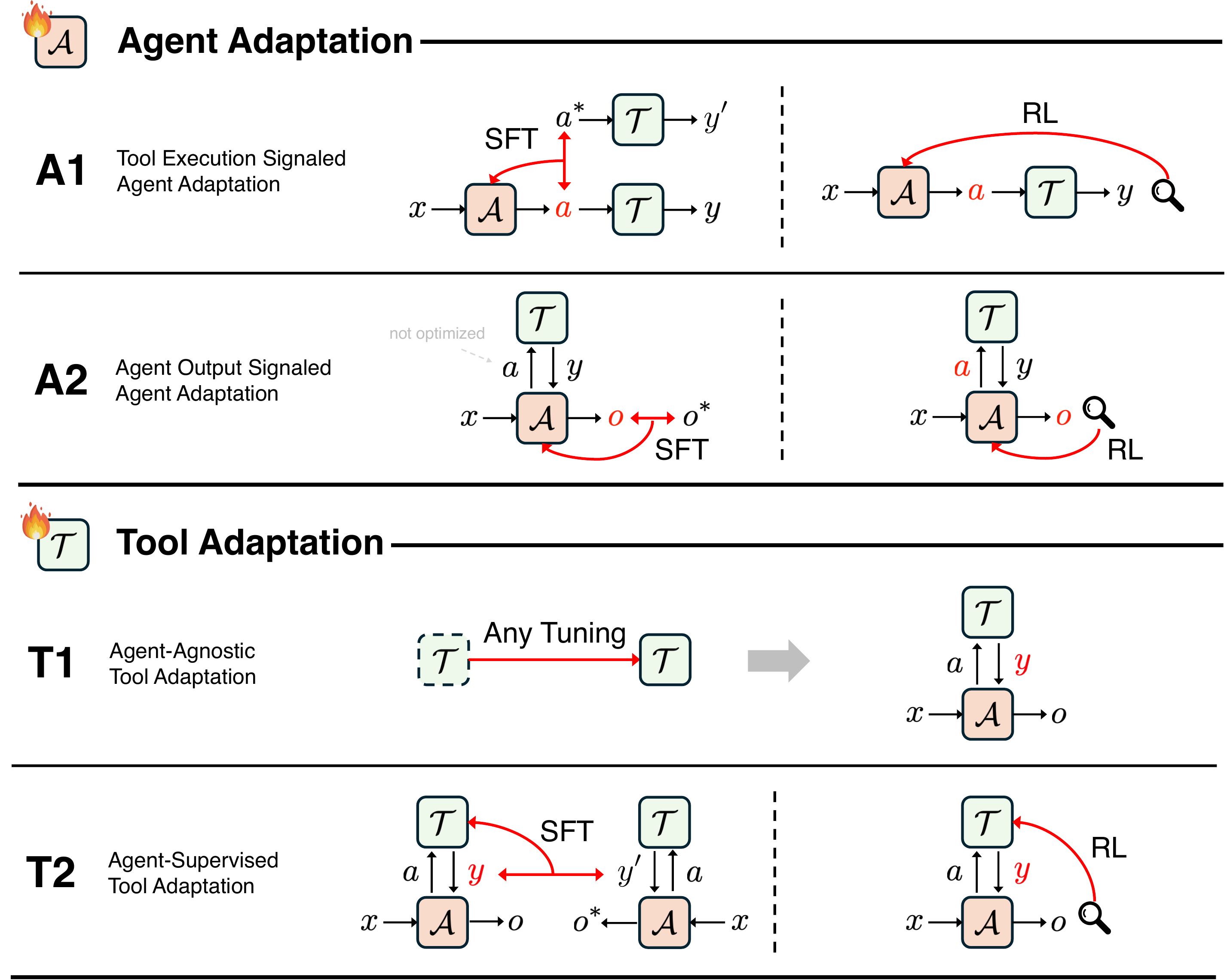}
    \caption{
    \textbf{Illustration of Four Adaptation Paradigms (A1, A2, T1, and T2)}. In all the panels, letters highlighted in \textcolor{red}{Red} denote the components \underline{directly} being optimized during adaptation. The red arrows show the sources of adaptation signals. The dotted black lines separate the cases of supervised fine-tuning (SFT) and reinforcement learning (RL).
    }
    \label{fig:a1a2_sft_rl}
\end{figure*}

\textbf{Scope and contributions.} By grounding the four-paradigm taxonomy in three concrete mechanisms---post-training, memory, and skills---we connect the abstract framework to the techniques that practitioners deploy in practice. Our key contributions are:

\begin{itemize}
    \item A \textbf{unified $2\times 2$ framework} (Figures~\ref{fig:agent_and_tool},~\ref{fig:a1a2_sft_rl},~\ref{fig:adaptation_landscape}) that decomposes agentic adaptation along two axes---\emph{what is adapted} (agent vs.\ tool) and \emph{how the signal is obtained} (execution-grounded vs.\ output-evaluated)---yielding four paradigms (A1/A2/T1/T2) with distinct cost, flexibility, and generalization profiles, together with \textbf{cross-paradigm empirical synthesis} (\S\ref{sec:comparison}) showing, e.g., that T2 tool adaptation can match A2 agent training accuracy with far fewer examples in retrieval settings, and that A1-trained agents can be frozen and redeployed as T1 tools.
    \item A \textbf{systematic treatment of agentic memory and skills} as adaptation mechanisms. We show how adaptive memory systems (episodic buffers, reflective databases, knowledge graphs) instantiate T2 adaptation, and how skill libraries bridge A1/A2 post-training with T1/T2 tool ecosystems (\S\ref{sec:tool_adaptation}).
    \item A \textbf{multi-dimensional evaluation framework} (\S\ref{sec:eval}) that maps benchmarks to paradigms, distinguishes verifiable execution metrics from holistic utility metrics, and identifies gaps in current evaluation practice, particularly for T2 and cross-paradigm comparison.
    \item \textbf{Domain-specific paradigm mapping} across deep research, software development, computer use, and drug discovery (\S\ref{sec:applications_and_existing_works}), showing how the dominant adaptation paradigm varies with the availability of verifiable feedback and the cost of agent retraining.
    \item Identification of \textbf{open challenges} including agent-tool co-adaptation, continual adaptation under non-stationary task distributions, safe exploration during on-policy RL, and efficient adaptation for resource-constrained deployment (\S\ref{sec:opportunities}).
\end{itemize}

\textbf{Survey methodology and scope.} We surveyed the literature on agentic adaptation published through early 2026, drawing primarily from top-tier venues (NeurIPS, ICML, ICLR, ACL, EMNLP, NAACL) and high-impact arXiv preprints. Methods are assigned to paradigms based on their \emph{dominant} locus of optimization and signal source; boundary cases are flagged explicitly. Quantitative cross-paradigm comparisons (e.g., data efficiency of T2 vs.\ A2) are drawn from case studies that may differ in confounding factors; we note these limitations where they arise.

\textbf{Organization.} Section~\ref{sec:background} introduces foundational concepts. Section~\ref{sec:overview} formalizes the four paradigms with illustrative examples. Sections~\ref{sec:agent_adaptation} and~\ref{sec:tool_adaptation} review agent adaptation (A1, A2) and tool adaptation (T1, T2) methods, respectively. Section~\ref{sec:comparison} compares paradigms along key dimensions. Section~\ref{sec:eval} presents a multi-dimensional evaluation framework. Section~\ref{sec:applications_and_existing_works} examines real-world applications with explicit paradigm mapping. Section~\ref{sec:opportunities} discusses open challenges, and Section~\ref{sec:conclusion} concludes the paper.
\section{Background}
\label{sec:background}
This section introduces the background necessary for the remainder of the survey. We first describe the core components of \textit{Agentic AI Systems} (\S\ref{subsec:background_1_agentic_system}), then discuss the two primary forms of \textit{Adaptation} (\S\ref{subsec:background_2_adaptation}) through which these systems specialize for particular tasks or deployment scenarios.

\subsection{Agentic AI Systems}
\label{subsec:background_1_agentic_system}

An agentic AI system combines a foundation model with planning, tool use, and memory modules to autonomously execute multi-step tasks, using environmental feedback to refine its behavior over time. 
This survey focuses on \textit{single-agent systems}, which provide a controlled yet expressive setting to study how an individual agent perceives, plans, and acts. 
Understanding single-agent adaptation is a prerequisite for studying \textit{multi-agent systems}, in which multiple agents coordinate, cooperate, or compete; see~\citep{fang2025comprehensive, luo2025large, liu2025advances} for overviews.

At the core of an agentic AI system lies a \textbf{foundation model}, typically a large language model (LLM) or multimodal model that serves as the agent's reasoning and control center. 
Several additional components extend the agent's autonomy:

\begin{itemize}[leftmargin=15pt]
    \item \textbf{Planning Module:} Decomposes complex goals into actionable steps and organizes their sequential or hierarchical execution. 
    Depending on the degree of feedback integration, planning takes two forms. 
    \textit{Static planning} methods, such as Chain-of-Thought~\citep{wei2022chain} and Tree-of-Thought~\citep{yao2023tree}, enable structured reasoning through single-path or multi-path task decomposition. 
    In contrast, \textit{dynamic planning} approaches, such as ReAct~\citep{yao2022react} and Reflexion~\citep{shinn2023reflexion}, incorporate feedback from the environment or past actions, allowing the agent to iteratively refine its plans and improve performance in long-horizon or partially observable scenarios.

    \item \textbf{Tool Use:} Enables the agent to interact with external resources and computational systems, extending its capabilities beyond the limitations of its internal knowledge. 
    Typical tools include web search engines, APIs, code execution environments, Model Context Protocols (MCPs), and browser automation frameworks~\citep{toollearning2024,xu2025llm}. 
    The tool-selection policy itself is an adaptation surface: agents must learn which tool to invoke for a given subtask and how to parse the tool's output for downstream reasoning.
    
    \item \textbf{Memory Module:} Allows the agent to retain, retrieve, and utilize past information for context-aware reasoning and long-term consistency. 
    Memory is typically divided into \textit{short-term memory}, which stores contextual information generated during the current task, and \textit{long-term memory}, which persists across sessions to accumulate reusable knowledge and experience~\citep{fang2025comprehensive, wu2025human}. 
    To access relevant information from long-term memory, many systems employ retrieval-augmented generation (RAG) mechanisms that retrieve and integrate stored knowledge into the agent’s reasoning process. 
    Designing an effective memory module involves challenges such as how to structure stored information, when and what to retain, how to retrieve relevant knowledge efficiently, and how to integrate it into ongoing reasoning and decision-making.\footnote{While memory is a fundamental component of an agent, this survey treats most \textit{adaptive memory systems} under the Tool Adaptation paradigm (discussed in \S\ref{subsubsec:4.2.3}). External, non-parametric memory stores (e.g., episodic buffers, reflective databases, skill libraries) that are updated using the frozen agent's outputs are classified as T2; pre-trained memory modules that operate independently of a specific agent fall under T1; and parametric or hybrid memory that modifies model weights occupies the boundary between tool and agent adaptation. A detailed discussion of these distinctions appears in \S\ref{sec:overview}.}
\end{itemize}

\subsection{Adaptation}
\label{subsec:background_2_adaptation}
Adaptation modifies the agent or its tools over time to improve performance on specific domains, tasks, or deployment conditions. 
Without adaptation, even capable foundation models struggle to generalize beyond their pre-training distribution. We distinguish two broad categories: prompt-based adaptation (\S \ref{subsubsec:background_2_1}) and fine-tuning-based adaptation (\S \ref{subsubsec:background_2_2}).

\subsubsection{Prompt Engineering}
\label{subsubsec:background_2_1}
Prompt engineering is a lightweight form of adaptation that guides the behavior of an agentic system without modifying model parameters. 
The agent's behavior is shaped by carefully crafted input prompts that define goals, constraints, and contextual instructions.

A \textit{prompt} refers to the input context provided to the agent’s core model, typically consisting of instructions, examples, or task descriptions that specify the desired behavior. By modifying or composing prompts, an agent can be adapted to new goals or environments without any additional model training, making this approach efficient and transferable across tasks. Such prompt-based adaptation has been widely adopted in recent agentic systems, such as CAMEL \citep{li2023camel}, AutoGen \citep{wu2024autogen}, MetaGPT \citep{hong2023metagpt} and ChatDev \citep{qian2024chatdev}. For a comprehensive overview of prompt engineering techniques and their design principles, we refer readers to the survey by~\citet{sahoo2024systematic}.

\subsubsection{Fine-Tuning}
\label{subsubsec:background_2_2}
In contrast, \textit{fine-tuning} adapts the agent by updating the internal parameters of the core model. 
Training on task-specific data shifts the model's distribution toward domain-appropriate behavior.

Fine-tuning can be performed at different granularities depending on data availability, computational cost, and the desired degree of adaptation. 
\textit{Full fine-tuning} updates all model parameters using labeled data, providing maximal flexibility but often requiring substantial resources. 
Alternatively, \textit{parameter-efficient fine-tuning} (PEFT) methods, such as low-rank adaptation (LoRA) \citep{hu2022lora}, update only a small subset of parameters. 
LoRA, for example, typically matches full fine-tuning quality while updating less than 1\% of parameters, making it practical for large agentic systems. For a comprehensive overview of PEFT methods, we refer readers to the survey by~\citet{han2024parameterefficient}.

Fine-tuning for adapting agents encompasses several major training paradigms. Supervised Fine-Tuning (SFT) \citep{wei2022finetuned} performs imitation learning on curated demonstrations. Preference-based methods, such as Direct Preference Optimization (DPO) \citep{rafailov2023direct} and its extensions \citep{xiao2024comprehensive}, align the model with human or automated preference signals. Reinforcement learning methods such as Proximal Policy Optimization (PPO) \citep{schulman2017proximal} and Group Relative Policy Optimization (GRPO) \citep{shao2024deepseekmath} optimize behavior through environment interaction; they require only a reward signal but are harder to stabilize than SFT. For a more comprehensive review of these approaches, see the survey by~\citet{zhang2025survey}.

\section{Overview of Adaptation Paradigms of Agentic AI}
\label{sec:overview}

We present the adaptation paradigms that form the analytical basis of this paper. 
We categorize existing studies on agentic AI systems according to \textbf{what is adapted} (the agent or the tool) and \textbf{how the adaptation signal is obtained}, yielding four canonical paradigms that capture the major directions of adaptation in recent literature.

\subsection{Mathematical Notations}
\label{subsec:math_notations}

We introduce the key mathematical notations used throughout this paper, organized into three categories: \textit{adaptation targets}, \textit{adaptation data sources}, and \textit{adaptation objectives}.

\paragraph{Adaptation Targets.}
The entities that undergo adaptation within an agentic AI system are as follows.
\begin{itemize}[leftmargin=15pt]
    \item \textbf{Agent} ($\mathcal{A}$): The foundation model that serves as the core reasoning and decision-making component of the system, parameterized by $\theta$. 
    Adaptation of the agent can occur through \textit{parameter updates}, \textit{prompt refinement}, or other modifications to its internal policy.
    \item \textbf{Tool} ($\mathcal{T}$): The set of external callable components that extend the agent’s capabilities beyond its internal parameters. 
    Tools can include retrievers, planners, executors, simulators, or other computational modules. 
    In this paper, we \textbf{treat external memory modules as components of $\mathcal{T}$} when they function as dynamic, updatable stores that interact with and learn from the agent's outputs. 
    The retrieval process for accessing stored information is typically performed through a dedicated retriever or search tool, which allows the agent to query and integrate relevant past knowledge into its reasoning process. We note that parametric and hybrid memory forms may blur the boundary between $\mathcal{T}$ and $\mathcal{A}$; see the discussion below (\S\ref{subsec:formalization}).
\end{itemize}

\paragraph{Adaptation Data Sources.}
The sources from which adaptation signals are obtained include the following.
\begin{itemize}[leftmargin=15pt]
    \item \textbf{Offline Data} ($\mathcal{D}$): Offline data that serve as alignment references or supervision sources for improving either the agent or the tool. 
    These data may include human-labeled demonstrations, synthetic trajectories, or logs of prior interactions.
    \item \textbf{Environment} ($\mathcal{E}$): The external environment in which the agent or tool interacts and receives feedback. 
    It provides online experience signals that reflect task performance or execution quality.
\end{itemize}

\paragraph{Adaptation Objectives.}
The objective that guides the adaptation process quantifies performance or alignment quality.
\begin{itemize}[leftmargin=15pt]
    \item \textbf{Objective Function} $\mathcal{O}(\cdot)$: The objective function optimized during adaptation, which evaluates how effectively the agent-tool system performs according to the designated evaluation protocol. 
    For example, the objective for offline data $\mathcal{D}$ may correspond to supervised or imitation learning losses such as supervised fine-tuning (SFT) or behavior cloning. 
    When adaptation relies on interactions with the environment $\mathcal{E}$, the objective is typically defined by outcome-based metrics such as task success rate.
\end{itemize}

\subsection{Four Adaptation Paradigms of Agentic AI}
\label{subsec:formalization}
Building on these notations, we present the four adaptation paradigms proposed in this paper. 
Adaptation is first categorized by the optimization target: the \textit{agent} or the \textit{tool}. 
For \textit{agent adaptation}, paradigms are further differentiated by the optimization signal: tool-execution feedback (A1) or evaluations of the agent's final output (A2). 
For \textit{tool adaptation}, the distinction concerns whether tools are optimized independently of any agent (T1) or adapted under the supervision of a fixed agent (T2). 

\subsubsection{A1: Tool Execution Signaled Agent Adaptation}
\label{subsub:a1_math}
In the A1 paradigm, the agent $\mathcal{A}$ is improved through feedback signals derived from the execution results of external tools $\mathcal{T}$, capturing scenarios where tool outcomes serve as a measurable basis for optimization.

\paragraph{Agent-Tool Interaction Process.}
The agent receives an input $x$ (e.g., a user query or task description) and generates a structured tool call or action $a=\mathcal{A}(x)$, which may include the tool name, arguments, and calling context. 
The tool set $\mathcal{T}$ then executes this call to produce a result $y=\mathcal{T}(a)$. 
The pair $(a, y)$ represents a single agent-tool interaction, and the overall process can be summarized as
\[
x \xrightarrow{\mathcal{A}} a \xrightarrow{\mathcal{T}} y.
\]
The pipeline captures how the agent uses tools to complete tasks. For simplicity and without loss of generality, we describe the interaction using a single tool invocation; multi-turn tool use follows as a direct extension of the formulation.

\paragraph{Optimization Objective.}
Given this interaction process, the general optimization goal is to adjust the agent $\mathcal{A}$ to generate high-quality tool call action $a$ such that the tool-executed outcomes achieve better performance. 
Formally,
\begin{align}
\text{\textbf{(A1)}} \quad
\mathcal{A}^* = \arg\max_{\mathcal{A}} \mathcal{O}_{\text{tool}}(\mathcal{A}, \mathcal{T}),
\end{align}
where $\mathcal{O}_{\text{tool}}$ measures the quality or correctness of the outputs obtained from invoking $\mathcal{T}$, such as tool execution success rate or retrieval scores. This optimization can be instantiated in two primary forms: (1) by imitating collected successful tool-call trajectories, or 
(2) by generating actions interactively and using the resulting tool feedback to optimize $\mathcal{A}$ via Reinforcement Learning. 
\begin{itemize}[leftmargin=15pt]
    \item \textbf{Supervised Fine-Tuning (SFT).} 
    When explicit target actions are available, the agent learns to imitate successful tool-using behaviors from recorded trajectories without performing online interaction. 
    Let $\mathcal{D}_{\text{succ}}=\{(x,a^*)\}$ denote a dataset of input $x$ and reference action $a^*$ that is known to lead to a correct or desirable tool outcome ($y'$). 
    The supervised objective is formulated as:
    \begin{align}
    \mathcal{A}^*
    = \arg\min_{\mathcal{A}}\; \mathbb{E}_{(x,a^*)\sim \mathcal{D}_{\text{succ}}}\big[\ell(\mathcal{A}(x), a^*)\big]
    \;\equiv\;
    \arg\max_{\mathcal{A}}\; \mathbb{E}_{(x,a^*)}\big[\log p_{\mathcal{A}}(a^*|x)\big],
    \end{align}
    where $\ell$ denotes the cross-entropy loss used for next-token prediction in language models.

    \item \textbf{Reinforcement Learning (RL).} 
    Alternatively, the agent can acquire adaptation signals through interactions with the environment, where it executes tool calls and receives evaluative feedback from the resulting outcomes. 
    The process follows:
    \[
    x \xrightarrow{\mathcal{A}} a \xrightarrow{\mathcal{T}} y,
    \quad\text{with reward}\quad R = \mathcal{O}_{\text{tool}}(y).
    \]
    Here, the agent $\mathcal{A}$ generates an action or tool call $a$ based on input $x$, the tool $\mathcal{T}$ executes $a$ to produce a result $y$, and the evaluation function $\mathcal{O}_{\text{tool}}$ assigns a scalar feedback $R$ indicating task success or quality. 
    The optimization objective can be expressed as:
    \begin{align}
    J(\mathcal{A})
    &= \mathbb{E}_{x\sim\mathcal{D}_0,\; a\sim\mathcal{A}(\cdot|x),\; y=\mathcal{T}(a)}[\mathcal{O}_{\text{tool}}(y)],
    \end{align}
    where $\mathcal{D}_0$ denotes the input distribution.
\end{itemize}

\subsubsection{A2: Agent Output Signaled Agent Adaptation}
\label{subsub:a2_math}
Unlike the A1 paradigm, where the adaptation signal is derived from tool-execution outcomes, the A2 paradigm obtains its optimization signal from the agent’s own final output. 
For simplicity, the following description focuses on a single-turn interaction; the multi-turn case extends naturally.

\paragraph{Agent-Tool Interaction Process.}
In the A2 paradigm, the agent first generates a tool call $a$ from the input $x$, the tool $\mathcal{T}$ executes this call and returns an executed result $y$, and the agent then integrates $x$ and $y$ to produce the final output $o$:
\[
x \xrightarrow{\mathcal{A}} a \xrightarrow{\mathcal{T}} y \xrightarrow{\mathcal{A}} o,
\]
where $o = \mathcal{A}(x, a, y)$. This formulation naturally includes the special case where the agent produces $o$ directly without calling any tools.

\paragraph{Optimization Objective.}
The goal of A2 adaptation is to optimize the agent such that its final output aligns with correctness, quality, or alignment criteria. Formally:
\begin{align}
\text{\textbf{(A2)}} \quad
\mathcal{A}^* = \arg\max_{\mathcal{A}}\, \mathcal{O}_{\text{agent}}(\mathcal{A}, \mathcal{T}),
\end{align}
where $\mathcal{O}_{\text{agent}}$ evaluates the final output $o$ generated by the agent. Similarly, A2 paradigm optimization also includes two main forms:
\begin{itemize}[leftmargin=15pt]

\item \textbf{Supervised Fine-Tuning (SFT).} Let $\mathcal{D}_{\text{ans}}=\{(x, y, a^*, o^*)\}$ denote a dataset of inputs, optional tool outputs, reference tool calls, and target final outputs. Supervising only the final output $o^*$ is insufficient for learning tool-use behavior, because the agent could improve final-answer likelihood without ever invoking tools. Effective A2-style SFT therefore combines final-output supervision with A1-style tool-call imitation:
\begin{align}
\mathcal{A}^* = \arg\max_{\mathcal{A}}\,
\mathbb{E}_{(x,y,a^*,o^*)}
\Big[
\underbrace{\log p_{\mathcal{A}}(a^*|x)}_{\text{tool-call (A1-style)}}
+
\underbrace{\log p_{\mathcal{A}}(o^*|x,a^*,y)}_{\text{final answer (A2-style)}}
\Big].
\end{align}
When no tools are invoked, $a^*$ and $y$ are empty and the objective reduces to standard answer-level SFT.

\item \textbf{Reinforcement Learning (RL).}
When explicit target outputs are unavailable, the agent learns from feedback assigned to its final response.  
The interaction follows:
\[
x \xrightarrow{\mathcal{A}} a \xrightarrow{\mathcal{T}} y \xrightarrow{\mathcal{A}} o,
\quad 
\text{with reward} \quad R = \mathcal{O}_{\text{agent}}(o).
\]
The optimization objective becomes:
\[
J(\mathcal{A})
= \mathbb{E}_{x\sim\mathcal{D}_0,\; a\sim\mathcal{A}(\cdot|x),\; y=\mathcal{T}(a),\; o = \mathcal{A}(x,a,y)}\big[\mathcal{O}_{\text{agent}}(o)\big],
\]
where $\mathcal{D}_0$ is the distribution of task inputs.  
Here, the agent receives rewards based solely on the quality of its final output, irrespective of how many intermediate tool calls were invoked.
    
\end{itemize}

\subsubsection{T1: Agent-Agnostic Tool Adaptation}
\label{subsub:t1_math}
In the T1 paradigm, the agent $\mathcal{A}$ is kept fixed, and adaptation is applied to only the external tool set $\mathcal{T}$.
The setting arises when the agent is a closed-source API (such as GPT, Claude, or Gemini) that cannot be fine-tuned, or when the goal is to augment a fixed agent by training specialized tools such as retrievers, rerankers, planners, simulators, or additional foundation models.
In this sense, a ``tool'' in T1 primarily refers to a trainable model, regardless of whether it is a traditional machine-learning model or a large-scale foundation model.

\paragraph{Optimization Objective.}
The goal of T1 is to optimize the tool in an agent-agnostic manner:
\[
\text{\textbf{(T1)}}\quad 
\mathcal{T}^{*} = \arg\max_{\mathcal{T}} \mathcal{O}_{\text{tool}}(\mathcal{T}),
\]
where $\mathcal{O}_{\text{tool}}(\mathcal{T})$ evaluates the quality of tool-produced results, often through metrics such as retrieval accuracy, ranking quality, simulation fidelity, or downstream task success. Since the agent is fixed and only $\mathcal{T}$ is trainable, T1 reduces to standard model training under various learning paradigms, such as supervised learning, contrastive learning, or reinforcement learning.

\subsubsection{T2: Agent-Supervised Tool Adaptation}
\label{subsub:t2_math}
In the T2 paradigm, tool adaptation is guided by the frozen agent $\mathcal{A}$. Unlike T1, where tools are trained independently, T2 adapts or constructs tools that complement the fixed agent and improve its overall capability. When the main agent is a closed-source foundation model, training auxiliary tools around it is often preferable to modifying the agent itself.

\paragraph{Agent-Tool Interaction Process.}
As before, we describe a single-turn interaction; the multi-turn case extends naturally. The agent receives input $x$ and produces a tool call $a=\mathcal{A}(x)$.  
The tool $\mathcal{T}$ executes this call to return a result $y=\mathcal{T}(a)$, and the agent integrates $(x,a,y)$ or $(x,y)$ to produce the final output $o$:
\[
x \xrightarrow{\mathcal{A}} a \xrightarrow{\mathcal{T}} y \xrightarrow{\mathcal{A}} o.
\]

\paragraph{Optimization Objective.}
The tool is optimized to improve the performance of the fixed agent-tool system:
\[
\text{\textbf{(T2)}} \quad
\mathcal{T}^{*} 
=
\arg\max_{\mathcal{T}}\, \mathcal{O}_{\text{agent}}(\mathcal{A}, \mathcal{T}),
\]
where $\mathcal{O}_{\text{agent}}$ evaluates how effectively the agent performs when equipped with tool~$\mathcal{T}$.  
The objective emphasizes that T2 adapts the tool specifically to the needs of the given agent. Tool adaptation in T2 generally takes two forms:

\begin{itemize}[leftmargin=15pt]

    \item \textbf{Supervised Learning.} In the supervised setting, the frozen agent provides signals that indicate how the tool should improve. The core idea is to adjust the tool so that its future outputs $\,\mathcal{T}(a)\,$ become more helpful for the agent's downstream reasoning. This can be instantiated in several ways. For example:

    \begin{itemize}[leftmargin=18pt]

        \item \textbf{Quality-Weighted Training.}
        The agent’s final output $o$ induces a quality score $w=\omega(o)$ that reflects the desirability or correctness of the agent’s behavior.
        The tool is trained by weighting each trajectory according to this score:
        \[
        \mathcal{T}^*
        = \arg\min_{\mathcal{T}}
        \; \mathbb{E}_{(a,y,o)}\!
        \Big[
        w(o)\,\ell\big(\mathcal{T}(a),\, y\big)
        \Big],
        \]
        where $\ell$ is a task-specific loss encouraging the tool’s output to improve.
        If $w(o)$ takes binary values $\{0,1\}$, this reduces to a data-selection scheme
        where only trajectories associated with desirable agent outputs $o$ are used to
        train the tool.

        \item \textbf{Output-Consistency Training.}
        The agent’s final output $o$ induces an implicit supervision target 
        $\tau=\phi(a,y,o)$,
        which prescribes how the tool output should change to better support the agent.
        The tool is updated by:
        \[
        \mathcal{T}^*
        = \arg\min_{\mathcal{T}}
        \; \mathbb{E}_{(a,y,o)}\!
        \Big[
        \ell\big(\mathcal{T}(a),\, \tau\big)
        \Big],
        \]
        where the mapping $\phi(a,y,o)$ extracts a learning target from the relationship between $y$ and $o$.
        This encourages the tool to produce outputs that more effectively align with the agent's downstream reasoning.

    \end{itemize}

    \item \textbf{Reinforcement Learning (RL).}
    The tool is updated using a scalar reward based on the final quality of the agent’s output.  
    Let $R=\mathcal{O}_{\text{agent}}(o)$ denote the reward assigned to the final output $o=\mathcal{A}(x,a, y)$.  
    The RL objective becomes:
    \[
    J(\mathcal{T})
    =
    \mathbb{E}_{x\sim\mathcal{D}_0,\; a=\mathcal{A}(x),\; y=\mathcal{T}(a),\;o=\mathcal{A}(x,a,y)}
    \big[
        \mathcal{O}_{\text{agent}}(o)
    \big],
    \]
    where $\mathcal{D}_0$ is the distribution of task inputs.

\end{itemize}

\paragraph{Memory and the Adaptation Paradigms.}
In this paper, the memory module is treated as a tool within $\mathcal{T}$. However, not all memory systems map cleanly onto a single paradigm; the appropriate classification depends on the memory's \emph{form} and \emph{update mechanism}:

\begin{itemize}[leftmargin=15pt]
    \item \textbf{External adaptive memory (predominantly T2).} When a frozen agent produces outputs that are used to update an external memory store (e.g., episodic buffers, reflective databases, skill libraries) through a fixed or learnable write function $\mathcal{M} \leftarrow \text{Update}(\mathcal{M},\, o)$, the process aligns with T2: the agent remains fixed, the adaptation signal originates from the agent's own output, and the memory module evolves to better support future reasoning.
    \item \textbf{Pre-trained or plug-in memory modules (predominantly T1).} Memory components that are trained independently of any specific agent, such as pre-trained dense retrievers, static knowledge bases, or off-the-shelf embedding indices, function as agent-agnostic tools under T1.
    \item \textbf{Parametric and hybrid memory (boundary cases).} Memory encoded within model parameters (e.g., LoRA adapters for knowledge injection, differentiable memory modules like Titans) or hybrid architectures that combine parametric and external storage (e.g., Memory$^3$) blur the boundary between tool adaptation and agent adaptation. When the memory update requires gradient-based changes to the agent's own parameters, the method is better classified under A1 or A2; when only an auxiliary parameter set is updated while the core agent remains frozen, the method occupies the T1/T2 boundary.
\end{itemize}

\noindent We adopt the convention that the \emph{dominant} form of memory adaptation determines the paradigm label. For most external, non-parametric memory systems discussed in this survey, the T2 label applies: the frozen agent's outputs supervise the evolution of the memory module. We flag boundary cases explicitly in \S\ref{sec:tool_adaptation} to avoid flattening the important architectural differences among memory forms.

\subsection{Illustrative Examples}
\label{subsec:examples}
We illustrate the adaptation paradigms through two representative settings: retrieval-augmented generation (RAG) and code-execution-based tasks. These settings highlight the central role of tool use in agentic AI while exhibiting distinct interaction patterns and evaluation protocols. 

For each application, we present paired A1 and A2 examples sharing the same tool-call action (document retrieval or code execution), allowing a direct contrast between tool-feedback-based adaptation (A1) and agent-output-based adaptation (A2). We also provide a T2 example in the RAG setting.

\subsubsection{Agent Adaptation Examples Across Two Applications}
We illustrate the A1 and A2 paradigms through two tool-use settings: retrieval-augmented generation (RAG) and code-execution-based question answering. For each, we describe the problem setup and then present paired examples that instantiate A1 and A2 under the same tool-call action.

\paragraph{Retrieval-Augmented Generation (RAG) Setting.}
In the RAG setting, the agent receives a query and performs a retrieval action to obtain relevant documents from a database. Formally, the agent produces a retrieval query $a$, the retriever returns a set of documents $y$, and the agent synthesizes these documents together with the original query to generate a final answer $o$.
\begin{itemize}[leftmargin=15pt]

    \item \textbf{A1 example}.
    DeepRetrieval~\citep{jiang2025deepretrieval} optimizes the agent using feedback signals computed directly from retrieval quality.  
    After generating a retrieval query $a$, the retriever returns documents $y$, and metrics such as recall or nDCG are computed from $y$ and used as the reward for updating the agent.  
    Since the adaptation signal depends solely on the tool-execution outcome, this represents the A1 paradigm.

    \item \textbf{A2 example}.
    Search-R1~\citep{jin2025search} follows the full RAG pipeline, where the agent first retrieves documents and then integrates them into its context to produce a final answer $o$.  
    The adaptation signal is computed from the correctness or quality of this final answer by calculating exact matching accuracy.  
    Because the optimization is guided by the agent’s final output rather than the retrieval result alone, this falls under the A2 paradigm.

\end{itemize}

\noindent\textit{Contrast.} DeepRetrieval (A1) directly optimizes retrieval quality but provides no gradient signal for answer synthesis; Search-R1 (A2) optimizes end-to-end answer correctness but must discover good retrieval strategies as a side effect, making credit assignment harder.

\paragraph{Code-Execution-Based Task Setting.}
In code-execution-based tasks, the agent receives a problem description and produces executable code as the tool-call action. The sandbox executes the code and returns an execution result $y$, which the agent may optionally use to generate a final answer $o$. 

\begin{itemize}[leftmargin=15pt]

    \item \textbf{A1 example:} DeepSeek-R1 (code)~\citep{guo2025deepseek}.
    During reinforcement learning, DeepSeek-R1 generates code that is executed inside a sandbox (when the reward is derived solely from test-case pass rates, this is A1).  
    The execution output, such as test-case pass rate or numerical correctness, is used directly as the reward for policy optimization.  
    Since adaptation is based entirely on the tool’s execution result, this example fits the A1 paradigm.

    \item \textbf{A2 example:}
    ReTool~\citep{feng2025retool} also generates executable code, but the sandbox result is fed back into the agent as additional context.  
    The agent then produces a final answer $o$, whose correctness determines the reward.  
    Because the adaptation signal depends only on the final output of the agent after integrating tool feedback, this corresponds to the A2 paradigm.

\end{itemize}

\noindent\textit{Contrast.} DeepSeek-R1's A1 formulation ties the reward to test-case pass rates, giving the agent a precise signal per code invocation. ReTool's A2 formulation, by contrast, rewards only the final answer, so the model must learn \emph{when} to invoke the code sandbox---a richer but noisier learning problem.

\subsubsection{Tool Adaptation Examples in the RAG Setting}
In many practical systems, the central agent is a closed-source API model (e.g., GPT, Claude, or Gemini) that cannot be fine-tuned. Training an open-source model to match such performance is challenging due to data curation requirements, scaling laws, and infrastructure demands. A more feasible strategy is to treat the closed-source model as a fixed agent and adapt auxiliary tools around it. In the RAG setting, this motivates tool adaptation for components such as retrievers.

\paragraph{T1 examples.}
\noindent\textbf{(1) Classic Dense Retrievers.}
A standard dense retriever (bi-encoder trained with contrastive learning) is the canonical T1 tool: given a query $a$, it returns documents $y=\mathcal{T}(a)$ optimized for recall, and any frozen agent can consume $y$ for downstream reasoning without having participated in the retriever's training.
\textbf{(2) Learned Subagents as Agent-Agnostic Tools.} Beyond classic dense retrievers, once agent adaptation has produced strong retrieval-oriented models, these learned models can themselves be reused as tools under the T1 paradigm. For example, a model trained in the DeepRetrieval~\cite{jiang2025deepretrieval} style can be deployed purely as a high-quality subagent that rewrites queries for improved retrieval over specific document databases. Given a tool call action $a$ (a retrieval query), the subagent returns a reformulated query or a curated document set $y=\mathcal{T}(a)$ with higher recall or relevance, which is then consumed by a fixed closed-source agent that performs the final reasoning and answer synthesis.

\paragraph{T2 examples.}
Under the T2 paradigm, the tool is adapted using supervision signals derived directly from a fixed agent’s final outputs.  
In the RAG setting, both \textsc{s3}~\citep{jiang2025s3} and AgentFlow~\cite{li2025flow} provide representative examples. The tool (a learnable search subagent) is updated based on the fixed agent's output signal so that its behavior becomes increasingly aligned with what the fixed agent needs for successful downstream reasoning.

Concretely, in \textsc{s3}~\citep{jiang2025s3}, given a question $x$, the learnable subagent ~$\mathcal{T}$ takes $x$ as input and internally generates a retrieval query $a'=\mathcal{T}(x)$.  
This query is executed on a static search engine to retrieve a document set $y$, which is then inserted into the frozen agent’s context.  
The fixed agent consumes $(x,y)$ and produces a final answer $o=\mathcal{A}(x,y)$. An evaluation function $\mathcal{O}_{\text{agent}}(o)$ (e.g., answer correctness) assigns a scalar reward, which is then used to update the tool so that its future retrieval behaviors yield document sets that more effectively support the agent’s downstream reasoning. Thus, \textsc{s3} directly realizes the T2 objective with the agent-output signaled tool adaptation specifically for the fixed agent’s downstream performance. AgentFlow~\citep{li2025flow} further extends this idea by training a more expressive planning-oriented subagent capable of multi-tool decision-making, applying T2 supervision signal to align a richer planning policy with the fixed agent’s final-output preferences.

\begin{figure*}[t]
    \centering
    \includegraphics[width=\textwidth]{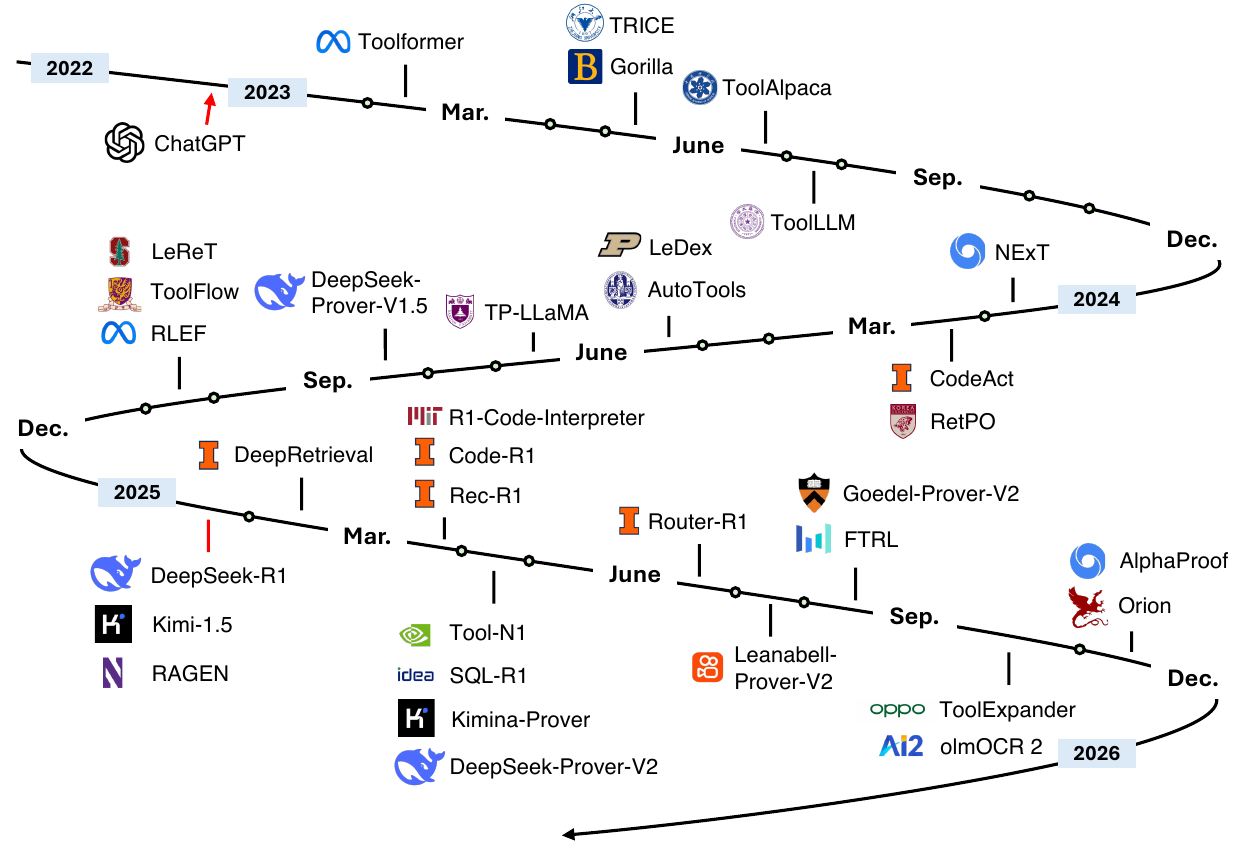}
    \caption{
    Development timeline of A1 methods (agent adaptation with tool-execution result as signal).
    }
    \label{fig:a1_timeline}
\end{figure*}
\section{Agent Adaptation}
\label{sec:agent_adaptation}

Agent adaptation refers to the mechanisms through which agents refine their behavior based on feedback from tools, environments, or their own outputs. A central design choice is \emph{where} the feedback signal originates, which determines both what the agent can learn and what failure modes remain invisible.  When the signal comes from tool execution (\textbf{A1}), the agent receives dense, causally grounded feedback tied to specific actions---but this signal is blind to whether the agent's overall reasoning strategy is sound.  When the signal comes from evaluating the agent's own outputs (\textbf{A2}), the agent can optimize holistic task success---but the sparser, episode-level reward makes credit assignment harder and training less data-efficient.  This A1/A2 tension between \emph{mechanistic precision} and \emph{strategic flexibility} structures the design space reviewed in this section.

Formally, let $\mathcal{A}$ denote an agent parameterized by its internal configuration or policy (prompt templates or model weights), and let $\mathcal{T}$ represent the set of accessible tools. 
The two paradigms correspond to distinct optimization objectives:
\[
\textbf{(A1)} \quad \mathcal{A}^{*} = \arg\max_{\mathcal{A}} \mathcal{O}_{\text{tool}}(\mathcal{A}, \mathcal{T}),
\qquad
\textbf{(A2)} \quad \mathcal{A}^{*} = \arg\max_{\mathcal{A}} \mathcal{O}_{\text{agent}}(\mathcal{A}, \mathcal{T}),
\]
where $\mathcal{O}_{\text{tool}}$ scores the correctness or utility of tool execution outcomes (e.g., code compilation success, retrieval precision), and $\mathcal{O}_{\text{agent}}$ scores the quality of the agent's generated outputs (reasoning validity, factual accuracy, or preference alignment). 
The distinction has practical consequences: $\mathcal{O}_{\text{tool}}$ yields per-action credit but is silent about reasoning quality, whereas $\mathcal{O}_{\text{agent}}$ captures end-to-end performance but conflates tool-use skill with reasoning skill, complicating diagnosis when either degrades.

\subsection{A1: Tool Execution Result as Signal}
\label{subsec:tool_execution_signal}

A1 adaptation treats tool and environment outputs as ground-truth supervision: because the signal is determined by external execution rather than model-internal beliefs, it is verifiable, reproducible, and dense enough to support both supervised and reinforcement-based learning.  The key challenge is that such signals optimize \emph{individual actions} (a single API call, a single code execution) and can miss system-level failures that only surface when multiple actions compose.

\definecolor{sectionblue}{RGB}{230,245,255}
\definecolor{sectiongreen}{RGB}{235,255,235}
\definecolor{cardinal}{RGB}{180,0,0}

\definecolor{white}{RGB}{255,255,255}

\small
%
%
\rowcolors*{1}{gray!7}{white}
\renewcommand{\arraystretch}{1.15}

\begin{longtable}[t]{>{\raggedright\arraybackslash}p{1.2cm}
                  >{\raggedright\arraybackslash}p{2.0cm}
                  >{\raggedright\arraybackslash}p{1.5cm}
                  >{\raggedright\arraybackslash}p{2.4cm}
                  >{\raggedright\arraybackslash}p{2.5cm}
                  >{\raggedright\arraybackslash}p{2.5cm}
                  >{\raggedright\arraybackslash}p{1.8cm}
                  >{\raggedright\arraybackslash}p{0.8cm}}
\rowcolor{white}
\caption{A1 Methods (Tool Execution Signaled): Earlier Methods (SFT \& DPO) and Recent RLVR-based Methods} \\

\toprule
\rowcolor{white}
\textbf{Time} & \textbf{Method} & \textbf{Venue} & \textbf{Task(s)} & \textbf{Tool(s)} & \textbf{Agent Backbone} & \textbf{Tuning} & \textbf{Links} \\
\midrule
\endfirsthead

\rowcolor{white}
\multicolumn{8}{c}{\tablename\ \thetable\ -- Continued from previous page} \\
\toprule
\rowcolor{white}
\textbf{Time} & \textbf{Method} & \textbf{Venue} & \textbf{Task(s)} & \textbf{Tool(s)} & \textbf{Agent Backbone} & \textbf{Tuning} & \textbf{Links} \\
\midrule
\endhead

\rowcolor{white}
\multicolumn{8}{r}{\textit{Continued on next page}} \\
\endfoot

\bottomrule
\endlastfoot

\rowcolor{sectionblue}\multicolumn{8}{c}{\textbf{SFT \& Off-Policy Methods}} \\
\midrule
2023.02 & Toolformer & NeurIPS'23 & QA, Math & Calculator, QA system, Search Engine, Translation System, Calendar & GPT-J & SFT & \href{https://arxiv.org/abs/2302.04761}{\textcolor{cardinal}{\faFilePdf}} \href{https://github.com/conceptofmind/toolformer}{\faGithub} \\
2023.05 & TRICE & NAACL'24 & Math Reasoning, QA & Calculator, WikiSearch, Atlas QA Model, NLLB Translator & ChatGLM, Alpaca, Vicuna & SFT, Contrastive Learning & \href{https://arxiv.org/abs/2305.13068}{\textcolor{cardinal}{\faFilePdf}} \href{https://github.com/zjunlp/TRICE}{\faGithub} \\
2023.05 & Gorilla & NeurIPS'24 & Tool-Calling, API Retrieval &  APIs & LLaMA & SFT & \href{https://arxiv.org/abs/2305.15334}{\textcolor{cardinal}{\faFilePdf}} \href{https://github.com/ShishirPatil/gorilla}{\faGithub} \\
2023.06 & ToolAlpaca & arXiv & Multi-Turn Tool-Use & Simulated APIs & Vicuna & SFT & \href{https://arxiv.org/abs/2306.05301}{\textcolor{cardinal}{\faFilePdf}} \href{https://github.com/tangqiaoyu/ToolAlpaca}{\faGithub} \\
2023.07 & ToolLLM & ICLR'24 & Tool-Calling, API Planning, Multi-Tool Reasoning & Real-World APIs & LLaMA, Vicuna & SFT & \href{https://arxiv.org/abs/2307.16789}{\textcolor{cardinal}{\faFilePdf}} \href{https://github.com/OpenBMB/ToolBench}{\faGithub} \\
2024.01 & NExT & ICML'24 & Program Repair & Code Executor & PaLM2 & SFT & \href{https://arxiv.org/abs/2404.14662}{\textcolor{cardinal}{\faFilePdf}} \\
2024.02 & CodeAct & ICML'24 & Coding & Code Executor & LLaMA2, Mistral & SFT & \href{https://arxiv.org/abs/2402.01030}{\textcolor{cardinal}{\faFilePdf}} \href{https://github.com/xingyaoww/code-act}{\faGithub} \\
2024.02 & RetPO & NAACL'25 & IR & Retriever & LLaMA2-7B& SFT, DPO & \href{https://arxiv.org/abs/2402.11827}{\textcolor{cardinal}{\faFilePdf}} \href{https://github.com/dmis-lab/RetPO}{\faGithub}\\
2024.03 & CYCLE &OOPSLA'24& Coding &Code Executor &CodeGen, StarCoder &SFT& \href{https://arxiv.org/abs/2403.18746}{\textcolor{cardinal}{\faFilePdf}} \\
2024.05 & AutoTools & WWW'25 & Tool-Calling &  APIs & GPT4, LLaMA3, Mistral & SFT & \href{https://arxiv.org/abs/2405.16533}{\textcolor{cardinal}{\faFilePdf}} \href{https://github.com/mangopy/AutoTools}{\faGithub} \\
2024.06 & TP-LLaMA & NeurIPS'24 & Tool-Calling &  APIs & LLaMA2 & SFT, DPO & \href{https://arxiv.org/abs/2406.07115}{\textcolor{cardinal}{\faFilePdf}} \\
2024.10 & ToolFlow & NAACL'25 & Tool-Calling &  APIs & LLaMA3.1 & SFT & \href{https://arxiv.org/abs/2410.18447}{\textcolor{cardinal}{\faFilePdf}} \\
2024.10 & LeReT & ICLR'25 & IR & Dense Retriever & LLaMA3, Gemma2 & DPO-like (IPO) & \href{https://arxiv.org/abs/2410.23214}{\textcolor{cardinal}{\faFilePdf}} \href{https://github.com/sher222/LeReT}{\faGithub} \\

\midrule
\rowcolor{sectiongreen}\multicolumn{8}{c}{\textbf{RLVR Methods}} \\
\midrule
2024.05 & LeDex & NeurIPS'24 & Coding & Code Executor & StarCoder \& CodeLlaMA & SFT, PPO & \href{https://arxiv.org/abs/2405.18649}{\textcolor{cardinal}{\faFilePdf}} \\
2024.08 & DeepSeek-Prover-V1.5 & ICLR'25 & Formal Theorem Proving & Lean 4 Prover & DeepSeek-Prover-V1.5-RL & SFT, GRPO & \href{https://arxiv.org/abs/2408.08152}{\textcolor{cardinal}{\faFilePdf}} \href{https://github.com/deepseek-ai/DeepSeek-Prover-V1.5}{\faGithub} \\
2024.10 & RLEF & ICML'25 & Coding & Code Executor & LLaMA3.1 & PPO & \href{https://arxiv.org/abs/2410.02089}{\textcolor{cardinal}{\faFilePdf}} \\
2025.01 & DeepSeek-R1-Zero (Code) & Nature & Coding & Code Executor & DeepSeek-V3-Base & GRPO & \href{https://arxiv.org/abs/2501.12948}{\textcolor{cardinal}{\faFilePdf}}  \\
2025.02 & DeepRetrieval & COLM'25 & Web Search, IR, Text2SQL & Search Engine, Retrievers, SQL exec. & Qwen2.5, LLaMA3.2 & PPO, GRPO & \href{https://arxiv.org/abs/2503.00223}{\textcolor{cardinal}{\faFilePdf}} \href{https://github.com/pat-jj/DeepRetrieval}{\faGithub} \\
2025.03 & Code-R1 & --- & Coding & Code Executor & Qwen2.5 & GRPO & \href{https://github.com/ganler/code-r1}{\faGithub} \\
2025.03 & ReZero & arXiv & Web Search, IR & Web Search Engine & LLaMA3.2 & GRPO & \href{https://arxiv.org/abs/2504.11001}{\textcolor{cardinal}{\faFilePdf}} \href{https://github.com/janhq/ReZero}{\faGithub} \\
2025.03 & Rec-R1 & TMLR'25 & Recommendation Optimization & Recommendation System & Qwen2.5, LLaMA3.2 & GRPO & \href{https://openreview.net/forum?id=YBRU9MV2vE}{\textcolor{cardinal}{\faFilePdf}} \href{https://github.com/linjc16/Rec-R1}{\faGithub} \\
2025.04 & SQL-R1 & NeurIPS'25 & Text2SQL Search & SQL Engine & Qwen2.5, OmniSQL & SFT, GRPO & \href{https://arxiv.org/abs/2504.08600}{\textcolor{cardinal}{\faFilePdf}} \href{https://github.com/DataArcTech/SQL-R1}{\faGithub} \\
2025.04 & Kimina-Prover & arXiv & Formal Theorem Proving & Lean 4 Compiler, Numina Lean Server & Qwen2.5 & SFT, GHPO & \href{https://arxiv.org/abs/2504.11354}{\textcolor{cardinal}{\faFilePdf}} \href{https://github.com/MoonshotAI/Kimina-Prover-Preview}{\faGithub} \\
2025.04 & DeepSeek-Prover-V2 & arXiv & Formal Theorem Proving & Lean 4 Compiler & DeepSeek-V3 & SFT, GRPO & \href{https://arxiv.org/abs/2504.21801}{\textcolor{cardinal}{\faFilePdf}} \href{https://github.com/deepseek-ai/DeepSeek-Prover-V2}{\faGithub} \\
2025.05 & Tool-N1 & arXiv & Tool-Calling & Tool APIs & Qwen2.5 & GRPO & \href{https://arxiv.org/abs/2505.00024}{\textcolor{cardinal}{\faFilePdf}} \href{https://github.com/NVlabs/Tool-N1}{\faGithub} \\
2025.05 & R1-Code-Interpreter & arXiv & Coding & Code Execution Sandbox & Qwen2.5 & GRPO & \href{https://arxiv.org/abs/2505.21668}{\textcolor{cardinal}{\faFilePdf}} \href{https://github.com/yongchao98/R1-Code-Interpreter}{\faGithub} \\
2025.06 & Router-R1 & NeurIPS'25 & Multi-Round Routing & LLM Routing Pool & Qwen2.5, LLaMA3.2 & PPO & \href{https://arxiv.org/abs/2506.09033}{\textcolor{cardinal}{\faFilePdf}} \href{https://github.com/ulab-uiuc/Router-R1}{\faGithub} \\
2025.07 & Leanabell-Prover-V2 & arXiv & Formal Theorem Proving & Lean 4 Verifier & Kimina, DeepSeek-V2 & SFT, DAPO & \href{https://arxiv.org/abs/2507.08649}{\textcolor{cardinal}{\faFilePdf}} \href{https://github.com/Leanabell-LM/Leanabell-Prover-V2}{\faGithub} \\
2025.08 & Goedel-Prover-V2 & arXiv & Formal Theorem Proving & Lean Compiler & Qwen3 & SFT, GRPO & \href{https://arxiv.org/abs/2508.03613}{\textcolor{cardinal}{\faFilePdf}} \href{https://github.com/Goedel-LM/Goedel-Prover-V2}{\faGithub} \\
2025.08 & FTRL & arXiv & Multi-Step Tool-Use & Simulated APIs & Qwen3 & GRPO & \href{https://arxiv.org/abs/2508.08791}{\textcolor{cardinal}{\faFilePdf}} \href{https://github.com/bytedance/FTRL}{\faGithub} \\
2025.09 & Tool-R1 & arXiv & Tool-Augmented Reasoning, QA & Code Execution, Multimedia Tools & Qwen2.5 & GRPO & \href{https://arxiv.org/abs/2509.12867}{\textcolor{cardinal}{\faFilePdf}} \href{https://github.com/YBYBZhang/Tool-R1}{\faGithub} \\
2025.09 & WebGen-Agent & arXiv & Website Generation & VLM, GUI Agent, Code Executor & Qwen2.5-Code, Qwen3 & SFT, Step-GRPO & \href{https://arxiv.org/abs/2509.22644}{\textcolor{cardinal}{\faFilePdf}} \href{https://github.com/mnluzimu/WebGen-Agent}{\faGithub} \\
2025.10 & ToolExpander & arXiv & Tool-Calling &  Tool APIs & Qwen2.5 & SFT, GRPO & \href{https://arxiv.org/abs/2510.07737}{\textcolor{cardinal}{\faFilePdf}} \\
2025.10 & AlphaProof & Nature & Formal Theorem Proving  & Lean Solver & Transformer (3B Enc-Dec) & SFT, AlphaZero, TTRL & \href{https://www.nature.com/articles/s41586-025-09833-y}{\textcolor{cardinal}{\faFilePdf}}\\
2025.10 & olmOCR2 & arXiv & Document OCR & Synthetic Document Verifier & Qwen2.5-VL & SFT, GRPO & \href{https://arxiv.org/abs/2510.19817}{\textcolor{cardinal}{\faFilePdf}} \href{https://github.com/allenai/olmocr}{\faGithub} \\
2025.11 & Orion & arXiv & IR & Retrievers & LFM2 & GRPO & \href{https://arxiv.org/abs/2511.07581}{\textcolor{cardinal}{\faFilePdf}}\\
\end{longtable}
\normalsize

\subsubsection{Earlier Works: SFT \& Off-Policy Methods}
\label{subsubsec:3.1.1}

Early A1-type methods train agents from pre-collected data via SFT or DPO.  Their shared limitation is an off-policy data distribution: the agent learns from trajectories it did not generate, which can cause a mismatch between training and deployment behavior.  Within this family, methods diverge along a key axis---\textbf{what counts as correct?}---progressing from answer-level correctness to format-level validity to raw execution outcomes.

\textbf{Toolformer}~\cite{schick2023toolformer} (NeurIPS 2023) illustrated the earliest point on this axis: tool outcomes serve as \textbf{self-supervised signals}. 
The model inserts candidate API calls into text, executes them, and retains a call only if it reduces perplexity beyond a threshold ($L_i^- - L_i^+ \ge \tau_f$).
Because the supervision is derived from the model's own likelihood rather than an external correctness criterion, Toolformer can discover \emph{when} tools help but cannot distinguish \emph{how well} a tool call was constructed.
Subsequent work addressed this gap through three progressively stronger notions of correctness:
\begin{itemize}
\item \textbf{Golden-answer alignment:} the tool-augmented output must match a known correct response or expert trajectory.
\item \textbf{Golden-format alignment:} the tool call must be structurally and syntactically valid, independent of downstream output.
\item \textbf{Direct-execution alignment:} supervision emerges from actually running the tool, closing the gap between training signal and deployment behavior.
\end{itemize}

\textbf{\underline{Golden-answer alignment.}}
The simplest instantiation defines correctness as matching a known answer or expert trajectory.  A key question within this family is \emph{how to exploit negative examples}: early methods discard failed trajectories, while later ones treat them as contrastive signal.

\noindent\textbf{ToolAlpaca}~\cite{tang2023toolalpaca} pioneered the closed-loop generate-execute-evaluate-finetune cycle, where the model calls a tool, observes the runtime outcome (returned values, errors, or completion states), and retrains on the result.  Iteration over this loop aligns the model's internal representation with actual tool semantics and enables generalization to unseen tools---but only positive outcomes contribute to learning.
\textbf{TRICE}~\cite{qiao2024making} (NAACL 2024) introduces a ranking loss that scores candidate responses by comparing execution results against ground-truth answers, teaching the model to invoke tools only when they improve accuracy.
\textbf{TP-LLaMA}~\cite{chen2024advancing} (NeurIPS 2024) goes further by explicitly mining failure information that ToolAlpaca and earlier systems (e.g., ToolLLaMA~\cite{qin2024toolllm}) discard: at each decision node along a successful trajectory, the expert's correct next step is preferred ($y_w$) and any failed branch is dispreferred ($y_l$), and DPO on these pairs converts failure signals into contrastive supervision.  The progression from ToolAlpaca to TP-LLaMA thus illustrates a recurring theme in A1 methods: the richer the exploitation of negative feedback, the more data-efficient the adaptation.

\textbf{\underline{Golden-format alignment.}}
A complementary axis defines correctness structurally: a tool call is correct if it is syntactically and logically valid, regardless of whether the final answer is right.  This decoupling is valuable when ground-truth answers are expensive to obtain but canonical call formats exist.

\noindent\textbf{Gorilla}~\cite{patil2024gorilla} (NeurIPS 2024) fine-tunes a retrieval-augmented LLaMA model on a large set of machine learning APIs, defining correctness via Abstract Syntax Tree (AST) subtree matching---more robust than string matching because it tolerates differences in parameter order or optional arguments.
\textbf{ToolFlow}~\cite{wang2025toolflow} (NAACL 2025) complements Gorilla by tackling data quality: a tool graph built from parameter and return-value similarities guides the selection of interacting tool subsets, and a planned generation step organizes multi-turn requests for logical consistency.  Together, these two works show that format-level supervision can scale to large, evolving API surfaces where golden answers are impractical, but they also reveal a limitation: format correctness does not guarantee functional correctness, so models trained this way can produce well-formed but semantically wrong calls.

\textbf{\underline{Direct-execution alignment.}}
The strongest form of A1 supervision closes the loop entirely: the tool is executed, and its output---success, failure, returned values---becomes the training signal.  This eliminates the need for pre-labeled answers or canonical formats but introduces a new challenge: designing execution environments that are safe, reproducible, and informative enough to support learning.

Two design patterns have emerged.  The first generates executable code as the action representation.
\textbf{CodeAct}~\cite{wang2024executable} (ICML 2024) replaces textual or JSON-based commands with sandboxed code actions, so the environment's execution feedback directly grounds the model in tool causality.
\textbf{NExT}~\cite{ni2024next} (ICML 2024) applies the same principle to program repair via an iterative Sample-Filter-Train loop: candidate fixes are validated by unit tests, and only passing solutions enter the next fine-tuning round, yielding progressively sharper execution-aware rationales.
The second pattern targets autonomous tool discovery.
\textbf{ToolLLM}~\cite{qin2024toolllm} and \textbf{AutoTools}~\cite{shi2025tool} (WWW 2025) have the LLM itself parse raw API documentation into callable functions (tool encapsulation) and then compose multi-function programs to solve user queries (tool programming), with execution outcomes driving iterative self-improvement.

A third pattern applies execution-grounded preference learning to retrieval.
\textbf{LeReT}~\cite{hsugrounding} (ICLR 2025) generates diverse search queries, scores retrieved documents via a reward function, and fine-tunes the query generator through Identity Policy Optimization (IPO):
\[
\mathcal{L}_{\text{IPO}} = \mathbb{E}_{(x, y_w, y_l) \sim D_p} \Big[ (\tilde{r}_\phi(x, y_w) - \tilde{r}_\phi(x, y_l) - 0.5\tau^{-1} )^2 \Big],
\]
where $\tilde{r}_\phi(x, y) = \log \frac{\pi_\phi(y|x)}{\pi_\text{ref}(y|x)}$ and $\tau$ controls the target margin.  Because the reward comes from running the retriever rather than from labeled answers, LeReT adapts to arbitrary off-the-shelf retrievers without modifying the generator.
\textbf{RetPO}~\cite{yoon2025ask} (NAACL 2025) follows the same logic at lower cost: GPT-4 generates candidate query rewrites, an off-the-shelf retriever (e.g., BM25) scores each by retrieval quality, and a smaller open-source LM is trained via DPO to produce high-reward rewrites.

\paragraph{Summary of SFT \& off-policy A1 methods.}
The three alignment stages trace a clear arc: from Toolformer's self-supervised likelihood reduction, through answer- and format-level matching, to direct execution feedback.  Each step tightens the coupling between training signal and deployment behavior, reducing the ``reality gap'' between what the model optimizes and what determines success at inference time.  Yet all off-policy methods share a fundamental ceiling: because they learn from pre-collected trajectories, they cannot explore novel tool-use strategies that were absent from the training distribution.  This limitation motivates the on-policy RLVR methods described next.

\subsubsection{RLVR-Based Methods}
\label{subsubsec:3.1.2}

Reinforcement learning with verifiable reward (RLVR) removes the off-policy ceiling of SFT and DPO by letting the agent explore on-line: it proposes actions, executes them, observes verifiable outcomes, and updates its policy---all within the same training loop.  The on-policy setting introduces two new design axes absent from the methods above: (i)~\emph{reward density}---whether feedback arrives per step (as in theorem proving) or only at episode end (as in multi-tool reasoning)---which governs credit assignment difficulty; and (ii)~\emph{reward composition}---how task-specific, format, and regularization terms are combined.  We organize the domain-specific instantiations below and distill cross-domain principles at the end.

\textbf{\underline{Web search and information retrieval.}}
\textbf{DeepRetrieval}~\cite{jiang2025deepretrieval} (COLM 2025) established the template: query reformulation is cast as an MDP with retrieval metrics (Recall@K, NDCG, or SQL execution accuracy) as the reward, optimized via KL-regularized PPO:
\[
\hat{\pi} = \arg\max_{\pi} \, \mathbb{E}_{q, q' \sim \pi} \big[ r(q, q') - \beta \log \tfrac{\pi(q'|q)}{\pi_{\text{ref}}(q'|q)} \big],
\]
where $r(q, q') = r_{\text{retrieval}}(q, q') + r_{\text{format}}(q')$.  A single framework thereby adapts across literature search, QA retrieval, and text-to-SQL, yielding roughly a threefold recall improvement (65.1\% vs.\ 24.7\%) on real-world search engines.

Subsequent work addresses two limitations of this single-step formulation.
\textbf{ReZero}~\cite{dao2025rezero} adds retry-aware reward shaping via GRPO, teaching the agent to persist after failed searches in partially observable web environments.
\textbf{Orion}~\cite{vijay2025think} extends from single-step to multi-turn adaptive search, using turn-level rewards based on normalized similarity and rank; notably, compact 350M--1.2B models learn effective multi-hop search strategies, demonstrating that on-policy RLVR can substitute for large-controller scale when reward density is sufficient.

\textbf{\underline{Code-based tools.}}
Code execution provides a near-ideal A1 environment: feedback is deterministic, sandboxable, and obtainable at every compilation or test-run step.  The central design question is how to keep this feedback \emph{reliable} as tasks grow more heterogeneous.
\textbf{LeDex}~\cite{jiang2024ledex} (NeurIPS 2024) uses a composite PPO reward combining unit-test correctness (CodeBLEU) with explanation quality (semantic similarity), illustrating the value of multi-faceted rewards.
\textbf{RLEF}~\cite{gehring2025rlefgroundingcodellms} (ICML 2025) formalizes code synthesis as a partially observable MDP where the agent generates, receives public-test feedback, and iterates---showing that multi-turn interaction is key to solving problems beyond the model's single-shot capability.
\textbf{Code-R1}~\cite{code-r1} shifts focus from the policy to the \emph{reward pipeline}: by eliminating false positives from faulty tests, unsolvable prompts, and mismatched sandboxes, it demonstrates that reward quality dominates reward quantity.
\textbf{R1-Code-Interpreter}~\cite{chen2025r1} addresses the heterogeneity of code-interpreter tasks (math, retrieval, data analysis) through multi-stage curriculum learning that prioritizes samples by their improvement potential, mitigating sparse and unstable rewards.
\textbf{Tool-R1}~\cite{zhang2025tool} tackles sample efficiency directly via a dynamic sample queue that caches and reuses high-quality trajectories, reducing the exploration cost of multi-step reasoning.

\textbf{\underline{Formal theorem proving}}~\cite{polu2020generative,leandojo,lin2025goedel,xindeepseek,wang2025kimina,chen2025seed,song2025leancopilotlargelanguage,tsoukalas2024putnambench,yang2024formalmathematicalreasoningnew} provides a canonical domain for RLVR under the A1 paradigm, as proof assistants provide ground-truth, tool-execution-signaled feedback at every step.
In this setting, the agent proposes one or more tactics (i.e., proof steps), a formal proof checker (the tool) deterministically verifies their validity, and the resulting validated proof-state transition is returned to the agent.
The verification outcome (e.g., whether a tactic is accepted, whether it advances the proof state, or whether a complete proof is achieved) serves directly as a verifiable reward signal for policy optimization.
Compared to code-execution RLVR, where unit tests may be sparse or incomplete, theorem proving offers step-wise semantic verification with minimal ambiguity, enabling denser rewards and substantially easing long-horizon credit assignment.
Recent systems such as \textbf{AlphaProof}~\cite{hubert2025olympiad} (Nature 2025), \textbf{DeepSeek-Prover-V2}~\cite{ren2025deepseek} (ICLR 2025), \textbf{Kimina-Prover}~\cite{wang2025kimina}, and \textbf{Leanabell-Prover-V2}~\cite{ji2025leanabell} use this verifier feedback to train multi-step proof search policies via reinforcement learning,
while a complementary line of work augments the native proof checker feedback with auxiliary guidance signals to prioritize trajectories, shape exploration, or stabilize optimization on top of verifier-grounded rewards~\cite{xin2025scaling,huang2025leanprogress,achim2025aristotleimolevelautomatedtheorem,sanchez2025qedcartographer,wu2024internlm2}.
While RLVR is well suited for learning proof strategies under a fixed prover snapshot, formal theorem proving also highlights a broader adaptation challenge: formal libraries (e.g., \textsc{mathlib}~\cite{mathlib}) and large, actively evolving formalization projects~\cite{gowers2023conjecturemarton} built by the Lean community grow continuously, expanding the available premise space.
Addressing this non-stationarity often requires complementary continual or low-resource adaptation mechanisms beyond pure RLVR, which we discuss in \S\ref{subsec:continual_adapt}.

\textbf{\underline{Multi-tool reasoning systems.}}
When multiple tools must be composed sequentially or conditionally, the agent faces a combinatorial action space and sparser episode-level rewards.  Methods in this category address the challenge through three complementary strategies: (i)~\emph{routing}, (ii)~\emph{environment construction}, and (iii)~\emph{dense step-level rewards}.

\textbf{Router-R1}~\cite{zhang2025router} (NeurIPS 2025) exemplifies routing: a policy LLM learns to alternate between internal reasoning and external model selection, dynamically invoking different LLMs from a routing pool.
\textbf{FTRL}~\cite{ye2025feedback} tackles environment construction by automatically synthesizing diverse tool-use training environments through a multi-stage pipeline, so the agent can train without hand-crafted external toolsets; a verifiable reward balances invocation accuracy with task completion.
\textbf{Tool-N1}~\cite{zhang2025nemotron} separates internal reasoning (\texttt{<think>} tags) from external execution (\texttt{<tool\_call>} tags), giving the RL optimizer distinct credit-assignment channels for thought and action.
\textbf{WebGen-Agent}~\cite{lu2025webgen} demonstrates the dense-reward strategy: appearance scores from a visual-language model and functionality scores from a GUI-agent are combined via Step-GRPO to supervise interactive website code generation at every turn.
\textbf{ToolExpander}~\cite{chen2025toolexpander} targets the resource-constrained regime with Dynamic Multi-Round Hard Sampling (replacing hard samples with few-shot exemplars) and a Self-Exemplifying Thinking mechanism that rewards self-generated analysis, stabilizing GRPO for small LLMs.

\textbf{\underline{Domain-specific extensions.}}
A1-type RLVR has also been applied beyond the core domains above.
\textbf{Rec-R1}~\cite{lin2025rec} (TMLR 2025) casts recommendation as an RL policy over LLM generation, using NDCG and Recall as rewards.
\textbf{SQL-R1}~\cite{ma2025sql} uses a composite reward (format, execution, result, and length) for NL-to-SQL.
\textbf{olmOCR 2}~\cite{poznanski2025olmocr2} replaces the SFT-based teacher imitation of olmOCR 1~\cite{poznanski2025olmocr} with diverse binary unit tests (text presence, reading order, table accuracy, formula rendering) as the reward signal for document OCR.
These applications reinforce a common pattern: whenever a domain provides a cheap, deterministic correctness oracle, A1-style RLVR can yield large gains with modest engineering effort.

\paragraph{Cross-domain principles.}
Despite the diversity of domains (web search, code, theorem proving, multi-tool reasoning, and specialized applications), four recurring patterns cut across all A1 RLVR work:
\begin{itemize}[leftmargin=12pt]
    \item \textbf{Signal density determines learning efficiency.} Domains with dense, per-step feedback (theorem proving, code execution) enable faster convergence than those with sparse, episode-level rewards (multi-tool reasoning). Dense feedback explains why A1 methods in code and proof domains often require less data than A2 counterparts.
    \item \textbf{Reward quality dominates reward quantity.} Code-R1~\cite{code-r1} and DeepRetrieval~\cite{jiang2025deepretrieval} both show that investing in clean, verified reward signals yields larger gains than scaling training data, a finding consistent across retrieval, code, and SQL domains.
    \item \textbf{Format rewards are necessary but not sufficient.} Nearly all successful RLVR methods combine a task-specific reward with a format compliance reward (e.g., DeepRetrieval's $r_{\text{format}}$, Tool-N1's structured output tags). Format rewards alone cannot drive meaningful adaptation, but their absence leads to degenerate outputs.
    \item \textbf{Stabilization mechanisms are domain-agnostic.} KL regularization, dynamic sampling, and curriculum scheduling appear across all domains, suggesting that the challenge of stabilizing on-policy RL for language agents is fundamental rather than domain-specific.
\end{itemize}
These principles notwithstanding, RLVR remains expensive: it requires careful reward design, interactive compute, and explicit stabilization---costs that motivate the tool-centric adaptation paradigms discussed in later sections.

\begin{figure*}[t]
    \centering
    \includegraphics[width=\textwidth]{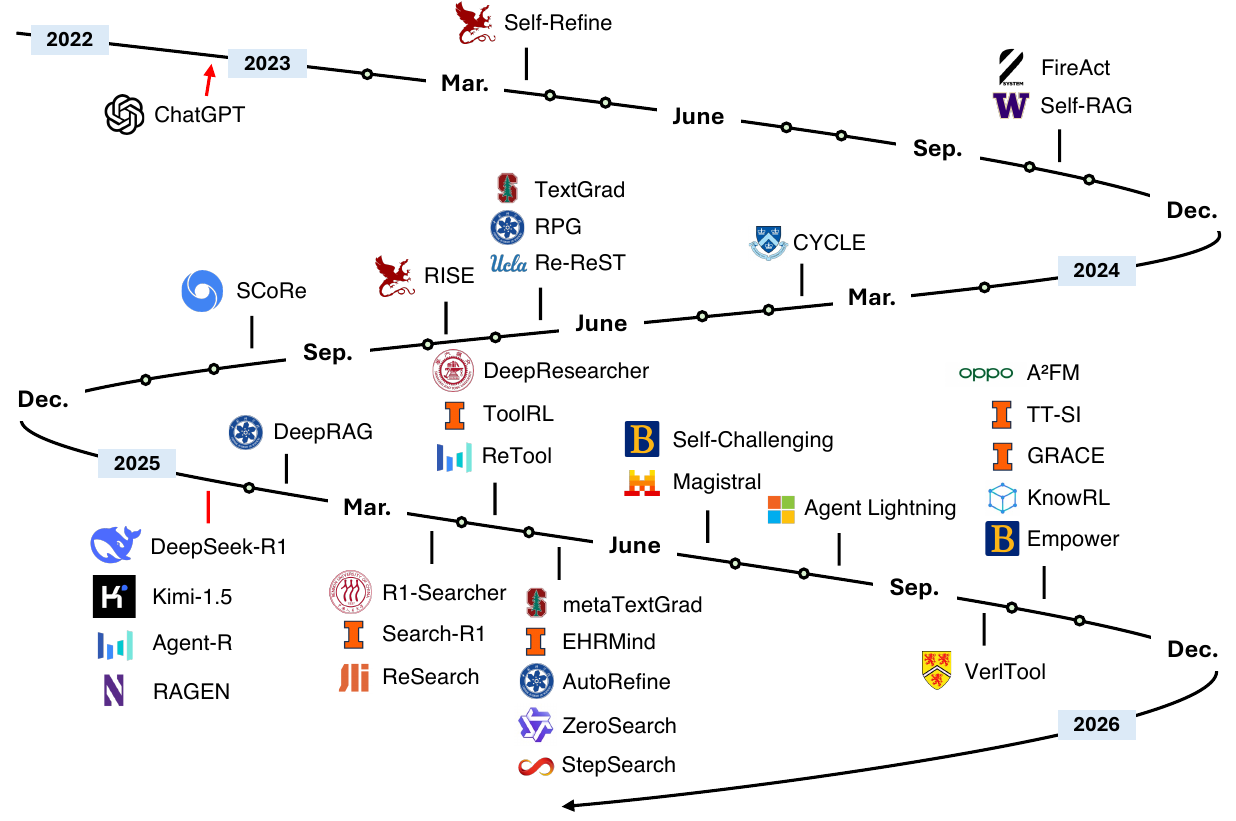}
    \caption{
    Development timeline of A2 methods (agent adaptation with agent output as signal).
    }
    \label{fig:a2_timeline}
\end{figure*}
\subsection{A2: Agent Output as Signal}
\label{subsec:agent_output_as_signal_for_agent}
Whereas A1 optimizes the mechanics of individual tool calls, A2 optimizes the agent's end-to-end output---reasoning chains, final answers, or generated artifacts---using evaluations of those outputs as the training signal.
The shift from action-level to output-level feedback brings two consequences: (i)~the agent can learn \emph{strategic} behaviors (when to invoke a tool, how deeply to search, whether to self-correct) that A1 signals cannot reward, and (ii)~credit assignment becomes harder because a single episode-level reward must be attributed across many internal decisions.

We distinguish two settings based on whether tools participate in the evaluation loop:
\begin{itemize}
    \item \textbf{Agent Adaptation w/o Tools:} The agent improves intrinsic reasoning (mathematics, coding, logical inference) by optimizing against evaluations of its generated solutions.  The design space here is defined by the \emph{granularity} of the evaluation signal (scalar correctness vs.\ structured linguistic critique) and \emph{whether weights are updated} (RL-based vs.\ inference-time refinement).
    \item \textbf{Agent Adaptation w/ Tools:} The agent's outputs are assessed in conjunction with tool interactions, so the training signal reflects both reasoning quality and tool-coordination skill.  The key additional challenge is learning the \emph{meta-policy}---when and how to invoke tools---alongside the reasoning policy.
\end{itemize}

\definecolor{sectionblue}{RGB}{230,245,255}
\definecolor{sectiongreen}{RGB}{235,255,235}
\definecolor{cardinal}{RGB}{180,0,0}

\definecolor{white}{RGB}{255,255,255}

\small
%
%
\rowcolors*{1}{gray!7}{white}
\renewcommand{\arraystretch}{1.15}

\begin{longtable}[t]{>{\raggedright\arraybackslash}p{1.2cm}
                  >{\raggedright\arraybackslash}p{2.2cm}
                  >{\raggedright\arraybackslash}p{1.5cm}
                  >{\raggedright\arraybackslash}p{2.4cm}
                  >{\raggedright\arraybackslash}p{2.3cm}
                  >{\raggedright\arraybackslash}p{2.5cm}
                  >{\raggedright\arraybackslash}p{2.1cm}
                  >{\raggedright\arraybackslash}p{0.8cm}}
\rowcolor{white}
\caption{A2 Methods: Agent Output Signaled Agent Adaptation} \\

\toprule
\rowcolor{white}
\textbf{Time} & \textbf{Method} & \textbf{Venue} & \textbf{Task(s)} & \textbf{Tool(s)} & \textbf{Agent Backbone} & \textbf{Tuning} & \textbf{Links} \\
\midrule
\endfirsthead

\rowcolor{white}
\multicolumn{8}{c}{\tablename\ \thetable\ -- Continued from previous page} \\
\toprule
\rowcolor{white}
\textbf{Time} & \textbf{Method} & \textbf{Venue} & \textbf{Task(s)} & \textbf{Tool(s)} & \textbf{Agent Backbone} & \textbf{Tuning} & \textbf{Links} \\
\midrule
\endhead

\rowcolor{white}
\multicolumn{8}{r}{\textit{Continued on next page}} \\
\endfoot

\bottomrule
\endlastfoot
\rowcolor{red!5}\multicolumn{8}{c}{\textbf{w/o Tools}} \\
\midrule
2023.03 & Self-Refine & NeurIPS'23 &Dialogue, Math, Coding &--- & GPT3.5, GPT4, CODEX & Prompt Engineering &\href{https://arxiv.org/abs/2303.17651}{\textcolor{cardinal}{\faFilePdf}} \href{https://github.com/madaan/self-refine}{\faGithub} \\ 
2024.06 & TextGrad & Nature& Code Optimization, Molecule Optimization, etc. & --- & GPT3.5, GPT4o & Prompt Engineering & \href{https://arxiv.org/abs/2406.07496}{\textcolor{cardinal}{\faFilePdf}}  \href{https://github.com/zou-group/textgrad}{\faGithub} \\ 
2024.07 & RISE &NeurIPS'24& Math &---&LLaMA2, LLaMA3, Mistral&SFT& \href{https://arxiv.org/abs/2407.18219}{\textcolor{cardinal}{\faFilePdf}} \href{https://github.com/cmu-mind/RISE}{\faGithub} \\
2024.09 & SCoRe & ICLR'25 &Math, Coding, QA &--- & Gemini1.0 Pro, Gemini1.5 Flash & REINFORCE &\href{https://arxiv.org/abs/2409.12917}{\textcolor{cardinal}{\faFilePdf}} \href{https://github.com/BY571/SCoRe}{\faGithub} \\
2025.01 & DeepSeek-R1-Zero (Math) & Nature & Math & --- & DeepSeek-V3 & GRPO & \href{https://arxiv.org/abs/2501.12948}{\textcolor{cardinal}{\faFilePdf}}  \\ 
2025.01 & Kimi k1.5 & arXiv & Math, Coding & --- & Kimi k1.5 & GRPO & \href{https://arxiv.org/abs/2501.12948}{\textcolor{cardinal}{\faFilePdf}} \href{https://github.com/MoonshotAI/Kimi-k1.5}{\faGithub} \\ 
2025.05 & EHRMind & arXiv & EHR-based Reasoning & --- &  LLaMA3 & SFT, GRPO & \href{https://arxiv.org/abs/2505.24105}{\textcolor{cardinal}{\faFilePdf}}  \\ 
2025.05 & metaTextGrad & NeurIPS'25 & QA, Math, Word Sorting & --- & Qwen3-235B-A22B, Claude-3.5-Sonnet & Prompt Engineering & \href{https://arxiv.org/abs/2505.18524}{\textcolor{cardinal}{\faFilePdf}}  \href{https://github.com/zou-group/metatextgrad}{\faGithub} \\ 
2025.06 & Magistral & arXiv & Math, Coding & --- & Magistral & PPO, GRPO & \href{https://arxiv.org/abs/2506.10910}{\textcolor{cardinal}{\faFilePdf}} \\ 
2025.10 & GRACE & arXiv & Embedding Tasks & --- & Qwen2.5, Qwen3, LLaMA3.2 & GRPO & \href{https://arxiv.org/abs/2510.04506}{\textcolor{cardinal}{\faFilePdf}} \href{https://github.com/GasolSun36/GRACE}{\faGithub} \\ 
2025.10 & KnowRL & arXiv & Knowledge Calibration & --- &LLaMA3.1, Qwen2.5 & REINFORCE++ & 
\href{https://arxiv.org/abs/2510.11407}{\textcolor{cardinal}{\faFilePdf}} \href{https://anonymous.4open.science/r/KnowRL-5BF0}{\faGithub} \\ 
2025.10 & Empower & arXiv & Coding & --- & Gemma3 & SFT & \href{https://arxiv.org/abs/2510.13709}{\textcolor{cardinal}{\faFilePdf}} \href{https://github.com/festusev/codegen_empowerment/tree/main}{\faGithub} \\  
\midrule
\rowcolor{orange!5}\multicolumn{8}{c}{\textbf{w/ Tools}} \\
\midrule
2023.10 & FireAct & arXiv & QA &Search API& GPT3.5, LLaMA2, CodeLLaMA &SFT& \href{https://arxiv.org/abs/2310.05915}{\textcolor{cardinal}{\faFilePdf}} \href{https://fireact-agent.github.io}{\faGithub}  \\
2023.10 & Self-RAG &ICLR'24& QA, Fact Verification &Retriever&LLaMA2&SFT& \href{https://arxiv.org/abs/2310.11511}{\textcolor{cardinal}{\faFilePdf}} \href{https://github.com/AkariAsai/self-rag}{\faGithub} \\
2024.06 & RPG &EMNLP'24& QA, Reasoning&Search Engine, Retriever &LLaMA2, GPT3.5&SFT& \href{https://arxiv.org/abs/2406.14979}{\textcolor{cardinal}{\faFilePdf}} \href{https://github.com/haruhi-sudo/RPG}{\faGithub} \\
2024.06 & Re-ReST &EMNLP'24& QA, VQA, Sequential Decision, Coding &Tool APIs&Various Models&DPO& \href{https://arxiv.org/abs/2406.01495}{\textcolor{cardinal}{\faFilePdf}} \href{https://github.com/PlusLabNLP/Re-ReST}{\faGithub} \\
2025.01 & Agent-R & arXiv & Various Tasks &Monte Carlo Tree Search& Qwen2.5, LLaMA3.2&SFT& \href{https://arxiv.org/abs/2501.11425}{\textcolor{cardinal}{\faFilePdf}} \href{https://github.com/ByteDance-Seed/Agent-R}{\faGithub}  \\
2025.02 & RAS & arXiv & QA & Retriever& LLaMA2, LLaMA3.2 &SFT& \href{https://arxiv.org/abs/2502.10996}{\textcolor{cardinal}{\faFilePdf}} \href{https://github.com/pat-jj/RAS}{\faGithub}  \\
2025.03 & R1-Searcher & arXiv & QA &Retriever& LLaMA3.1, Qwen2.5&REINFORCE++& \href{https://arxiv.org/abs/2503.05592}{\textcolor{cardinal}{\faFilePdf}} \href{https://github.com/RUCAIBox/R1-Searcher}{\faGithub}  \\
2025.03 & Search-R1 & COLM'25 & QA &Search Engine, Retriever& Qwen2.5 &PPO, GRPO& \href{https://arxiv.org/abs/2503.09516}{\textcolor{cardinal}{\faFilePdf}} \href{https://github.com/PeterGriffinJin/Search-R1}{\faGithub}  \\
2025.03 & ReSearch & NeurIPS'25 & QA &Search Engine, Retriever& Qwen2.5 &GRPO& \href{https://arxiv.org/abs/2503.19470}{\textcolor{cardinal}{\faFilePdf}} \href{https://github.com/Agent-RL/ReCall}{\faGithub}  \\
2025.04 & ReTool & arXiv & Math &Code  Interpreter& Qwen2.5 &PPO& \href{https://arxiv.org/abs/2504.11536}{\textcolor{cardinal}{\faFilePdf}} \href{https://github.com/ReTool-RL/ReTool}{\faGithub}  \\
2025.04 & DeepResearcher & arXiv & QA, Reasoning, Deep Research &Web Search API, Web Browser& Qwen2.5&GRPO& \href{https://arxiv.org/abs/2504.03160}{\textcolor{cardinal}{\faFilePdf}} \href{https://github.com/GAIR-NLP/DeepResearcher}{\faGithub}  \\
2025.04 & ToolRL & arXiv & Tool Calling & Tool APIs& Various Models&GRPO& \href{https://arxiv.org/abs/2504.13958}{\textcolor{cardinal}{\faFilePdf}} \href{https://github.com/qiancheng0/ToolRL}{\faGithub}  \\
2025.05 & AutoRefine & NeurIPS'25 & QA &Retriever& Qwen2.5&GRPO& \href{https://arxiv.org/abs/2505.11277}{\textcolor{cardinal}{\faFilePdf}} \href{https://github.com/syr-cn/AutoRefine}{\faGithub}  \\
2025.05 & ZeroSearch & arXiv & QA &Search Engine, Web Search& Qwen2.5, LLaMA3.2&REINFORCE, GPRO, PPO, SFT& \href{https://arxiv.org/abs/2505.04588}{\textcolor{cardinal}{\faFilePdf}} \href{https://github.com/Alibaba-NLP/ZeroSearch}{\faGithub}  \\
2025.05 & StepSearch & EMNLP'25 & QA &Search Engine, Retriever& Qwen2.5&StePPO& \href{https://arxiv.org/abs/2505.15107}{\textcolor{cardinal}{\faFilePdf}} \href{https://github.com/Zillwang/StepSearch}{\faGithub}  \\
2025.06 & Self-Challenging & arXiv &Multi-Turn Function-Calling, Calculation &Code Interpreter, Web Browser & LLaMA3.1   & REINFORCE, SFT &\href{https://arxiv.org/abs/2506.01716}{\textcolor{cardinal}{\faFilePdf}} \\
2025.06 & MMSearch-R1 & arXiv & QA, VQA &Image Search, Web Browser, Retriever& Qwen2.5&REINFORCE, SFT& \href{https://arxiv.org/abs/2506.20670}{\textcolor{cardinal}{\faFilePdf}} \href{https://github.com/EvolvingLMMs-Lab/multimodal-search-r1}{\faGithub}  \\
2025.07 & DynaSearcher & arXiv & QA &Document Search, KG Search& Qwen2.5, LLaMA3.1&GRPO& \href{https://arxiv.org/abs/2507.17365}{\textcolor{cardinal}{\faFilePdf}} \href{https://modelscope.cn/home}{\faGithub}  \\
2025.07 & CodePRM & ACL'25 &Coding &Code Executor & Qwen2.5-Coder   & SFT &\href{https://aclanthology.org/2025.findings-acl.428/}{\textcolor{cardinal}{\faFilePdf}} \\
2025.08 & Agent Lightning & arXiv & Text2SQL, Math &SQL Executor, Retriever, Calculator& LLaMA3.2 & LightningRL & \href{https://arxiv.org/abs/2508.03680}{\textcolor{cardinal}{\faFilePdf}} \href{https://github.com/microsoft/agent-lightning}{\faGithub}  \\
2025.08 & MedResearcher-R1 & arXiv & Medical QA &Medical Retriever, Web Search API, Document Reader& MedResearcher-R1&SFT, GRPO& \href{https://arxiv.org/abs/2508.14880}{\textcolor{cardinal}{\faFilePdf}} \href{https://github.com/AQ-MedAI/MedResearcher-R1}{\faGithub}  \\
2025.09 & VerlTool & arXiv & Math, QA, SQL, Visual, Web Search, Coding & Code Interpreter, Search Engine, SQL Executor, Vision Tools & Qwen2.5, Qwen3 & GRPO & \href{https://arxiv.org/abs/2509.01055}{\textcolor{cardinal}{\faFilePdf}} \href{https://github.com/TIGER-AI-Lab/verl-tool}{\faGithub} \\ 
2025.10 & A$^2$FM & arXiv & Web Navigation, Math, QA &Search Engine, Crawl, Code Executor& Qwen2.5 & APO,GRPO & \href{https://arxiv.org/abs/2510.12838}{\textcolor{cardinal}{\faFilePdf}} \href{https://github.com/OPPO-PersonalAI/Adaptive_Agent_Foundation_Models}{\faGithub}  \\
2025.10 & TT-SI & arXiv & Tool Calling &Tool APIs& Qwen2.5 & Test-Time Fine-Tuning  & \href{https://arxiv.org/abs/2510.07841}{\textcolor{cardinal}{\faFilePdf}}  \\
\end{longtable}
\normalsize

\subsubsection{Agent Adaptation w/o Tools}
\label{subsec:agent_adaptation_no_tools}

Three design axes organize the methods below: (1)~the \emph{reward granularity} ranges from binary final-answer correctness to structured natural-language critiques; (2)~the \emph{optimization locus} ranges from weight updates (RL, SFT) to inference-time refinement (no weight changes); and (3)~the \emph{objective scope} ranges from task-specific correctness to broader goals such as calibration or human empowerment.

\textbf{\underline{Scalar-reward RL (the R1 paradigm).}}
\textbf{DeepSeek-R1}~\cite{guo2025deepseek} (Nature 2025) demonstrated that RLVR with binary final-answer correctness is sufficient to unlock strong mathematical and coding reasoning in large agents, revealing a scalable pathway beyond SFT.
\textbf{Kimi-1.5}~\cite{team2025kimi} extended this to multi-modal agents with simplified policy optimization, matching or surpassing DeepSeek-R1 across reasoning benchmarks.  Together, these works established the \textbf{R1 paradigm}: RL on verifiable output evaluations as the primary driver of reasoning improvement.

Subsequent R1-style studies explore the boundaries of this paradigm by varying the objective:
\textbf{EHRMind}~\cite{lin2025training} applies RLVR to clinical reasoning (EHR interpretation), finding that a lightweight SFT warm-up is necessary before RL to ground domain knowledge---a pattern that recurs across specialized domains.
\textbf{KnowRL}~\cite{kale2025knowrl} replaces task correctness with self-knowledge calibration: the agent is rewarded for accurately assessing its own confidence, improving reliability without targeting any single task.
\textbf{Empower}~\cite{ellis2025training} goes further by defining the objective as maximizing human empowerment rather than correctness, aligning assistive agents to multi-turn coding tasks using only offline text data.
\textbf{GRACE}~\cite{sun2025grace} transforms contrastive-learning objectives into RL-style policy signals, bridging generative reasoning with representation learning.
A related study, \textbf{Rec-R1}~\cite{lin2025rec}, applies RL to product re-ranking using recommendation metrics (NDCG, Recall) as rewards.  Rec-R1 straddles the A1/A2 boundary---A1 when the reward is computed from the tool's ranking output, A2 when it evaluates the agent's overall recommendation quality---illustrating that the distinction is not always binary but depends on where in the agent-tool pipeline the evaluation signal is computed.

\textbf{\underline{Inference-time self-refinement (no weight updates).}}
A complementary line of work achieves adaptation at inference time, avoiding the cost of retraining.
\textbf{Self-Refine}~\cite{kumar2025training} (NeurIPS 2023) introduced the generate-critique-revise loop: the same LLM produces an initial response, evaluates it, and revises it based on self-generated textual feedback, improving output quality across dialogue, mathematics, and code without any supervised data or auxiliary models.
\textbf{SCoRe}~\cite{kumar2025training} (ICLR 2025) makes self-correction trainable: multi-turn on-policy RL encourages the model to iteratively refine its own responses, operationalizing the Self-Refine loop into a stable learning signal.  The progression from Self-Refine to SCoRe mirrors the SFT-to-RLVR progression in A1: inference-time heuristics are replaced by learned policies that improve with scale.

\textbf{\underline{Structured linguistic feedback.}}
\textbf{TextGrad}~\cite{yuksekgonul2025optimizing} (Nature 2025) generalizes the scalar reward to natural-language critiques that describe \emph{how} to improve the output. These ``textual gradients'' enable optimization across black-box LLM systems without parameter access, improving GPT-4o's zero-shot code accuracy on \textsc{LeetCode-Hard} from 26\% to 36\% and \textsc{MMLU}-Physics from 91.2\% to 95.1\%.
\textbf{metaTextGrad}~\cite{xu2025metatextgrad} (NeurIPS 2025) applies the same principle recursively to the optimizer itself, using validation feedback to refine the optimizer's prompts.  TextGrad and metaTextGrad occupy a unique position: unlike RL methods, they require no reward function design; unlike Self-Refine, they propagate structured credit through multi-component systems.

\paragraph{Synthesis of A2 without tools.}
The methods above map onto a two-dimensional design space.  Along the first axis (reward granularity), scalar final-answer correctness (DeepSeek-R1) is broadly applicable but sparse, while structured linguistic feedback (TextGrad) is information-rich but requires a capable critic model.  Along the second axis (optimization locus), weight-update methods (RL, SFT) amortize adaptation cost across future queries, while inference-time methods (Self-Refine, TextGrad) pay per-query cost but require no training infrastructure.  A recurring finding across both axes is that SFT warm-up followed by RL yields more stable training than RL alone: SFT provides domain grounding, while RL refines the policy toward the evaluation objective.

\subsubsection{Agent Adaptation w/ Tools}
\label{subsec:agent_adaptation_with_tools}
When tools enter the A2 loop, the agent must learn a \emph{meta-policy}: not only how to reason but when to invoke a tool, which tool to choose, and how to integrate tool outputs back into the reasoning chain.  The reward is still computed on the agent's final output, but the policy space is much larger, making credit assignment and data efficiency the central challenges.

\textbf{\underline{Retrieval-based tool learning.}}
Prior distillation-SFT approaches such as \textbf{Self-RAG}~\cite{asai2024self} (ICLR 2024) and its successors~\cite{lyu2024retrieve, jiang2025ras, guan2025deeprag} teach retrieval tool use from expert demonstrations; RL-based methods instead let the agent discover search strategies on its own.
\textbf{R1-Searcher}~\cite{song2025r1}, \textbf{Search-R1}~\cite{jin2025search} (COLM 2025), \textbf{ReSearch}~\cite{chen2025learning} (NeurIPS 2025), and their successors~\cite{zheng2025deepresearcher, sun2025zerosearch, wang2025stepsearch, hao2025dynasearcher, yu2025medresearcher, shi2025search,wu2025mmsearch} all train LLMs to autonomously generate and refine search queries during multi-turn reasoning.
The distinguishing design choice across this family is the \emph{action representation}: R1-Searcher incentivizes search API calls via a two-stage RL framework (up to 24\% over RAG baselines), Search-R1 formulates search invocation as a joint reward over retrieved evidence and final correctness, and ReSearch integrates queries and results directly into the reasoning chain via \texttt{<think>}/\texttt{<search>}/\texttt{<result>} tags, optimized with GRPO.  ReSearch is notable for exhibiting emergent reflection and self-correction behaviors---strategic capabilities that were not explicitly supervised but arose from holistic task optimization.

\textbf{\underline{Code- and execution-based tool learning.}}
Code execution provides a natural bridge between A1 and A2: the execution outcome is verifiable (A1-like), but the training signal evaluates the agent's overall problem-solving trajectory (A2-like).
\textbf{CodePRM}~\cite{li2025codeprm} (ACL 2025) scores intermediate reasoning steps via a process reward model trained on code execution results, forming a Generate-Verify-Refine pipeline that corrects errors during inference.
\textbf{ReTool}~\cite{feng2025retool} integrates real-time code execution directly into RL rollouts, so the agent learns when to invoke an interpreter to offload computation---illustrating how A2 training can discover tool-use strategies that A1 training cannot, because the reward depends on whether the agent's overall answer (not just the tool call) is correct.

\textbf{\underline{General multi-tool and agentic learning.}}
In the most general setting, agents interact with heterogeneous APIs and environments.  Three recurring themes organize the methods below: \emph{data generation}, \emph{self-reflection}, and \emph{infrastructure}.

On the data side, \textbf{Self-Challenging Agents}~\cite{zhou2025self} introduce a self-generated curriculum---the model first creates tool-use tasks as a challenger, then solves them via RL as an executor---achieving over twofold improvement on multi-turn benchmarks.
\textbf{Re-ReST}~\cite{dou2024re} uses environment feedback (e.g., unit-test results) to refine low-quality trajectories through reflection, yielding large gains on HotpotQA and AlfWorld.
On the self-reflection side, \textbf{Agent-R}~\cite{yuan2025agent} formalizes iterative self-correction via model-guided critique and MCTS rollouts (+5.6\% across interactive environments), and \textbf{Test-Time Self-Improvement}~\cite{acikgoz2025self} performs on-the-fly fine-tuning at inference time on self-identified uncertain cases.
On the infrastructure side, \textbf{Agent Lightning}~\cite{luo2025agent} decouples agent execution from training to support complex multi-agent workflows, \textbf{A$^{2}$FM}~\cite{chen20252} unifies reasoning and acting under cost-regularized RL (dynamically choosing between thinking, tool use, and direct answering), and \textbf{VerlTool}~\cite{jiang2025verltool} provides asynchronous rollout infrastructure that eliminates synchronization bottlenecks in agentic RL.

\paragraph{Synthesis of A2 with tools.}
A2 methods with tools share a defining characteristic: the agent must learn not only \emph{how} to use tools (also targeted by A1) but \emph{when} to use them and how to fold their outputs into a coherent reasoning chain.  This strategic dimension explains why A2 training frequently produces emergent behaviors---self-correction, adaptive search depth, query refinement---that are never explicitly supervised.  The cost is data efficiency: the sparse, episode-level reward requires either large training sets (e.g., Search-R1's $\sim$170k examples) or carefully designed curricula (e.g., Self-Challenging Agents' self-generated tasks).  The tension between A1's per-action precision and A2's end-to-end strategic optimization is analyzed quantitatively in \S\ref{sec:comparison}.

\section{Tool Adaptation}
\label{sec:tool_adaptation}
Tool adaptation shifts the optimization target from the agent itself to its ecosystem. Instead of modifying the agent's parameters through fine-tuning or reinforcement learning, this paradigm targets the external components (pre-trained models, retrievers, planners, or executors) that the agent invokes through language or code. Methods in this category (1) employ pre-trained machine learning models as plug-and-play components, ranging from simple classifiers to complex LLM-based subagents as discussed in \S\ref{sec:agent_adaptation}; or (2) use the agent's outputs as supervision or reinforcement signals to train, align, or refine the tool itself.

Formally, let $\mathcal{A}$ denote an agent, parameterized by its internal configuration or policy (which include prompt templates or model weights) and $\mathcal{T}$ denote a tool or a set of tools that can be trained or optimized based on task feedback. 
The adaptation process can be characterized by two complementary paradigms:

\[
\textbf{(T1)} \quad \mathcal{T}^{*} = \arg\max_{\mathcal{T}} \mathcal{O}_{\text{tool}}(\mathcal{T}), 
\qquad 
\textbf{(T2)} \quad \mathcal{T}^{*} = \arg\max_{\mathcal{T}} \mathcal{O}_{\text{agent}}(\mathcal{A}, \mathcal{T}),
\]

where $\mathcal{O}_{\text{tool}}$ quantifies task-specific or environment-driven improvements that are independent of the agent, such as retrieval accuracy or planning efficiency, while $\mathcal{O}_{\text{agent}}$ incorporates agent-derived supervision, where the agent’s outputs provide learning signals to refine or align the tool. 
Here, $\mathcal{T}^{*}$ denotes the optimized tool configuration that maximizes the respective objective, illustrating how tool adaptation complements agent-level optimization within the broader agent-tool ecosystem.

\subsection{T1: Agent-Agnostic Tool Adaptation}
\label{subsec:agent_agnostic_tool_training}

The foundational architecture for tool-augmented systems uses pre-trained models as plug-and-play tools for frozen agents. The agent orchestrates tool usage through prompting alone, never updating its parameters, while relying on tools trained independently on diverse data sources before deployment.

While T1 is the most straightforward paradigm (any pre-trained model an agent can invoke qualifies), it provides the compositional substrate on which all other paradigms build: A1/A2 agents invoke T1 tools, T2 subagents are often initialized from T1 components, and the ``graduation lifecycle'' (A1$\rightarrow$T1, discussed in \S\ref{sec:comparison_t1}) continually enriches the T1 ecosystem. Understanding T1's design space is therefore essential for the entire framework.

\subsubsection{Foundational Systems and Architectures}
\label{subsubsec:4.1.1}

Early systems established the architectural foundations for how frozen agents orchestrate external models. These works illustrate distinct mechanisms (functional, prompt-based, code-based, and graph-based) that shaped the design space for modern tool-augmented AI systems.

\textbf{Operator-Learning Tools: Neural Operators}~\cite{kovachki2023neural} (JMLR).
Before large-scale LLM-based orchestration emerged, Neural Operators represented an early example of \textit{agent-agnostic tool learning}: models trained to approximate mappings between infinite-dimensional function spaces, serving as differentiable surrogates for complex simulators.
Unlike conventional neural networks tied to discrete grids, Neural Operators are \textbf{discretization-invariant}: they learn the underlying operator itself, not its finite discretization, and can generalize across resolutions and geometries.
The \textbf{Fourier Neural Operator (FNO)} achieves $\mathcal{O}(J\log J)$ inference via spectral convolution and outperforms classical solvers on Navier-Stokes, Darcy flow, and elasticity equations by orders of magnitude in speed.
FNO and its variants (Graph-, Low-rank-, Multipole-NO) represent the first wave of ``frozen tools'' that agents can query repeatedly for reasoning, planning, or control without retraining.
In modern agentic pipelines, they serve as plug-in surrogates: fast, differentiable black-box functions invoked within decision or inference loops.

\textbf{HuggingGPT}~\cite{shen2023hugginggpt} (NeurIPS 2023) introduced the orchestration paradigm by allowing ChatGPT to command 1000+ machine learning models from HuggingFace Hub without any fine-tuning. The frozen LLM executes a four-stage workflow: task planning (decomposing user requests), model selection (choosing from tool descriptions), task execution (invoking models), and response generation (synthesizing outputs). The architecture shows that tool descriptions in natural language suffice for the frozen agent to coordinate complex multimodal workflows. On composite cross-modal tasks, HuggingGPT demonstrates that orchestrating specialized vision, speech, and language models can close the gap between GPT-3.5 and much larger multimodal models. The primary limitation lies in latency from sequential LLM calls and token length constraints for tool descriptions.
\textbf{ViperGPT}~\cite{suris2023vipergpt} (ICCV 2023) introduced code generation as the orchestration mechanism. The frozen GPT-3 Codex generates Python code that composes vision models (GLIP for detection, SAM for segmentation, MiDaS for depth estimation) into executable programs. The code-based approach achieves strong zero-shot performance on GQA visual reasoning, outperforming end-to-end models by 10--15\% on compositional tasks. Python functions provide more flexible tool composition than fixed API calls. Each tool exposes simple functions like \texttt{find(image, object\_name)} or \texttt{compute\_depth(image)}, which Codex chains programmatically without learning tool-specific interfaces.
\textbf{SciToolAgent}~\cite{ding2025scitoolagent} (Nature Computational Science 2025) scales tool orchestration to scientific domains through graph-based organization. The frozen GPT-4o accesses 500+ biology, chemistry, and materials science tools via SciToolKG, a knowledge graph encoding tool metadata, dependencies, and safety constraints. Graph-based retrieval for tool selection achieves 94\% accuracy on scientific query benchmarks, representing a 15--20\% improvement over GPT-4o without tool access. The system successfully automates protein engineering workflows chaining ESMFold for structure prediction, BLAST for sequence alignment, and custom analysis tools. Structured knowledge graphs address the scalability challenges inherent in prompt-based descriptions.

These foundational systems illustrate the dominant integration patterns. HuggingGPT exemplifies \textbf{prompt-based orchestration}, where the agent parses tool calls from text. ViperGPT uses \textbf{code generation}, exposing tools as Python functions. SciToolAgent uses \textbf{knowledge graph retrieval}, using RAG to select from structured tool graphs. A fourth common pattern is \textbf{multimodal bridging}, which converts non-textual modalities into text representations; for example, Visual ChatGPT's~\cite{wu2023visual} prompt manager serializes vision operations as text API calls. The usability of these patterns depends on clear \textbf{interface design}, such as programmatic function signatures (e.g., \texttt{find\_object(image: PIL.Image, ...)}), structured JSON schemas, or simple natural language descriptions. \textbf{Execution modes} are similarly varied, ranging from direct API calls and code generation to HTTP requests and command-line invocations.

\textbf{Model Context Protocol (MCP) and Code Execution Environments}~\cite{anthropic2025mcp}. 
As large-scale agent ecosystems began connecting thousands of heterogeneous tools, the \textit{Model Context Protocol} (MCP) emerged as an open standard for unifying how agents interface with external systems. 
MCP provides a universal API layer that allows frozen agents to discover, invoke, and coordinate tools across domains using a consistent schema, rather than embedding long tool definitions directly into the model's context. 
Anthropic's ``\textit{Code Execution with MCP}'' paradigm introduced an execution-centric design in which the agent writes executable code to interact with MCP servers instead of performing token-level tool calls. 
Agents can thereby load only the necessary tool definitions, filter or aggregate data within a sandboxed environment, and pass compact results back to the model, substantially reducing context usage while preserving compositionality. 
MCP represents a scalable \textbf{T1-style tool adaptation infrastructure} that decouples execution from inference, while the code-execution mode bridges toward \textbf{T2-style optimization} by dynamically improving efficiency under frozen agents.

\subsubsection{Categories and Training Methodologies}
\label{subsubsec:4.1.2}

While the preceding systems define \emph{how} agents invoke tools, the choice of \emph{which} tool modalities to deploy shapes the entire adaptation surface.

Tool adaptation encompasses a range of pre-trained model categories. We organize these by modality and highlight how their training methodology determines their amenability to T2 optimization: models producing structured, evaluable outputs (code, retrieval results) are most naturally upgraded to T2, because the frozen agent can score their outputs automatically; models producing perceptual features (vision, speech) are harder to supervise via T2 and remain predominantly T1 plug-ins.

\textbf{Vision models} dominate T1 deployments as plug-and-play tools. CLIP~\cite{radford2021learning}, trained contrastively on 400M image-text pairs, provides zero-shot classification and semantic understanding through frozen encoders. SAM~\cite{kirillov2023segment}, trained on 11M images with 1B masks via human-in-the-loop data engines, provides promptable segmentation with point, box, or mask inputs. SAM-CLIP~\cite{wang2024sam} merges these capabilities through multi-task distillation with frozen teachers, achieving +6.8\% mIoU on Pascal VOC for zero-shot semantic segmentation while retaining both parent models' strengths. These vision tools require no task-specific fine-tuning - frozen agents invoke them directly via APIs for image understanding, segmentation, and classification tasks.

\textbf{Speech and audio tools} rely on massive pre-training for robust performance. Whisper~\cite{radford2023robust}, trained on 680K hours of multilingual audio with weak supervision, provides speech recognition, translation, and language identification as a frozen API for multimodal agents. The encoder-decoder Transformer architecture supports zero-shot transcription across languages and domains, with strong robustness to accents, noise, and technical terminology. Agents simply pass audio inputs to the frozen Whisper model and process text outputs without any model adaptation.

\textbf{Code execution tools} encompass models that learn to compose and execute functions through code. CodeAct~\cite{wang2024executable} shows that representing tool use in executable Python rather than static JSON improves compositional reasoning, achieving over 20\% higher success rates on API-Bank benchmarks. The dynamic nature of code allows agents to flexibly construct, parameterize, and combine tools without predefined schemas.

\textbf{Search and retrieval tools} comprise pre-trained dense retrievers such as DPR~\cite{karpukhin2020dense}, ColBERT~\cite{khattab2020colbert}, Contriever~\cite{izacard2021unsupervised}, and e5~\cite{wang2022text}, often deployed as frozen components within retrieval-augmented generation pipelines. These bi-encoder models, trained on passage ranking tasks, support semantic search over large corpora.

\textbf{Scientific tools} extend capabilities to specialized fields. AlphaFold2~\cite{jumper2021highly} and ESMFold~\cite{lin2023evolutionary} provide protein structure prediction from sequences. Materials science models like CGCNN~\cite{xie2018crystal} predict crystal properties. Molecule representation learning approaches~\cite{xu2024smiles,jin2017predicting,jin2018learning,zheng2019predicting,rong2020selfsupervised,fang2023knowledge,jiang2025bilevel} have been developed for molecular property prediction, whereas some encoder-decoder frameworks~\cite{qi2024predicting,tong2024deep} aim to predict transcriptional profiles elicited by chemical perturbations. These tools represent years of domain-specific model development, deployed as-is for frozen agents tackling scientific queries.

Beyond these static models, adaptive agents introduced in \S\ref{sec:agent_adaptation} (such as DeepRetrieval~\cite{jiang2025deepretrieval} for search query rewriting and Code-R1~\cite{code-r1} for code generation) illustrate how trained reasoning agents themselves can function as dynamic tools. Once frozen, they extend the tool ecosystem by reformulating queries, generating executable code, or performing reasoning-driven actions, thereby bridging the gap between pre-trained models and the environment or offline data.

\subsection{T2: Agent-Supervised Tool Adaptation}
\label{subsec:agent_output_as_signal_for_tool}

The T2 paradigm inverts the standard adaptation question: rather than asking ``how can we modify the agent to better use its tools?'' (the A1/A2 question), T2 asks ``how can we modify the tools to better serve a fixed agent?'' This reframes the expensive, monolithic foundation model as a stable source of supervision rather than the target of optimization.

Training or fine-tuning billion-parameter foundation models is computationally expensive and risks catastrophic forgetting. Peripheral tools (retrievers, planners, memory modules) are orders of magnitude smaller and cheaper to train. T2 exploits this asymmetry: the frozen agent provides supervision signals derived from its pre-trained knowledge, while the tools learn to reshape information into the form the agent can best exploit.

The evolution of T2 methods from 2023 to 2025 shows a progression from using internal proxy signals (perplexity, preferences) to train passive retrieval tools, to using verifiable outcome signals (task success, accuracy gains) to train active, multi-turn agentic tools.

\begin{figure*}[t]
    \centering
    \includegraphics[width=\textwidth]{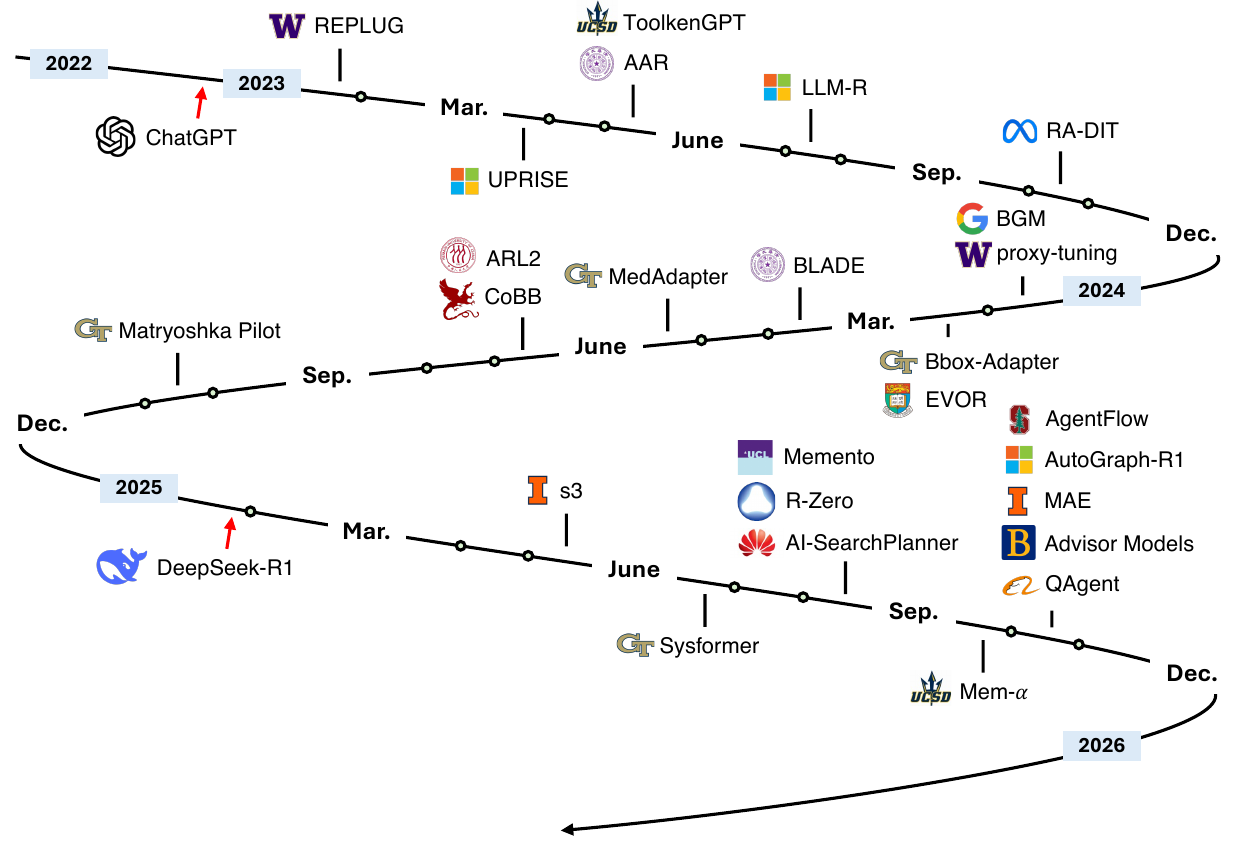}
    \caption{
    Development timeline of T2 methods (agent-supervised tool adaptation, classic memory-related methods are not included in this figure due to space limitation).
    }
    \label{fig:t2_timeline}
\end{figure*}
\definecolor{sectionblue}{RGB}{230,245,255}
\definecolor{sectiongreen}{RGB}{235,255,235}
\definecolor{cardinal}{RGB}{180,0,0}

\definecolor{white}{RGB}{255,255,255}

\small
%
%
\rowcolors*{1}{gray!7}{white}
\renewcommand{\arraystretch}{1.15}

\begin{longtable}[t]{>{\raggedright\arraybackslash}p{1.2cm}
                  >{\raggedright\arraybackslash}p{2.0cm}
                  >{\raggedright\arraybackslash}p{1.5cm}
                  >{\raggedright\arraybackslash}p{2.4cm}
                  >{\raggedright\arraybackslash}p{2.5cm}
                  >{\raggedright\arraybackslash}p{2.5cm}
                  >{\raggedright\arraybackslash}p{1.8cm}
                  >{\raggedright\arraybackslash}p{0.8cm}}
\rowcolor{white}
\caption{T2 Methods: Tool Adaptation w/ Agent Supervision} \\

\toprule
\rowcolor{white}
\textbf{Time} & \textbf{Method} & \textbf{Venue} & \textbf{Task(s)} & \textbf{Tool Backbone} & \textbf{Agent Backbone} & \textbf{Tuning} & \textbf{Links} \\
\midrule
\endfirsthead

\rowcolor{white}
\multicolumn{8}{c}{\tablename\ \thetable\ -- Continued from previous page} \\
\toprule
\rowcolor{white}
\textbf{Time} & \textbf{Method} & \textbf{Venue} & \textbf{Task(s)} & \textbf{Tool Backbone} & \textbf{Agent Backbone} & \textbf{Tuning} & \textbf{Links} \\

\midrule
\endhead

\rowcolor{white}
\multicolumn{8}{r}{\textit{Continued on next page}} \\
\endfoot

\bottomrule
\endlastfoot
\rowcolor{sectionblue}\multicolumn{8}{c}{\textbf{Earlier Methods}} \\
\midrule
2023.01 & REPLUG & NAACL'24 & QA & Contriever & GPT3-175B, PaLM, Codex, LLaMA-13B   &  Proxy-Tuning, LSR & \href{https://aclanthology.org/2024.naacl-long.463.pdf}{\textcolor{cardinal}{\faFilePdf}} \href{https://github.com/swj0419/REPLUG}{\faGithub} \\
2023.03 & UPRISE  & EMNLP'23 & Zero-shot NLU (QA, NLI, etc.) &GPT-Neo-2.7B &BLOOM-7.1B, OPT-66B, GPT-3-175B& Contrastive Learning& \href{https://aclanthology.org/2023.emnlp-main.758/}{\textcolor{cardinal}{\faFilePdf}} \href{https://github.com/microsoft/LMOps}{\faGithub}\\
2023.05 & ToolkenGPT &	NeurIPS'23 &Numerical Reasoning, QA, Plan Generation & Token Embedding &GPT-J 6B, OPT-6.7B, OPT-13B & Proxy-Tuning & \href{https://arxiv.org/abs/2305.11554}{\textcolor{cardinal}{\faFilePdf}} \href{https://github.com/Ber666/ToolkenGPT}{\faGithub}\\
2023.05 & AAR & ACL'23  & Zero-Shot Generalization (MMLU, PopQA) & ANCE, Contriever  & Flan-T5-Small, InstructGPT & Contrastive Learning & \href{https://arxiv.org/abs/2305.17331}{\textcolor{cardinal}{\faFilePdf}} \href{https://github.com/OpenMatch/Augmentation-Adapted-Retriever}{\faGithub}\\
2023.06 & LLM-R & EACL'24  &  Zero-shot NLU (QA, NLI, Paraphrase, etc.) & E5-base  & GPT-Neo-2.7B, LLaMA-13B, GPT-3.5-Turbo & Contrastive Learning & \href{https://arxiv.org/abs/2307.07164}{\textcolor{cardinal}{\faFilePdf}} \href{https://github.com/microsoft/LMOps/tree/main/llm_retriever}{\faGithub}\\
2023.10 & RA-DIT & ICLR'24  & Knowledge-Intensive Tasks (MMLU, NQ, TQA, ELI5, HotpotQA, etc.) & DRAGON+  & LLaMA-65B & SFT, LSR & \href{https://arxiv.org/abs/2310.01352}{\textcolor{cardinal}{\faFilePdf}} \\
2024.01 & BGM & ACL'24  & QA & T5-XXL-11B  & PaLM2-S  & SFT, PPO & \href{https://arxiv.org/abs/2401.06954}{\textcolor{cardinal}{\faFilePdf}} \\
2024.01 & Proxy-Tuning & COLM'24 & QA, Math, Code & LLaMA2-7B & LLaMA2-70B & Proxy-Tuning & \href{https://arxiv.org/abs/2401.08565}{\textcolor{cardinal}{\faFilePdf}} \href{https://github.com/alisawuffles/proxy-tuning}{\faGithub}\\
2024.02 & Bbox-Adapter & ICML'24 & QA & DeBERTa-v3-base (0.1B), DeBERTa-v3-large (0.3B) &  GPT-3.5-Turbo, Mixtral-8x7B &Contrastive Learning & \href{https://arxiv.org/abs/2402.08219}{\textcolor{cardinal}{\faFilePdf}} \href{https://github.com/haotiansun14/BBox-Adapter}{\faGithub}\\
2024.02 & EVOR & EMNLP'24 &  Coding & GPT-3.5-Turbo &  GPT-3.5-Turbo, CodeLLaMA &Prompt Engineering & \href{https://arxiv.org/abs/2402.12317}{\textcolor{cardinal}{\faFilePdf}} \href{https://github.com/xlang-ai/EVOR}{\faGithub}\\
2024.02 & ARL2 & ACL'24 & QA & LLaMA2-7B  &  GPT-3.5-Turbo &Contrastive Learning & \href{https://arxiv.org/abs/2402.13542}{\textcolor{cardinal}{\faFilePdf}} \href{https://github.com/zhanglingxi-cs/ARL2}{\faGithub}\\
2024.03 & BLADE & AAAI'25 & QA &  BLOOMZ-1b7 &  ChatGPT, ChatGLM, Baichuan, Qwen &SFT, BPO & \href{https://arxiv.org/abs/2403.18365}{\textcolor{cardinal}{\faFilePdf}} \href{https://github.com/CSHaitao/BLADE}{\faGithub}\\
2024.05 & Medadapter & EMNLP'24 & Medical QA, NLI, RQE & BERT-Base-Uncased  &  GPT-3.5-Turbo &SFT, BPO & \href{https://arxiv.org/abs/2405.03000}{\textcolor{cardinal}{\faFilePdf}} \href{https://github.com/wshi83/MedAdapter}{\faGithub}\\
2024.06 & CoBB & EMNLP'24 & QA, Math & Mistral-7b-inst-v2  &  GPT-3.5-Turbo, Claude-3-Haiku, Phi-3-mini-4k-inst, Gemma-1.1-7B-it, Mistral-7B-inst-v2 &SFT, ORPO & \href{https://arxiv.org/abs/2406.18695}{\textcolor{cardinal}{\faFilePdf}} \href{https://github.com/bbuing9/CoBB}{\faGithub}\\
2024.10 & Matryoshka Pilot & NeurIPS'25 & Math, Planning, Reasoning & LLaMA3-8B, Qwen2.5-7B  &  GPT-4o-Mini, GPT-3.5-Turbo &DPO, IDPO & \href{https://arxiv.org/abs/2410.20749}{\textcolor{cardinal}{\faFilePdf}} \href{https://github.com/lichangh20/Matryoshka}{\faGithub}\\
2025.06 & Sysformer & arXiv & QA & Small Transformer  &  LLaMA-2-7B, LLaMA-3.1-8B, Mistral-7B, Phi-3.5-mini, Zephyr-7B-beta &Supervised Learning & \href{https://arxiv.org/abs/2506.15751}{\textcolor{cardinal}{\faFilePdf}} \\
\midrule
\rowcolor{sectiongreen}\multicolumn{8}{c}{\textbf{RLVR Methods}} \\
\midrule
2025.05 & s3 & EMNLP'25 & QA & Qwen2.5-7B  &  Qwen2.5-7B, Qwen2.5-14B, Claude-3-Haiku &PPO & \href{https://arxiv.org/abs/2505.14146}{\textcolor{cardinal}{\faFilePdf}} \href{https://github.com/pat-jj/s3}{\faGithub}\\
2025.08 & R-Zero & arXiv & Math, Reasoning & Qwen3-4B, Qwen3-8B, OctoThinker-3B, OctoThinker-8B  &  Qwen3-4B, Qwen3-8B, OctoThinker-3B, OctoThinker-8B &GRPO & \href{https://arxiv.org/abs/2508.05004}{\textcolor{cardinal}{\faFilePdf}} \href{https://github.com/Chengsong-Huang/R-Zero}{\faGithub}\\
2025.08 & Memento & arXiv & Long-Horizon Reasoning, Web Research, QA, Academic Reasoning & Q-function (two-layer MLPs) & GPT-4.1  &  Soft Q-Learning & \href{https://arxiv.org/abs/2508.16153}{\textcolor{cardinal}{\faFilePdf}} \href{https://github.com/Agent-on-the-Fly/Memento}{\faGithub}\\
2025.08 & AI-SearchPlanner & arXiv & QA & Qwen3-32B & Qwen2.5-7B &  PPO & \href{https://arxiv.org/abs/2508.20368}{\textcolor{cardinal}{\faFilePdf}} \\
2025.09 & Mem-$\alpha$ & arXiv &  Test-Time Learning, Long-Range Understanding & Qwen3-4B & Qwen3-4B, Qwen3-32B, GPT-4.1-Mini  &  GRPO & \href{https://arxiv.org/abs/2509.25911}{\textcolor{cardinal}{\faFilePdf}} \\
2025.10 & AgentFlow & arXiv &  Web Search, Planning, Reasoning, Math & Qwen2.5-7B & Qwen2.5-7B  &  Flow-GRPO & \href{https://arxiv.org/abs/2510.05592}{\textcolor{cardinal}{\faFilePdf}}\\
2025.10	& AutoGraph-R1	&arXiv	&KG Construction &KG Constructor (Qwen2.5-3B/7B)	&Frozen RAG Generator (Qwen2.5-7B)	&GRPO  & \href{https://arxiv.org/abs/2510.15339}{\textcolor{cardinal}{\faFilePdf}}\\
2025.10 & MAE & arXiv &  Math, Coding, QA & Qwen2.5-3B & Qwen2.5-3B  &  REINFORCE++ & \href{https://arxiv.org/abs/2510.23595}{\textcolor{cardinal}{\faFilePdf}}\\
2025.10 & Advisor Models& arXiv &  Math, Reasoning & Qwen2.5-7B, Qwen3-8B & GPT-4o-Mini, GPT-5, Claude4-Sonnet, GPT-4.1-Mini  &  GRPO & \href{https://arxiv.org/abs/2510.02453}{\textcolor{cardinal}{\faFilePdf}}\\
2025.10 & QAgent& arXiv &  QA & Qwen2.5-3B & Qwen-7B &  GRPO & \href{https://arxiv.org/abs/2510.08383}{\textcolor{cardinal}{\faFilePdf}}\\
\end{longtable}
\normalsize

\subsubsection{Earlier Methods: From Proxy Signals to Structured Preferences}
\label{subsubsec:4.2.1}

The earliest T2 methods emerged from the RAG community, where researchers sought to optimize dense retrievers for compatibility with large language models. These works established the principle that a frozen LM's internal computations could serve as supervision, but they also reveal the limitations of relying on proxy metrics that may not align with downstream task objectives.

\textbf{REPLUG}~\cite{shi2024replug} (NAACL 2024) introduced a general framework for adapting frozen language models through black-box supervision, using perplexity reduction as a training signal for the retriever. If conditioning the LM on a retrieved document lowers its perplexity for a given query, the document likely provides informative context. Formally, the retriever is optimized to align its retrieval distribution with the distribution induced by the LM’s perplexity-based preferences:
\begin{equation}
\mathcal{L}_{\text{REPLUG}} = D_{\text{KL}}\!\left( P_{\text{retriever}}(d|q) \,\|\, P_{\text{LM-perplexity}}(d|q) \right), \nonumber
\end{equation}
where $P_{\text{LM-perplexity}}(d|q)$ reflects how strongly each document reduces the LM’s perplexity when conditioned on query $q$. 
REPLUG thus enabled retrieval adaptation without parameter access to the LM, establishing a family of agent-supervised methods that optimize external modules based solely on frozen-agent feedback. 
\textbf{BLADE}~\cite{li2025blade} (AAAI 2025) further extended this paradigm by replacing traditional retrievers with domain-specific models that synthesize auxiliary knowledge: it couples a frozen general LLM with a small, domain-specific LM optimized via Bayesian Prompted Optimization (BPO). 
The small LM learns to generate domain-relevant knowledge and soft prompts that improve the black-box LLM’s responses, extending REPLUG’s black-box adaptation principle from retrieval to generative, domain-specialized tool co-adaptation. \textbf{BBox-Adapter}~\cite{sun2024bbox} (ICML 2024) reframed adaptation as an energy-based modeling problem, introducing a ranking-based noise-contrastive estimation loss and an online update framework to align outputs from black-box APIs like GPT-3.5 without access to internal probabilities. These methods collectively shifted the focus from likelihood-based alignment to utility-driven adaptation, anticipating the reinforcement-learning-based search and reasoning frameworks that followed. \textbf{proxy-tuning}~\cite{liu2024tuning} (COLM 2024) extends black-box, agent-supervised adaptation to decoding time: a small tuned “expert” and its untuned “anti-expert” provide logit offsets that steer a frozen large LM without modifying its weights, effectively training a lightweight steering \emph{tool} under a frozen agent (T2).
\textbf{EVOR}~\cite{su2024evor} (EMNLP 2024) further extends to the domain of code generation, formulating retrieval and knowledge evolution as a co-adaptive process driven by execution feedback from a frozen LLM. 

\paragraph{Preference Learning: Toward Task Alignment.}
While black-box adaptation methods such as REPLUG and BBox-Adapter relied on indirect proxy signals like perplexity or ranking scores, subsequent studies~\cite{yu2023augmentation,lin2023ra,shi2024medadapter,kim2024learning,zhang2024arl2} moved toward explicit preference-based supervision that better reflects task utility. 
\textbf{AAR}~\cite{yu2023augmentation} (ACL 2023) introduced augmentation-aware retrieval, where a frozen source LM (Flan-T5-250M) constructs preference pairs by comparing documents that most improve its own likelihood against human-annotated references. 
The retriever is then trained to reproduce these preferences via a contrastive loss, yielding a signal that directly encodes the LM’s notion of ``helpful context.'' 
These LM-derived preferences transfer effectively across architectures and scales, improving even 175B-parameter models. 
\textbf{RA-DIT}~\cite{lin2023ra} (ICLR 2023) formalized this idea by defining document utility as the log-probability gain for producing correct answers:
\begin{equation}
\text{Utility}(d, q, a) = \log P_{\text{LM}}(a|q,d) - \log P_{\text{LM}}(a|q), \nonumber
\end{equation}
training retrievers to prefer documents that yield higher expected gains. 
The preference-based approach aligns retrieval more directly with downstream reasoning objectives while still operating through a single forward pass of the frozen LM. 
Together, these works mark a conceptual shift from proxy-based to task-aligned supervision, setting the stage for reinforcement-style feedback and multi-turn optimization in later frameworks.

\paragraph{Multi-stage architectures: distilling complex preferences.}

\textbf{LLM-R}~\cite{wang2024learning} (EACL 2024) introduced a multi-stage distillation pipeline that increases the fidelity of training signals. Rather than training the retriever directly on the frozen LM's outputs, LLM-R first trains an intermediate cross-encoder ``reward model'' to capture the frozen LM's nuanced preferences over in-context examples. The reward model, which can afford to be slow and expressive because it is used only during training, is then distilled into a fast bi-encoder retriever. The key insight is that the complexity of the training signal need not be constrained by inference-time efficiency requirements. By decoupling the preference modeling from the final retrieval tool, LLM-R achieves both high-quality supervision and fast deployment.

\textbf{UPRISE}~\cite{cheng2023uprise} (EMNLP 2023) extends this paradigm beyond documents to prompts, training a prompt retriever using the frozen LLM's task performance across diverse tasks. By training on multiple tasks simultaneously, UPRISE learns a generalizable meta-skill: selecting prompts that improve LLM performance in zero-shot settings. The cross-task transfer (+8.5\% on reading comprehension, +14.6\% on paraphrase detection) suggests that T2-trained tools can internalize abstract principles of ``what helps an LM'' rather than memorizing task-specific heuristics.

By 2024, a consensus emerged that optimizing retrieval in isolation is insufficient. Even a retriever that scores well on traditional IR metrics (NDCG, MRR) may produce results poorly suited for LLM reasoning. The observation motivated research on bridge tools: rerankers, query reformulators, and document selectors that align retrieval outputs with LLM preferences.

\paragraph{The architecture of preference translation.}

\textbf{BGM}~\cite{ke2024bridging} addresses the systematic ``preference gap'' between what traditional retrievers optimize for (surface-level relevance, lexical overlap) and what LLMs find useful for reasoning (contextual coherence, inferential support). BGM addresses this by training a T5-XXL ``bridge model'' that sits between a frozen retriever and a frozen generator (PaLM2-S), transforming the retriever's output into an LLM-friendly context.

Stage 1 uses supervised learning on synthetically generated preference data (documents that improve LLM task performance vs.\ those that do not), while Stage 2 employs reinforcement learning where the frozen LLM's final task success provides the reward signal. On HotpotQA, the bridged system achieves 35.6\% exact-match accuracy compared to 25.8\% with the best prior retriever, a relative improvement of 38\%.

BGM shows that specialized adaptation layers can be more effective than end-to-end fine-tuning. Rather than making a single retriever simultaneously satisfy IR metrics and LLM preferences, BGM decomposes the problem: the retriever handles broad recall, while the bridge model handles preference alignment. The same modularity recurs in the most successful T2 systems.

\paragraph{Synthesis: the multi-tool ecosystem.}

These advances reveal an architectural pattern: cascaded tool adaptation. State-of-the-art T2 systems now employ a pipeline of specialized tools (query reformulators, retrievers, selectors), each trained using different aspects of the frozen LLM's behavior as supervision. The decomposition offers several advantages:
\begin{itemize}
\item \textbf{Separation of concerns:} Each tool can optimize a specific sub-problem (recall vs. precision, speed vs. quality).
\item \textbf{Composability:} Tools can be mixed and matched; a new reranker can be trained without retraining the retriever.
\item \textbf{Efficiency:} Expensive operations (LLM inference) are deferred to the end of the pipeline, after cheaper tools have filtered the space.
\end{itemize}

An open question is how deep this hierarchy can go before compounding errors overwhelm the benefits. Empirical results suggest that 2--3 stages of tool adaptation (e.g., query reformulator $\rightarrow$ retriever $\rightarrow$ reranker) strike a good balance.

\subsubsection{Subagent-as-Tool}
\label{subsubsec:4.2.2}
The year 2025 marked a transition in T2 research from training reactive tools (retrievers that respond to queries) to training proactive subagents (autonomous systems that explore, plan, orchestrate, and refine their operations over multiple turns while serving frozen primary agents). The shift, enabled by advances in RLVR, applies not only to information retrieval but also to meta-cognitive processes such as workflow orchestration and memory management. 
The evolution can be organized into four families of subagents: (i) agentic searchers, (ii) memory-construction subagents, (iii) meta-cognitive planners and orchestrators, and (iv) self-evolving subagents.

\paragraph{Agentic searchers.}
\textbf{s3}~\cite{jiang2025s3} (EMNLP 2025) showed that training agentic tools can be radically more data-efficient than training agentic LLMs. The system trains a lightweight 7B ``searcher'' that performs multi-turn iterative search: generate queries, retrieve documents, select evidence, decide whether to search again or feed context to the frozen generator. The frozen generator (agent, Qwen2.5-14B or Claude) never updates, but provides the ultimate training signal through a metric called Gain Beyond RAG (GBR):
\begin{equation}
\text{GBR} = \text{Accuracy}\left(\mathcal{G}_{\text{frozen}}(q, D_{s3}), a\right) - \text{Accuracy}\left(\mathcal{G}_{\text{frozen}}(q, D_{\text{naive}}), a\right) \nonumber
\end{equation}
where $D_{s3}$ are documents retrieved by the trained searcher and $D_{\text{naive}}$ are documents from naive top-$k$ retrieval.

The reward directly measures the value added by the search tool, focusing training on examples where naive retrieval fails. s3 achieves 58.9\% average generation accuracy with only \textit{2.4k training samples}, 70$\times$ less data than Search-R1 (an A2-style agent requiring 170k examples) and 33$\times$ faster wall-clock training time. On specialized medical QA, s3 trained on general QA reaches 76.6\% accuracy versus 71.8\% for Search-R1, indicating that T2-trained tools learn more \textbf{generalizable search skills} than agents trained end-to-end.
The efficiency advantage arises because A2-style agent training must simultaneously learn (1) domain knowledge, (2) tool-use skills, and (3) task-specific reasoning, creating a high-dimensional optimization landscape. In T2, the frozen generator already possesses domain knowledge and reasoning ability, so the tool need only learn the procedural skill of effective search.

Similar to this idea, \textbf{DynamicRAG}~\cite{sun2025dynamicrag} (NeurIPS 2025) ``agentifies'' reranking: instead of static reorderings, an RL policy adapts how many and which documents to pass based on query hardness and retrieval noise, balancing quality and context cost. The training combines imitation learning on expert trajectories (to bootstrap reasonable behavior) with policy gradient RL where the generator's output provides the reward signal. The learned policy exhibits emergent adaptive behavior: for simple queries with high-quality initial retrieval, it presents fewer documents; for complex queries with noisy retrieval, it retrieves more broadly and reranks more aggressively.

\textbf{QAgent}~\cite{jiang2025qagent} further clarifies how to robustly train such search subagents. Its Stage 1 trains a 3B search agent end-to-end, rewarding it based on whether its own generated answer is correct, but this encourages reward hacking, where the agent prefers shallow, easily copyable evidence over genuinely informative documents. Its Stage 2 corrects this by switching to evaluation from a stronger frozen generator:
\begin{equation}
R_{\text{Stage2}} = \mathbb{I}\left[\mathcal{G}_{\text{frozen}}(q, D_{\text{agent}}) = a_{\text{correct}}\right], \nonumber
\end{equation}
rewarding the searcher only when the frozen model can answer correctly using its retrieved documents. The decoupling forces the subagent to optimize for retrieval quality rather than its own myopic behavior, reinforcing a core T2 principle: the frozen generator should not only consume a tool's outputs but also supervise its learning.

\paragraph{Learning to construct memory as a subagent.}
Extending beyond search, a complementary line of work treats long-term memory construction itself as a T2-style subagent problem. 
Mem-$\alpha$~\cite{wang2025mem} formulates memory management as RL over an explicit memory API, 
training a lightweight Qwen3-4B controller to operate a three-part external memory (core summary, semantic facts, episodic events) for a frozen backend generator. 
Only the memory-writing policy is optimized; the generator and retriever for downstream QA remain frozen. 
Rewards derive from verifiable outcomes (question-answering accuracy over long horizons, tool-call correctness, effective compression, and semantic validity of memory entries), so the subagent learns to construct compact yet sufficient memories that maximize the frozen model's utility. 
In experiments, Mem-$\alpha$ outperforms prior memory baselines and generalizes from $\sim$30k-token training sequences to contexts exceeding 400k tokens, serving as a concrete instance of a memory-construction subagent that adaptively curates the information diet for a fixed reasoning core.

\textbf{AutoGraph-R1}~\cite{tsang2025autograph} applies this symbiotic principle to the construction of structured Knowledge Graphs (KGs). Rather than relying on static extraction heuristics, it optimizes an LLM-based ``constructor subagent'' to generate KGs from raw text. The supervision signal is derived directly from the frozen agent's performance on downstream reasoning tasks (GraphRAG~\cite{xiao2025graphrag}) using the generated graph. The constructor thereby learns a policy that prioritizes functional utility, creating connectivity and paths that specifically facilitate the host agent's retrieval and reasoning, over intrinsic metrics like triple density.

\paragraph{Meta-cognitive and control subagents.}
Recent work trains subagents that shape how frozen models think (planning, steering, and budgeting computation) rather than what they retrieve or store.

\textbf{AI-SearchPlanner}~\cite{mei2025ai} introduces multi-objective optimization that balances effectiveness with efficiency. The system trains a planner tool (Qwen2.5-7B) that generates multi-step search strategies for a frozen generator, optimizing
\begin{equation}
\mathcal{J} = \mathbb{E}\left[R_{\text{outcome}} + \lambda \cdot R_{\text{process}} - \alpha \cdot \text{Cost}\right], \nonumber
\end{equation}
where $R_{\text{outcome}}$ measures final task success, $R_{\text{process}}$ evaluates the rationality of the search plan (critiqued by the frozen generator), and Cost penalizes excessive planning. By combining both outcome and process rewards~\cite{lightman2023let}, the frozen model acts as executor and teacher, so the planner internalizes not only ``what works'' but ``why it works.'' Varying $\lambda$ traces a Pareto frontier between cost and quality, yielding planners tailored to different deployment budgets.

\textbf{Advisor Models}~\cite{asawa2025train} generalize this idea to instance-wise natural-language steering. A small advisor model learns, via GRPO, to prepend context-specific advice that nudges a frozen foundation model toward preferred behaviors (style, safety, reasoning depth) without touching its weights. Within our taxonomy, such advisors function as trainable control interfaces or parametric memories that encode environment- and user-specific latents.

Bridging from advising to driving, \textbf{Matryoshka Pilot}~\cite{limatryoshka} (NeurIPS 2025) formalizes a controller-generator loop where a small white-box LLM controls a larger black-box LLM by emitting intermediate decomposition steps, plans, and summaries. Treating the black-box model as an environment, M-Pilot collects trajectory-level success signals and optimizes the controller with Iterative DPO. The method yields $\approx$3--7\% gains across reasoning, planning, and personalization benchmarks, and transfers plug-and-play across multiple black-box backends, further reinforcing the view of control subagents as portable T2 tools.

\paragraph{Learning to orchestrate frozen specialists.}
Orchestration-focused subagents train a dedicated policy to coordinate multiple frozen specialists.

\textbf{AgentFlow}~\cite{li2025flow} decomposes an agent into modules (planner, tool executor, verifier, and solution generator) implemented mostly as frozen Qwen2.5-7B-Instruct models. Only the planner is trained. Using Flow-GRPO, a single trajectory-level reward (correct vs.\ incorrect, judged by GPT-4o) is broadcast to all decisions in each rollout, with group-normalized advantages permitting effective credit assignment despite sparse rewards. 
A 7B AgentFlow planner achieves 57.3\% on search-intensive tasks (+14.9\% over AutoGen), 51.5\% on mathematical reasoning (+14.5\% over ToRL), and 33.1\% on GAIA, outperforming the much larger GPT-4 on several setups, demonstrating that learned orchestration of frozen specialists can rival or surpass monolithic models.

\paragraph{Self-evolving (sub)agent.}
A more advanced branch of the subagent-as-tool paradigm allows the tools themselves to co-evolve through self-generated tasks and rewards. 
\textbf{R-Zero}~\cite{huang2025rzero} instantiates two roles, a Solver and a Challenger, from the same base LLM. 
When the Solver is frozen, its successes, failures, and uncertainty (via self-consistency) define rewards that train the Challenger to propose tasks near the Solver’s capability frontier. 
Alternating these phases creates a bidirectional loop, yet each step still follows the T2 principle of optimizing a lightweight subagent under signals from a stronger or temporarily fixed core.
\textbf{Multi-Agent Evolve (MAE)}~\cite{chen2025mae} extends this design into a triadic architecture with a Proposer, Solver, and Judge. 
The Proposer and Judge operate as adaptive T2 subagents: the Judge learns to evaluate trajectories produced by the system, and the Proposer learns to generate diverse, high-quality, Solver-challenging tasks. 
Rather than tuning the main Solver, MAE improves performance by training these peripheral subagents to shape data, rewards, and curricula. 
Together, R-Zero and MAE illustrate a second generation of subagent-as-tool methods: self-evolving ecosystems that autonomously construct the learning conditions for otherwise frozen reasoning cores.

\paragraph{Synthesis: the maturation of T2.}

The subagent-as-tool paradigm progressively broadens the scope of T2: retrieval, memory construction, planning and orchestration, and finally self-evolution. 
Agentic searchers (s3, DynamicRAG, QAgent) optimize information acquisition for frozen generators; memory-construction subagents (Mem-$\alpha$) curate long-horizon state; meta-cognitive controllers and orchestrators (AI-SearchPlanner, Advisor Models, Matryoshka Pilot, AgentFlow) decide how tools and specialists are deployed; and self-evolving frameworks (R-Zero, Multi-Agent Evolve) autonomously generate curricula and reward signals. 
Decoupling tool training from generator training, while allowing tools to adapt to one another, yields systems that are more data-efficient, modular, and generalizable than monolithic alternatives.

\subsubsection{Agentic Memory and Skills}
\label{subsubsec:4.2.3}

An agent's memory system can be framed as an adaptive tool. As discussed in \S\ref{sec:overview}, the paradigm label depends on the memory's form and update mechanism: external non-parametric stores updated by a frozen agent's outputs are predominantly T2; pre-trained or plug-in modules are T1; and parametric or hybrid architectures occupy the boundary between tool and agent adaptation. The majority of systems reviewed below are T2, where the frozen agent's downstream task performance or outputs serve as the supervisory signal for tuning the memory module (how it writes, retrieves, reflects, and forgets). Several recent surveys provide complementary perspectives on this space. Hu et al.~\cite{hu2025memorySurvey} propose a unified \textit{forms, functions, dynamics} framework that organizes agent memory along three orthogonal dimensions: \textbf{forms} (token-level, parametric, and latent memory, distinguished by where and how information is stored), \textbf{functions} (factual memory for knowledge of users and environments, experiential memory for accumulated interaction outcomes, and working memory for active context management), and \textbf{dynamics} (the lifecycle of memory formation, evolution, and retrieval). This tripartite lens clarifies that memory adaptation operates simultaneously on the storage substrate, the content type, and the update policy. Within the experiential function, Hu et al.\ further distinguish three abstraction levels: \textit{case-based memory} (raw trajectory storage), \textit{strategy-based memory} (distilled workflows and heuristics), and \textit{skill-based memory} (executable code, APIs, and MCP protocols). This hierarchy maps naturally onto our adaptation framework: case-based and strategy-based memory are predominantly T2 (the frozen agent's success or failure determines what is retained), while skill-based memory bridges T2 (skill libraries curated by agent feedback) and T1 (pre-trained tool APIs that are agent-agnostic).

Zhang et al.~\cite{zhang2025survey} categorize the mechanisms that can be optimized as T2 tools, spanning short-term buffers, long-term experiential databases, and structured knowledge. Zhang et al.~\cite{zhang2024memorymechanism} provide a complementary survey focused specifically on the memory mechanisms of LLM-based agents, organizing existing work along the dimensions of memory formation (how memories are created), management (how they are updated and pruned), and utilization (how they inform future actions). Huang et al.~\cite{huang2026rethinking} offer a broader treatment that connects cognitive-science memory models (complementary learning systems, working memory) to foundation-agent architectures, while Jiang et al.~\cite{jiang2026anatomy} provide an empirical analysis of evaluation and system limitations, identifying benchmark saturation, metric validity gaps, and backbone-model dependence as open problems. Jiang et al.~\cite{jiang2025longtermmemory} argue that long-term memory is the foundation of AI self-evolution, framing persistent memory as the substrate that enables agents to accumulate knowledge, refine strategies, and improve autonomously over extended horizons. At a higher level of abstraction, Sumers et al.~\cite{sumers2023coala} propose the \textbf{Cognitive Architectures for Language Agents (CoALA)} framework, which organizes agent memory into working memory (the active context window), episodic memory (records of past interactions), semantic memory (general knowledge), and procedural memory (learned action routines). CoALA provides a principled decomposition that maps directly onto the T2 design space: each memory type can be independently adapted under frozen-agent supervision, and the framework clarifies which memory subsystem should be optimized for a given task profile. Together, these surveys establish that agentic memory is not a single mechanism but a design space with multiple axes: storage modality (parametric vs.\ explicit vs.\ latent), temporal scope (transient, session-level, persistent), content type (factual, experiential, working), abstraction level (cases, strategies, skills), and update policy (append-only, reflective, RL-optimized).

Closely related is the concept of \textit{agent skills}. Wu and Zhang~\cite{wu2025agentskills} argue that agent skills are best understood through the lens of \textit{procedural memory}: structured, reusable knowledge of \textit{how} to perform tasks, paralleling the cognitive-science distinction between declarative knowledge (knowing \textit{that}) and procedural knowledge (knowing \textit{how}). Under this framing, skills follow a lifecycle of \textbf{acquisition} (learning from demonstrations, exploration, or trajectory distillation), \textbf{representation} (as executable code, API specifications, MCP protocols, or textual procedures), \textbf{invocation} (retrieval and execution at inference time), and \textbf{refinement} (iterative improvement through reflection, RL, or collective sharing). Xu and Yan~\cite{xu2026agentskills} complement this procedural-memory view with a systems perspective: they emphasize progressive context loading, the role of \texttt{SKILL.md} and MCP as interface standards, and the security/governance problems that arise once skills become executable, shareable artifacts rather than passive memories. Fang et al.~\cite{fang2025memp} provide an empirical investigation of procedural memory in LLM agents, showing that the format and granularity of stored procedures significantly affect downstream task performance, and that agents benefit from adaptive selection between code-based and natural-language procedure representations depending on task complexity. Cao et al.~\cite{cao2025rememberme} propose a dynamic procedural memory framework in which the agent continuously refines its stored procedures based on execution outcomes, demonstrating that iterative memory refinement outperforms static memory accumulation on long-horizon planning tasks. The skill lifecycle maps onto the experiential memory hierarchy proposed by Hu et al.~\cite{hu2025memorySurvey}: case-based memory stores raw trajectories, strategy-based memory distills transferable heuristics and workflows, and skill-based memory compiles executable capabilities. Memory and skills are thus two facets of the same adaptation mechanism. Memory provides the storage substrate and organizational structure; skills provide the executable, composable content that makes that storage actionable for future tasks.

\paragraph{\textbf{Dynamic memory stores.}}
Foundational T2 memory architectures cluster into three design families based on how they organize stored information\cite{park2023generative, packer2023memgpt, zhong2024memorybank, lu2023memochat, modarressi2023ret, liang2023unleashing}:
\emph{hierarchical/OS-inspired} systems (MemGPT, Memory OS) impose explicit tiers (working memory vs.\ long-term storage) with page-eviction or garbage-collection policies;
\emph{reflection-based} systems (Generative Agents, A-MEM) let the agent's own outputs decide what to consolidate and when;
and \emph{graph/structured} systems (HippoRAG, SHIMI) index memories by relational or semantic structure rather than recency.
The design choice determines the system's trade-off between recall speed and associative richness. \textbf{MemGPT}~\cite{packer2023memgpt} formalizes this idea through an operating-system inspired memory hierarchy: a limited main-context window acts as ``RAM,'' while an unbounded external store serves as ``disk.'' The agent issues explicit read/write operations to move information between tiers, and a page-eviction policy decides what to retain in the active context. Generative Agents~\cite{park2023generative} maintain a memory stream of natural-language observations and use the agent's own reflections to periodically consolidate raw entries into higher-level abstractions, creating a two-tier store (observations and reflections) that supports planning over day-long time horizons. More recently, Guti{\'e}rrez et al.~\cite{gutierrez2025ragmemory} recast the transition from static retrieval-augmented generation to persistent agent memory as a form of non-parametric continual learning, showing that memory stores can absorb new information without catastrophic forgetting of earlier knowledge, a property that parametric fine-tuning struggles to guarantee. Kang et al.~\cite{kang2025memoryos} propose \textbf{Memory OS}, a layered architecture that treats memory management as an operating-system service, providing standardized APIs for storage, indexing, retrieval, and garbage collection that decouple memory logic from the agent's reasoning loop. \textbf{A-MEM}~\cite{xu2025amem} takes an agentic approach to memory management: the LLM itself decides how to organize its memory store, dynamically creating, linking, and restructuring entries based on content relevance and task demands rather than relying on fixed indexing heuristics. \textbf{HippoRAG}~\cite{gutierrez2024hipporag} draws on the neuroscience of hippocampal memory indexing, separating a neocortical component (an LLM that encodes passages into a knowledge graph) from a hippocampal index (a retrieval module that performs pattern completion over the graph). The design mirrors complementary learning systems theory and enables associative retrieval across passages that share no lexical overlap, yielding substantial gains over standard RAG baselines on multi-hop QA benchmarks. \textbf{SHIMI}~\cite{helmi2025shimi} proposes a decentralized, semantic hierarchical memory index designed for scalable multi-agent reasoning, where each agent maintains a local memory shard organized by semantic similarity, and a coordination protocol enables cross-agent memory retrieval without centralizing all information in a single store. Liu et al.~\cite{liu2024raise} demonstrate that fine-tuning a small LLM specifically for memory-enhanced conversation (retrieval, summarization, and persona maintenance) can transform a base model into a conversational agent with persistent memory, illustrating that the memory management pipeline itself can be a trainable component.

\paragraph{\textbf{Experiential and reflective memory.}}
A substantial line of T2-aligned research focuses on memory modules that learn from experience. These tools allow a frozen agent to store, reflect on, and learn from its own output (e.g., entire trajectories), often using verbal reinforcement or self-correction, thereby building a curriculum of strategies and avoiding repeated failures without updating the core LLM's weights \cite{shinn2023reflexion, yaoretroformer, liu2023think}. \textbf{Reflexion}~\cite{shinn2023reflexion} pioneered verbal reinforcement learning: after each task attempt, the frozen agent generates a natural-language self-critique that is appended to an episodic memory buffer, and subsequent attempts condition on these reflections. The approach converts scalar reward signals into rich textual feedback without gradient updates, achieving 91\% pass@1 on HumanEval and strong results on AlfWorld and HotPotQA. \textbf{Retroformer}~\cite{yaoretroformer} extends this idea by training a dedicated retrospective model that generates targeted feedback for the frozen actor, separating the ``what went wrong'' analysis from the ``try again'' generation. \textbf{Think-in-Memory}~\cite{liu2023think} introduces a recall-then-post-think pipeline: the agent first retrieves relevant historical thoughts from memory, then performs post-thinking to recombine and adapt them to the current context, enabling long-term coherence without expanding the context window. \textbf{Agent Workflow Memory (AWM)}~\cite{wang2024awm} takes a complementary approach by extracting reusable \textit{workflows} (structured action sequences) from an agent's past successful trajectories and storing them in a dedicated memory module. When the agent encounters a new task, it retrieves the most relevant workflow and uses it as a template, reducing planning errors on web navigation benchmarks by 24\% compared to agents without workflow memory. AWM shows that the unit of experiential memory need not be a raw trajectory or a verbal reflection; structured procedural abstractions can serve as a more efficient retrieval target. Pan et al.~\cite{pan2025personalmemory} investigate memory construction and retrieval specifically for personalized conversational agents, showing that the choice of memory granularity (individual facts vs.\ user-level summaries) and retrieval strategy (recency-biased vs.\ relevance-biased) significantly affects the agent's ability to maintain coherent, personalized dialogue over hundreds of turns. When these accumulated experiences are distilled into reusable, composable capabilities, the result is a \textit{skill library}, discussed in a dedicated paragraph below.

\paragraph{\textbf{Structured memory (graphs, trees, and databases).}}
To move beyond linear text, some T2 memory tools structure information in richer forms, such as knowledge graphs, trees, or symbolic databases. The frozen agent's outputs are used as signals to ``tune'' this structured tool, for example by adding new nodes, updating relationships, or writing to a database. \textbf{AriGraph}~\cite{anokhin2024arigraph} builds an episodic knowledge graph during interaction: the agent emits observations that are parsed into entity-relation triples and merged into a persistent graph, which the agent later queries for multi-hop reasoning. \textbf{ChatDB}~\cite{hu2023chatdb} externalizes memory into a relational database, translating the agent's natural-language outputs into SQL operations (INSERT, SELECT, UPDATE) that maintain structured records across conversation turns. Rezazadeh et al.~\cite{rezazadeh2024isolated} propose a hierarchical tree memory that organizes conversations into nested topic clusters, enabling efficient retrieval at multiple granularity levels. \textbf{Zep}~\cite{rasmussen2025zep} introduces a temporal knowledge graph architecture for agent memory that explicitly models time as a first-class dimension, allowing the agent to reason about when facts were learned, how they have changed, and which are most current, a capability that flat vector stores lack. These structured representations can be more efficiently queried and reasoned over by the frozen agent, effectively externalizing complex memory management into a specialized, adaptive tool \cite{cheng2024information}.

\paragraph{Parametric and hybrid memory architectures.}
A parallel line of research explores parametric memory mechanisms that complement explicit external stores. \textbf{Memory$^3$}~\cite{yang2024memory3} introduces a three-tier memory hierarchy for language models: (1) model weights as implicit long-term memory, (2) an explicit memory pool of retrievable text chunks as semi-parametric memory, and (3) the context window as working memory. A lightweight ``memory circuitry'' learns to route information between tiers, enabling the model to offload factual knowledge to the explicit pool while reserving parametric capacity for reasoning. On language modeling benchmarks, Memory$^3$ with a 2.4B parameter model and an external memory pool matches the perplexity of a 6.4B model without external memory, demonstrating that explicit memory can substitute for raw parameter count. \textbf{Titans}~\cite{behrouz2025titans} incorporates a differentiable long-term memory module directly into the attention mechanism, maintaining a persistent memory state that is updated through gradient-based learning during inference. Unlike standard transformers that rely solely on the fixed context window, Titans can selectively store and retrieve information across arbitrarily long sequences. These parametric approaches complement the external-store paradigm (MemGPT, Generative Agents) by demonstrating that memory adaptation can also occur within the model's computational graph, blurring the boundary between T1 (pre-trained memory modules) and T2 (agent-supervised memory adaptation).

\paragraph{Episodic memory as a trainable module.}

\textbf{Memento}~\cite{zhou2025memento} shows that an agent's memory system can be optimized as an external tool without any modification to the LLM planner. The system combines a frozen GPT-4.1 high-level planner with a trainable episodic case memory module. The memory stores past problem-solving trajectories, and the tool being trained is a neural Q-function that learns a case retrieval policy: which past cases to present to the frozen planner when facing a new problem.

The training signal is binary task success or failure: the sparse, trajectory-level reward is broadcast to all case-selection decisions in that trajectory, and a soft Q-learning algorithm updates the retrieval policy. The frozen LLM never sees the Q-values or policy internals; it simply receives retrieved cases as context and generates its plan.
Memento achieves top-tier performance: 87.88\% on GAIA validation (ranked 1st), 79.40\% on GAIA test (3rd place), and 95.0\% on SimpleQA. Ablations show that case-based memory adds 4.7--9.6\% absolute improvement on out-of-distribution tasks. Only the memory is trained; the same frozen LLM that performed worse without memory now excels because its information diet has been optimized.

\paragraph{\textbf{Memory operations as learnable skills.}}
A recent line of work reframes the memory management problem itself: rather than hand-designing fixed operations (append, retrieve, summarize), the operations themselves become learnable and evolvable. \textbf{MemSkill}~\cite{zhang2026memskill} introduces a controller-executor-designer loop in which a controller selects which memory skill to apply (e.g., extract key facts, consolidate related entries, prune stale information), an LLM executor generates skill-guided memory content, and a designer module analyzes failure cases to evolve the skill set over time. The skill library starts with a small seed set and grows as the designer identifies recurring failure patterns that existing skills cannot address. On LongMemEval~\cite{wu2024longmemeval}, LoCoMo, HotpotQA, and ALFWorld, MemSkill outperforms static memory baselines by learning task-appropriate memory granularity, for instance extracting fine-grained entity attributes for QA tasks while retaining coarser episode summaries for planning tasks. The approach demonstrates that memory adaptation need not be limited to what is stored; how the storage operations themselves are performed can also be optimized under frozen-agent supervision.

\paragraph{\textbf{Test-time memory curation.}} Another prominent example of T2 memory adaptation at inference-time is \textbf{Dynamic Cheatsheet (DC)}~\cite{suzgun2025dynamic}, a lightweight framework that provides a ``persistent, evolving memory'' for black-box LMs. The system operates ``without modifying their underlying parameters'' and requires no gradient-based updates. The framework consists of two core modules: a Solution Generator and a Memory Curator. The \textbf{Memory Curator} is the adaptive T2 tool: it operates without access to ground-truth labels, assessing the correctness and efficiency of solutions autonomously after they are produced by the frozen generator. Based on this self-assessment, the curator updates the memory by storing concise, transferable snippets such as ``reusable strategies, code snippets, and general problem-solving insights'', rather than full, uncurated transcripts. \textbf{ReasoningBank}~\cite{ouyang2025reasoningbank} extends this test-time curation concept by creating a memory framework that explicitly distills generalizable reasoning strategies from both successful and self-judged failed experiences. Unlike methods that store raw trajectories or only successful routines, ReasoningBank analyzes failures to extract ``crucial preventative lessons''. The curated bank of reasoning strategies is then retrieved to guide the agent in future tasks. The framework also introduces memory-aware test-time scaling, which uses the curated memory to guide a scaled exploration, where the diverse experiences from scaling help forge stronger, more generalizable memories. ReasoningBank demonstrates effectiveness on complex benchmarks like WebArena~\citep{zhou2024webarena} and SWE-Bench~\citep{jimenez2024swebench}.

\paragraph{Skill libraries as adaptive tools.}
A particularly important form of T2 memory adaptation is the construction of reusable \textit{skill libraries}, which instantiate skill-based experiential memory~\cite{hu2025memorySurvey}. Rather than storing raw trajectories (case-based) or abstract heuristics (strategy-based), these systems distill successful experiences into modular, composable, and executable capabilities that the frozen agent can invoke in future tasks. The skill representations span a continuum from fine-grained code snippets to standardized APIs and MCP protocols~\cite{wu2025agentskills}.

Table~\ref{tab:skill_libraries} organizes representative skill-library systems along two axes: \emph{skill granularity} (what is stored---code, heuristics, trajectories, specialist agents, or synthesized tools) and \emph{acquisition mechanism} (how skills are obtained---demonstration, exploration, reflection, or RL). We discuss representative systems from each cell below; the table provides a concise reference for the remainder.

\begin{table*}[t]
\centering
\caption{Skill-library design space. Systems are organized by \emph{skill granularity} (what is stored) and \emph{acquisition mechanism} (how skills are obtained). Domain abbreviations: MC = Minecraft, CU = Computer Use, Web = Web Navigation, Sci = Scientific Domains, Gen = General/Multi-domain.}
\label{tab:skill_libraries}
\small
\begin{tabular}{llp{2.6cm}p{2.2cm}p{4.8cm}}
\toprule
\textbf{Granularity} & \textbf{Acquisition} & \textbf{Representative Systems} & \textbf{Domain} & \textbf{Key Distinction} \\
\midrule
 & Exploration & Voyager, OS-Copilot, CRADLE & MC, CU & Frozen agent generates \& verifies code skills \\
\multirow{-2}{*}{Executable code} & RL & PAE & Web & Proposer--Agent--Evaluator RL loop \\
\midrule
 & Reflection & EXPEL & Gen & Extracts rules from success \& failure \\
\multirow{-2}{*}{Abstract heuristics} & Demonstration & Synapse & Gen & Stores state--action exemplars \\
\midrule
Full trajectories & Exploration & JARVIS-1 & MC & Multimodal (visual+textual) indexing \\
\midrule
 & Exploration & SkillWeaver & Web & Community-shared API-level skills \\
\multirow{-2}{*}{API / MCP protocols} & Reflection & CASCADE & Sci & Validates against domain criteria \\
\midrule
Programs (compositional) & Reflection & ASI~\cite{wang2025asi} & Gen & Structured programs with loops \& conditionals \\
\midrule
 & Demonstration & LATM & Gen & Maker--user cost amortization \\
\multirow{-2}{*}{Synthesized tools} & Reflection & CREATOR & Gen & Abstract plan $\to$ tool $\to$ verify loop \\
\midrule
Specialist agents & Exploration & AgentStore & CU & Registry + meta-controller selection \\
\bottomrule
\end{tabular}
\end{table*}

\textbf{Voyager}~\cite{wang2023voyager} pioneered this approach in Minecraft: a frozen GPT-4 agent generates executable code for each new task, and successful programs are stored in a growing skill library indexed by natural-language descriptions. When the agent encounters a related task, it retrieves and composes previously verified skills instead of reasoning from scratch. The skill library thus functions as a T2 tool whose contents are curated entirely by the frozen agent's success or failure signals. Voyager discovers 63 unique items in Minecraft (3.3$\times$ more than baselines) and demonstrates continual skill accumulation without catastrophic forgetting, because new skills are added to the library without modifying existing ones. Related Minecraft agents adopt complementary strategies: \textbf{GITM}~\cite{zhu2023gitm} (Ghost in the Minecraft) equips a frozen LLM with text-based knowledge and memory modules that decompose goals into executable sub-goal sequences, while \textbf{DEPS}~\cite{wang2023deps} uses an interactive describe-explain-plan-select loop that enables the agent to recover from failures by re-planning based on environmental feedback. Both systems maintain persistent memory of past interactions, but neither constructs a reusable skill library, highlighting Voyager's contribution of \textit{composable, code-level skill accumulation}.

\textbf{JARVIS-1}~\cite{wang2023jarvis1} extends the skill-library paradigm to open-world multi-task settings by combining multimodal memory with a pre-trained action planner. The system maintains a multimodal memory that stores successful task-completion trajectories as (observation, plan, action) tuples. Given a new task, JARVIS-1 retrieves the most relevant past trajectories from memory and uses them as in-context examples for the planner. The memory is indexed by both visual similarity (using CLIP embeddings of game frames) and textual similarity (using descriptions of task goals), enabling cross-modal retrieval. On 200+ Minecraft tasks spanning crafting, combat, and exploration, JARVIS-1 achieves substantially higher success rates than Voyager and other baselines, particularly on long-horizon tasks requiring 10+ sequential sub-goals, where memory-guided planning reduces compounding errors.

\textbf{EXPEL}~\cite{zhao2024expel} generalizes this principle beyond game environments: a frozen LLM extracts reusable ``insights'' (abstract rules and heuristics) from both successful and failed trajectories, storing them in a persistent experience pool. On ALFWorld and WebShop, EXPEL improves task success by 18\% and 12\% respectively over non-adaptive baselines, demonstrating that experiential skill extraction transfers across task instances.

\textbf{Synapse}~\cite{zheng2024synapse} combines trajectory-level memory with state-action exemplars, enabling a frozen agent to select from a library of demonstrated skills at each decision step.

\textbf{OS-Copilot}~\cite{wu2024oscopilot} extends the skill-library paradigm to general-purpose computer use. The system maintains a self-improving library where each skill is a Python function with a natural-language docstring. When the agent encounters a novel task, it first searches the library for relevant skills; if none exist, it generates a new skill, executes it, and upon success stores it for future reuse. The self-improvement loop operates without human annotation: the agent's own execution outcomes (success or failure) determine which skills enter the library. On the GAIA benchmark, OS-Copilot with its accumulated skill library outperforms baselines that lack persistent skill storage, confirming that cross-session skill reuse is a key advantage of the T2 approach.

\textbf{CRADLE}~\cite{tan2024cradle} applies skill-centric design to general computer control, maintaining a skill curation module that stores, retrieves, and composes reusable action sequences for GUI interaction. The system decomposes complex computer tasks into sub-goals, matches each sub-goal against the skill library, and executes the retrieved skill or generates a new one. Successful skills are refined through repeated use, creating a self-improving repertoire that grows with deployment experience.

\textbf{AppAgent}~\cite{zhang2023appagent} demonstrates autonomous skill discovery for smartphone use. The agent interacts with mobile apps, discovers UI element functionalities through trial and error, and stores the learned interaction patterns as reusable procedural knowledge in an external document. When encountering a new app or task, the agent retrieves relevant skills from this document to guide its actions. The exploration-based discovery requires no human demonstration and transfers across applications that share similar UI patterns.

\textbf{Agent S}~\cite{agashe2024agents} introduces experience-augmented hierarchical planning for computer use, where the agent maintains a library of learned ``experience'' consisting of successful task-completion strategies. The system uses a Manager-Worker architecture: the Manager decomposes tasks into sub-goals using retrieved experience, and the Worker executes individual actions. The experience retrieval mechanism matches new sub-goals against the library using both semantic similarity and structural task-graph matching, enabling transfer of learned strategies to structurally similar but previously unseen tasks.

\textbf{AgentStore}~\cite{jia2024agentstore} addresses a different facet of the skill management problem: integrating heterogeneous specialist agents as reusable skills within a unified system. Rather than building a monolithic agent, AgentStore maintains a registry of specialized agents (each trained for a particular domain or tool), and a meta-controller learns to select and compose these specialists for new tasks. The system supports dynamic registration of new specialist agents, enabling the skill library to grow without retraining the meta-controller.

\textbf{ExACT}~\cite{yu2024exact} combines reflective Monte Carlo Tree Search with exploratory learning to teach agents to systematically explore unfamiliar environments. The agent builds a skill tree through exploration, where each node represents a discovered interaction pattern. Successful exploration trajectories are stored as reusable skills, and the MCTS-guided exploration ensures diverse coverage of the action space. The approach bridges the gap between trial-and-error skill discovery (as in AppAgent) and more structured planning-based approaches. Zhao et al.~\cite{zhao2024agenticskill} formalize the problem of \textit{agentic skill discovery}, proposing a framework in which an LLM agent autonomously identifies, abstracts, and catalogs reusable behavioral patterns from its interaction history, providing a theoretical grounding for the empirical skill-library systems described above.

\textbf{SkillWeaver}~\cite{zheng2025skillweaver} demonstrates that web agents can autonomously discover, create, and refine reusable API-level skills through exploration. The agent navigates websites, identifies recurring interaction patterns, and encodes them as callable API functions that can be shared across agents and tasks. The collective skill library grows through a community mechanism in which multiple agents contribute discovered skills, enabling rapid coverage of diverse web domains without centralized supervision. \textbf{PAE}~\cite{zhou2025pae} (Proposer-Agent-Evaluator) formalizes autonomous skill discovery as a three-component loop: a Proposer generates candidate tasks, an Agent attempts them, and an Evaluator scores the outcomes to produce reward signals for reinforcement learning. The decoupled design allows each component to be optimized independently, and the RL-based training enables the agent to acquire skills that are difficult to specify through demonstrations alone. On web navigation benchmarks, PAE discovers skills that transfer across websites and task types, achieving competitive performance with substantially less human supervision than imitation-learning baselines. \textbf{SAGE}~\cite{wang2025sage} pushes this RL direction further by treating the skill library itself as part of the learning loop: during sequential rollouts across related tasks, newly generated skills accumulate in the library and are immediately available for subsequent tasks, while a skill-integrated reward encourages both useful skill creation and effective reuse. This is a particularly clean instantiation of T2-style self-improvement because the agent's future competence changes through the evolving skill substrate rather than through monolithic retraining alone. Wang et al.~\cite{wang2025asi} take a complementary programmatic approach: rather than storing skills as natural-language descriptions or raw code, ASI induces structured programs (with conditionals, loops, and subroutine calls) that capture the compositional structure of agentic tasks. The programmatic representation enables systematic generalization to longer task horizons and novel task compositions that flat skill libraries struggle with. \textbf{CASCADE}~\cite{huang2025cascade} extends skill creation to scientific domains, where an LLM agent autonomously develops and evolves a library of computational chemistry skills (e.g., molecular property prediction, reaction pathway search) through cumulative self-improvement. Each skill is validated against domain-specific correctness criteria before entering the library, and the system tracks skill dependencies to enable compositional reuse across multi-step scientific workflows.

A related but distinct form of skill construction is \textit{tool creation}: rather than storing successful action traces, the agent synthesizes new reusable tools (typically as executable functions) that encapsulate solutions to recurring subproblems. \textbf{LATM}~\cite{cai2024latm} (Large Language Models as Tool Makers, ICLR 2024) formalizes a two-phase protocol in which a ``tool maker'' LLM creates Python functions from task demonstrations, and a lightweight ``tool user'' LLM applies these functions to new instances. Once created, a tool can be reused across hundreds of instances without regeneration, amortizing the cost of the expensive maker model. On reasoning benchmarks (GSM8K, MATH, TabMWP), LATM matches the performance of a single large model while reducing per-instance cost by up to 79\%, because the tool user need only call the pre-built function rather than re-derive the solution strategy. \textbf{CREATOR}~\cite{qian2023creator} (EMNLP 2023 Findings) disentangles abstract reasoning from concrete implementation: given a problem, the agent first formulates an abstract solution plan, then creates a tool (code function) that implements the plan, and finally applies the tool. By separating ``what to do'' from ``how to do it,'' CREATOR enables the agent to correct tool implementations through a verification-and-refinement loop without re-deriving the abstract strategy. On Creation Challenge and MATH benchmarks, CREATOR outperforms both chain-of-thought and direct code-generation baselines, and the created tools transfer to new problem variants. Both LATM and CREATOR demonstrate that skill construction can operate at the tool-creation level: the agent does not merely retrieve past experiences but synthesizes new, reusable abstractions that compress recurring reasoning patterns into callable functions.

These systems share a common architecture: the frozen agent generates experience, a curation mechanism (code verification, self-reflection, or success filtering) selects what to retain, and a retrieval interface makes accumulated skills available for future reasoning. The architecture maps directly onto the T2 paradigm, where the agent remains fixed and the skill library evolves under agent-derived supervision. A key distinction among these systems is the \textit{granularity} of the stored skill: Voyager and OS-Copilot store executable code, EXPEL stores abstract heuristics, JARVIS-1 stores full trajectories, AgentStore stores entire specialist agents, LATM and CREATOR store synthesized tool functions, and SkillWeaver stores API-level callable functions. The choice of granularity trades off composability (fine-grained code skills compose more flexibly) against transfer breadth (coarse heuristics generalize more broadly across task distributions). A second axis of variation is the \textit{acquisition mechanism}: demonstration-based (Synapse), exploration-based (AppAgent, ExACT, PAE), reflection-based (EXPEL, Reflexion), and RL-based (PAE, Memento). Recent work on \textit{multi-agent procedural memory} adds a third axis. \textbf{LEGOMem}~\cite{han2025legomem} introduces modular procedural memory for multi-agent systems, where each agent maintains a local memory of learned workflows and a coordination layer enables agents to share, compose, and specialize procedural knowledge across the team. The modular design allows individual agents to refine their skill sets independently while benefiting from collective experience, a pattern that extends the single-agent skill library to collaborative settings. At the meta-level, \textbf{ADAS}~\cite{hu2024adas} (Automated Design of Agentic Systems) automates the design of agent architectures themselves, using an LLM to iteratively propose, evaluate, and refine agentic building blocks (prompts, tool configurations, control flows). ADAS can be viewed as a second-order skill management process: rather than accumulating task-level skills, the system accumulates reusable architectural patterns that improve performance across diverse benchmarks.

\paragraph{Skills in embodied and robotic settings.}
The skill-library paradigm extends naturally to embodied agents, where skills correspond to physical action sequences grounded in sensorimotor experience. \textbf{SayCan}~\cite{ahn2022saycan} introduced the idea of grounding LLM-generated plans in robotic affordances: the LLM proposes candidate skills in natural language, and a value function trained on real-world robot data scores each skill by its probability of successful execution in the current state. The product of the LLM's semantic score and the affordance score selects the next skill, ensuring that plans are both semantically reasonable and physically feasible. \textbf{ProgPrompt}~\cite{singh2022progprompt} takes a programmatic approach, prompting the LLM to generate Python-like programs that compose primitive robot actions (grasp, place, navigate) into multi-step task plans. The skill primitives are pre-defined, but the composition logic is generated dynamically, enabling zero-shot transfer to novel household tasks in VirtualHome. \textbf{Eureka}~\cite{ma2023eureka} addresses a different bottleneck in embodied skill learning: reward function design. The system uses an LLM to generate candidate reward functions as code, evaluates them through physics simulation, and iteratively refines the reward based on training statistics. On 29 robotic manipulation and locomotion tasks in IsaacGym, Eureka-generated rewards match or exceed human-designed rewards in 83\% of cases, enabling dexterous pen-spinning that was previously unsolved. \textbf{RoboGen}~\cite{wang2023robogen} combines LLM-based task generation with automated skill learning: the LLM proposes new tasks, generates simulation environments, and designs reward functions, while a reinforcement learning agent acquires the corresponding motor skills. The self-supervised loop enables open-ended skill accumulation without human task specification. These embodied systems illustrate that agent skills are not limited to digital environments; the same T2 principle (frozen high-level planner, adaptive skill repertoire) applies when skills involve physical actions grounded in sensorimotor feedback.

\paragraph{Memory navigation and interactive reading.}
A complementary approach treats long documents or conversation histories as environments that the agent navigates through memory-like retrieval actions. \textbf{MemWalker}~\cite{chen2023memwalker} constructs a tree-structured summary hierarchy over a long document and trains the agent to ``walk'' through this tree by iteratively selecting which branch to expand, converting the memory-retrieval problem into a sequential decision-making task. \textbf{ReadAgent}~\cite{lee2024readagent} takes a human-inspired approach: the agent first performs a ``gisting'' pass that compresses each page of a long document into a short summary, stores these gist memories, and then selectively re-reads original pages when detailed information is needed. On QuALITY and NarrativeQA, ReadAgent with gist memory achieves performance comparable to systems that process the full context, while using 3--5$\times$ fewer tokens. \textbf{CoRAG}~\cite{wang2025corag} (Chain-of-Retrieval Augmented Generation) decomposes complex queries into a chain of intermediate retrieval steps, each conditioned on the results of the previous step. A rejection-sampling training procedure teaches the retriever to produce retrieval chains that maximize downstream answer quality. On multi-hop QA benchmarks, CoRAG outperforms single-step and iterative retrieval baselines by learning retrieval strategies tailored to the reasoning structure of each query. \textbf{Adaptive-RAG}~\cite{jeong2024adaptiverag} learns to classify query complexity and route each query to the appropriate retrieval strategy (no retrieval, single-step, or multi-step), avoiding unnecessary retrieval for simple queries while ensuring thorough multi-hop retrieval for complex ones. The routing classifier is trained on silver labels derived from the correctness of different retrieval strategies, making the retrieval depth itself an adaptive, learnable parameter. These methods illustrate that memory adaptation extends beyond storage to include adaptive retrieval strategies that determine when and how deeply to access stored information.

\paragraph{Adapting the embedding space.}

An approach for tool scalability is \textbf{ToolkenGPT}~\cite{hao2023toolkengpt}, which represents tools as learnable token embeddings within the frozen LLM's vocabulary. The entire LLaMA-13B/33B model remains frozen; only a small embedding matrix $W_{\tau} \in \mathbb{R}^{|T| \times d}$ (where $|T|$ is the number of tools and $d$ is the embedding dimension) is trained. These ``toolkens'' are concatenated with the standard vocabulary, and the frozen LLM learns to predict them like any other token.

Training uses supervised learning on parallel sequences where ground-truth tool calls are replaced with toolken placeholders. The loss is masked so that only the toolken predictions (and subsequent argument tokens) contribute gradients to $W_{\tau}$. The approach is parameter-efficient: adding 234 tools requires training only $234 \times 4096 \approx 1M$ parameters (for LLaMA's 4096-dimensional embeddings), compared to the 13B+ parameters of the full model.
ToolkenGPT achieves 73\% one-hop accuracy on FuncQA (vs. 57\% for ReAct), 75\% supervised accuracy on 234-relation KAMEL, and 68\% success on VirtualHome with 58 action/object tools. New tools can be added by expanding $W_{\tau}$ and continuing training, without full model retraining.

ToolkenGPT shows that adaptation can occur at the interface layer (the embedding space) rather than the parameter layer (the LLM weights), offering a middle ground between fully frozen T1 systems and fully fine-tuned A1/A2 systems.

Beyond the paradigms above, several recent approaches further extend the tool adaptation framework. These methods introduce new training objectives, modalities, and architectural innovations that broaden the scope of tool adaptation.

\textbf{UniMuR}~\cite{wang2024unified} trains unified multimodal embeddings aligned with frozen LLM semantic representations, yielding 6.5\% R@1 improvement on MMDialog.
\textbf{DIFO}~\cite{tang2024source} adapts frozen CLIP through task-specific prompt learning via mutual information maximization for source-free domain adaptation.
\textbf{V2L Tokenizer}~\cite{zhu2024beyond} trains encoder-decoder structures mapping images to frozen LLM token space, using the frozen vocabulary as quantization codebook to enable low-level vision tasks with frozen text LLMs.
\textbf{Sysformer}~\cite{sharma2025sysformer} trains a small transformer that adapts the system-prompt embeddings based on each user prompt while keeping the LLM frozen. 
Supervision comes entirely from the frozen model’s own likelihoods over refusal and compliance targets, augmented by reconstruction and optional classifier losses. 

Common patterns emerge across T2 methods: lightweight training of small modules (millions of parameters) while keeping LLMs frozen (billions of parameters), semantic exploitation of rich representations (hidden states, token spaces, vocabularies), modality bridging between vision/retrieval/tools and frozen text LLMs, and strong generalization to zero-shot or unseen settings. 

\section{Comparison of Adaptation Paradigms}
\label{sec:comparison}

We now compare the four adaptation paradigms: (A1) Agent Adaptation with Tool Execution Signal, (A2) Agent Adaptation with Agent Output Signal, (T1) Agent-Agnostic Tool Adaptation, and (T2) Agent-Supervised Tool Adaptation. We first establish a conceptual framework, then analyze the agent-centric (A1/A2) and tool-centric (T1/T2) paradigms in depth, with special focus on the emergent ``subagent-as-tool'' and ``graduation'' concepts, and conclude with a quantitative synthesis of critical trade-offs.

\subsection{A Framework for Comparison}
\label{sec:comparison_framework}

We compare the four paradigms along four main axes.

\begin{itemize}[leftmargin=15pt]
    \item \textbf{Cost and Flexibility:} We use ``cost'' to refer to compute and engineering effort required for adaptation, and ``flexibility'' to mean how easily the system's behavior can be reconfigured. A1/A2 provide high parametric flexibility (the entire agent policy can change), whereas T1/T2 provide high system-level flexibility (capabilities can be added, swapped, or composed via tools) but remain bounded by the frozen agent's intrinsic reasoning power.
    \item \textbf{Data Efficiency:} Beyond raw compute, the amount of training data required differs substantially across paradigms. Recent evidence suggests that T2 methods can match or surpass A2-style end-to-end agent training with orders of magnitude less data, by only training small subagents around a frozen backbone.
    \item \textbf{Generalization Capability:} This axis captures how well an adaptation strategy transfers to new tasks, agents, or environments. T1 tools trained on broad data distributions generalize across different agents and tasks, while T2 tools often inherit cross-domain robustness from the frozen foundation models supervising them. A1/A2, especially on-policy variants, risk overfitting to specific environments without explicit regularization.
    \item \textbf{Modularity and System Evolution:} This axis focuses on engineering implications: how easily a system can be extended or maintained over time. Tool-centric paradigms (T1/T2) support modular evolution and hot-swapping of components; agent-centric paradigms (A1/A2) tend to be monolithic and may suffer from catastrophic forgetting when adapted repeatedly.
\end{itemize}

\noindent
In summary, agent adaptation (A1/A2) rewrites the entire policy in a single model, offering high parametric flexibility at the cost of expensive retraining and potential side-effects on unrelated behaviors. Tool adaptation (T1/T2) attaches specialized tools that can be added or replaced without destabilizing the base agent, bounded by what the frozen agent can understand and use. Data efficiency and generalization both favor tool-centric adaptation (see \S\ref{sec:comparison_synthesis} for quantitative evidence), while modularity---the ability to swap tools without retraining the core---is often more decisive than raw performance in practice.

\begin{table*}[t]
\centering
\caption{High-level qualitative comparison of the four adaptation paradigms. ``Flex.'' denotes the dominant form of flexibility: \emph{parametric} (within a single agent policy) vs. \emph{system-level} (via modular tools and orchestration).}
\label{tab:paradigm_comparison}
\resizebox{\textwidth}{!}{%
\begin{tabular}{@{}llllll@{}}
\toprule
\textbf{Paradigm} & \textbf{ID} & \textbf{Locus of Adaptation} & \textbf{Supervision Signal} & \textbf{Cost \& Flexibility} & \textbf{Modularity \& Evolution} \\
\midrule
Agent, Tool Signal & A1 & Core Agent Policy & Tool Execution  & High Cost, High \emph{Param.} Flex. & Monolithic, Risk of Overfitting \\
Agent, Output Signal & A2 & Core Agent Policy & Agent Output  & High Cost, High \emph{Param.} Flex. & Monolithic, Risk of Forgetting \\
Tool, Agent-Agnostic & T1 & External Tool & Agent-Independent & Low Cost, High \emph{System} Flex. & High (Plug-and-Play) \\
Tool, Agent-Supervised & T2 & External Tool & Frozen Agent Output & Low Cost, High \emph{System} Flex. & High (Symbiotic, No Forgetting) \\
\bottomrule
\end{tabular}%
}
\end{table*}

\subsection{Agent Adaptation Paradigms: A1 and A2}
\label{sec:comparison_agent}

The two agent-centric paradigms both modify the agent's core parameters but differ in their training signals and optimization objectives.

\subsubsection{A1: Optimizing Tool Mechanics via Causal Feedback}
\label{sec:comparison_a1}

A1 on-policy methods rely on causal, immediate, and fine-grained reward signals. The supervision source is the verifiable outcome of tool execution itself, not a downstream task metric. For example, DeepRetrieval~\cite{jiang2025deepretrieval} formalizes query reformulation as an MDP where reward is directly derived from retrieval metrics like Recall@K or NDCG, and RLEF~\cite{gehring2025rlefgroundingcodellms} frames code synthesis with rewards from test-case execution. The approach contrasts with A2 signals that only evaluate the final answer.

Conceptually, A1 on-policy RL optimizes tool-use mechanics: it teaches the agent how to wield tools correctly, grounding behavior in environment ``physics'' (``this syntax executes'', ``this query retrieves''). Direct engagement with ground-truth feedback drives strong performance in domains with verifiable, deterministic outcomes.

\textbf{Quantitative evidence.} Mechanistic optimization under A1 achieves strong performance in specialized domains:
\begin{itemize}[leftmargin=15pt]
\item \textbf{Retrieval:} DeepRetrieval achieves roughly $3\times$ improvement in recall (65.1\% vs.\ 24.7\%) on literature search~\cite{jiang2025deepretrieval}.
\item \textbf{Code reasoning:} R1-Code-Interpreter reaches 72.4\% accuracy on 37 test tasks through multi-stage RL~\cite{chen2025r1}.
\end{itemize}

However, learning through trial-and-error introduces practical challenges: it requires careful reward design, KL-regularized PPO or GRPO, curriculum learning, and dynamic sampling for stable convergence.

\subsubsection{A2: Optimizing Tool Strategy via Holistic Rewards}
\label{sec:comparison_a2}

A2 methods instead use holistic, sparse, and high-level rewards based on agent output quality (typically final answer correctness) that depend on tool usage but do not directly supervise individual tool calls. ReSearch~\cite{chen2025learning}, trained on multi-hop QA, optimizes when and how to search. The reward asks not ``was this particular search good?'' but ``did the entire process of thinking, searching, and reasoning lead to the correct answer?''

Thus A2 optimizes tool-use strategy and coordination. Rather than learning search mechanics (assuming a T1 retriever handles that), it learns the cognitive policy for when to search, what to search for, and how to integrate results. The strategic focus explains why ReSearch reports emergent reflection and self-correction behaviors during RL training~\cite{chen2025learning}.

\textbf{Quantitative evidence.} Strategic optimization under A2 proves effective for complex, multi-step reasoning:
\begin{itemize}[leftmargin=15pt]
\item \textbf{Retrieval-augmented QA:} ReSearch yields 9--22\% absolute gains over strong iterative RAG baselines~\citep{chen2025learning}.
\item \textbf{Factual accuracy:} R1-Searcher reports up to 24\% improvement over strong RAG baselines, with improved factual accuracy and reduced hallucination through learned retrieval policy~\cite{song2025r1}.
\end{itemize}

In terms of flexibility, A2 offers the richest parametric flexibility: the agent can change its entire global strategy for orchestrating tools and reasoning, but each such change requires expensive retraining, and the resulting policy is baked into a single large model.

\textbf{A1 \& A2: Signal Source as a Reliability Axis.} Beyond taxonomic categorization, the distinction between A1 and A2 determines the granularity and scope of the adaptation signal.

\begin{itemize}[leftmargin=15pt]
    \item Tool-execution signals (A1) are \textit{grounded}, \textit{causal}, and \textit{process-oriented}. The feedback is produced by an environment or tool whose semantics are independent of the agent’s internal beliefs (e.g., code execution, retrieval metrics, formal proof checkers). Such grounding enables learning that is tightly coupled to intermediate correctness and tool mastery, but often comes with higher interaction cost and environment dependence.
    
    \item Agent-output signals (A2) are \textit{holistic}, \textit{flexible}, and \textit{outcome-oriented}. Rewards are assigned to the agent’s final outputs, derived from either verifiable ground truths (e.g., gold answers, math solutions) or subjective preferences (e.g., reward models). While this allows for end-to-end task optimization, relying solely on terminal signals can make the agent vulnerable to shortcut learning (getting the right answer for the wrong reason) and sparse feedback issues compared to the dense signals of A1.
\end{itemize}

\subsection{Tool Adaptation Paradigms: T1 and T2}
\label{sec:comparison_tool}

Tool-centric paradigms shift optimization from the expensive agent to cheaper external tools. These paradigms sacrifice some parametric flexibility (the agent policy stays fixed) but gain system-level flexibility: the tool ecosystem can be grown, specialized, and rewired without touching the main agent.

\subsubsection{T1: The ``Graduated Agent'' as Subagent-as-Tool}
\label{sec:comparison_t1}

T1 is defined by agent-agnostic, pre-trained, plug-and-play components. A central concept within T1 is the subagent-as-tool, which follows a development lifecycle.

At one extreme, we have static, foundational tools like SAM~\cite{kirillov2023segment} or AlphaFold2~\cite{jumper2021highly}, trained once on massive datasets and deployed as fixed APIs. They primarily encapsulate learned representations or simulators and can be called by any agent.

At the other extreme are dynamic, graduated tools: adaptive agents from \S\ref{sec:comparison_agent} can be trained under A1 or A2 and then frozen and reused as T1 tools. The ``Graduation Lifecycle'' (A1 $\rightarrow$ T1) proceeds as:

\begin{enumerate}[leftmargin=15pt]
\item \textbf{Train (A1/A2):} Use on-policy RL or outcome-based RL to train an agent for a specific task (e.g., DeepRetrieval as a search-query rewriter, Code-R1 as a code generator).
\item \textbf{Freeze:} Once the agent reaches expert performance, freeze its parameters.
\item \textbf{Deploy (T1):} The frozen expert becomes a T1 ``subagent-as-tool'' callable by any higher-level agent.
\end{enumerate}

Concrete examples already follow this pattern. DeepRetrieval is trained via on-policy A1 RL as a query reformulation agent~\cite{jiang2025deepretrieval}, but once frozen it can be used as an interchangeable T1 retrieval-augmentation tool in many different pipelines. Similarly, SWE-Grep~\cite{cognition2025swegrep} is trained as a specialized RL subagent for fast, multi-turn, highly parallel code context retrieval, and then exposed as a tool that software-engineering agents (e.g., SWE-Agent or Cursor-style IDE agents) can call for high-quality repository search. In both cases, the ``graduated'' subagent encapsulates a learned policy (not just a representation) and slots into new systems without retraining.

From the flexibility perspective, T1 offers high system-level flexibility: different T1 tools can be assembled into various configurations, or one tool (e.g., a retriever) can be replaced without touching the agent. The cost of adding a capability is proportional to the size of the corresponding tool, not the backbone agent. The trade-off is that the tools are not tailored to any particular agent; the agent must adapt its prompts or orchestration logic to whatever interface the tool exposes.

\subsubsection{T2: Inverting the Optimization Target}
\label{sec:comparison_t2}

T2 inverts the conventional adaptation direction. Rather than adapting the agent to use tools better, T2 adapts the tools to better serve a fixed agent (see \S\ref{subsec:agent_output_as_signal_for_tool} for the full rationale). This reframes the foundation model from optimization target to supervision source.

In practice, the frozen host agent (e.g., GPT, Claude) supplies reasoning and reward signals, while lightweight subagents (e.g., 7B models) learn to reshape information for the host's consumption. The central advantage is the decoupling of skill from knowledge. A traditional A2 agent like Search-R1 must learn (1) domain knowledge, (2) tool-use skills, and (3) task reasoning simultaneously, which creates a complex optimization landscape. In T2, the frozen generator already possesses (1) and (3); the T2 subagent needs only learn procedural skill.

T2 subagent families also instantiate an architectural strategy: unbundling the agent's monolithic cognitive loop (Perceive-Plan-Act-Reflect) into specialized, independently trainable submodules:
\begin{itemize}[leftmargin=15pt]
\item \textbf{Optimizing ``Perception'' (Agentic Searchers):} Systems like s3, DynamicRAG, and QAgent train search subagents to decide what to query, where to search, and when to stop~\cite{jiang2025s3}.
\item \textbf{Optimizing ``Reflection'' (Memory Construction):} Subagents such as Mem-$\alpha$ learn memory-writing policies via RL, rewarded based on whether stored experiences improve future performance for the frozen generator.
\item \textbf{Optimizing ``Planning'' (Meta-Cognitive Planners):} Subagents like AI-Search Planner and AgentFlow decide how tools and specialists are deployed. AgentFlow~\cite{li2025flow} trains only a lightweight planner that orchestrates frozen specialists using trajectory-level rewards, achieving 33.1\% on GAIA and surpassing the much larger GPT-4.
\end{itemize}

T2 thus achieves high system-level flexibility: new T2 subagents can be trained and attached incrementally (e.g., a better planner, a domain-specific searcher, a new memory module), without retraining the host agent. Compared to T1, T2 trades some agent-agnosticity for tighter compatibility: tools are specialized for a given frozen agent, leading to higher data efficiency and better end-to-end performance under the same backbone.

\begin{table*}[t]
\centering
\caption{Quantitative comparison of flagship adaptation methods across paradigms and domains.}
\label{tab:quant_comparison}
\resizebox{\textwidth}{!}{%
\begin{tabular}{@{}llllll@{}}
\toprule
\textbf{Method} & \textbf{Paradigm} & \textbf{Domain} & \textbf{Training Signal} & \textbf{Key Result} & \textbf{Key Insight} \\
\midrule
DeepRetrieval~\cite{jiang2025deepretrieval} & A1 & Retrieval & Recall@$K$, nDCG  & $\sim$3$\times$ Recall (65.1\% vs.\ 24.7\%) & Causal RL optimizes tool mechanics \\
RLEF~\cite{gehring2025rlefgroundingcodellms} & A1 & Code & Test-case pass rate & Stable multi-turn PPO training & Dense execution signals enable A1 RL \\
Code-R1~\cite{code-r1} & A1 & Code & Sandboxed execution & Reward quality $>$ data quantity & Clean rewards reduce training cost \\
ReSearch~\cite{chen2025learning} & A2 & RAG QA & Final answer EM & 9--22\% gains over RAG & Holistic RL optimizes tool strategy \\
R1-Searcher~\cite{song2025r1} & A2 & RAG QA & Final answer correctness & 24\% over RAG baselines & Emergent search-reason interleaving \\
Memento~\cite{zhou2025memento} & T2 & Memory & Binary task success & +4.7--9.6\% on OOD tasks & Memory alone can transform performance \\
s3~\cite{jiang2025s3} & T2 & Retrieval & GBR from frozen gen. & 58.9\% Acc.\ w/ 2.4k samples & High data efficiency (see caveats in text) \\
AgentFlow~\cite{li2025flow} & T2 & Planning & Final answer correctness & 33.1\% on GAIA (beats GPT-4) & Learned orchestration of specialists \\
\bottomrule
\end{tabular}%
}
\end{table*}

\subsection{Synthesis: Data Efficiency and Modularity (A2 vs.\ T2)}
\label{sec:comparison_synthesis}
The sharpest empirical comparison arises between A2 and T2. Both aim to produce capable tool-using systems, but they place the learning burden in different places. A2 adapts the agent, letting it internalize tool-use strategies; T2 adapts the tools, letting them learn to support a fixed agent.

The retrieval-augmented generation domain offers an illustrative case study. Comparing Search-R1 (A2) and s3 (T2):
\begin{itemize}[leftmargin=15pt]
\item \textbf{A2 approach} (Search-R1): Trains the entire Qwen2.5 agent, requiring roughly {170k examples} to co-adapt internal knowledge, reasoning, and tool-use policy~\cite{jin2025search}.
\item \textbf{T2 approach} (s3): Trains only a lightweight 7B ``searcher'' subagent using frozen-generator feedback (GBR), achieving comparable performance (58.9\% average accuracy) with only {2.4k training samples}~\cite{jiang2025s3}.
\end{itemize}

\noindent\textbf{Caveats.} This comparison, while suggestive, is not a controlled experiment. Search-R1 trains the full Qwen2.5 agent end-to-end, whereas s3 trains only a 7B search subagent paired with a frozen generator (which may itself be Qwen2.5-7B, 14B, or Claude-3-Haiku). The two systems thus differ simultaneously in optimization target, backbone composition, and system architecture; the observed efficiency gap cannot be attributed to a single factor. Controlled cross-paradigm comparisons that isolate the effect of paradigm choice from confounding architectural differences remain an important open problem.

With this caveat, the case study illustrates a broader architectural principle. T2 simplifies the learning problem by assuming the backbone already handles most of knowledge and reasoning, and only learning a narrow procedural skill in a small subagent. A2's optimization landscape is higher-dimensional: the agent must simultaneously adjust its knowledge, reasoning style, and tool-use policy. On specialized medical QA, T2-trained s3 reaches 76.6\% accuracy vs.\ A2-trained Search-R1's 71.8\%~\cite{jiang2025s3}, consistent with the hypothesis that narrower optimization targets generalize more robustly, though alternative explanations (e.g., differences in training data distribution) cannot be ruled out.

From an engineering perspective, T2 offers modularity advantages. Updating an A2 agent requires retraining the monolithic model, potentially inducing catastrophic forgetting. In a T2 architecture, new tools can be trained and hot-swapped without touching the host agent, allowing continuous evolution of the peripheral ecosystem while the core remains stable.

\subsection{Strategic Recommendations}
\label{sec:comparison_summary}
Choosing an adaptation strategy requires balancing computational cost, data efficiency, and system modularity. The following guidelines distill the empirical and architectural evidence presented above into concrete recommendations for practitioners.

\noindent\textbf{A1} is best suited for local, mechanistic mastery of verifiable tools in stable domains such as retrieval, code execution, and SQL.
By optimizing directly on executable outcomes, A1 develops strong low-level competence and causal grounding, giving practitioners precise control over tool behavior with robust alignment to verifiable signals.
The cost is high: each training run requires substantial compute, and the resulting specialization often generalizes poorly across tasks or tool interfaces.

\noindent\textbf{A2} is appropriate when a single agent must orchestrate multiple tools and perform holistic reasoning.
A2 internalizes when, how, and why to invoke tools, yielding deeply integrated, end-to-end policies for complex workflows.
The price of this integration is expensive monolithic retraining and susceptibility to catastrophic forgetting when the agent is subsequently adapted to new domains.

\noindent\textbf{T1} provides horizontal scalability and reusability.
The category spans both static foundational models (e.g., SAM, AlphaFold2) and ``graduated'' subagents, A1/A2-trained experts that are frozen and redeployed as reusable modules (e.g., DeepRetrieval, SWE-Grep~\citep{cognition2025swegrep}).
These ``subagents-as-tools'' encapsulate learned procedural expertise while remaining decoupled from any specific host agent, enabling plug-and-play modularity and broad compositional flexibility.
Because T1 tools are trained without reference to a particular agent, they may be under-optimized for any given host's reasoning style.

\noindent\textbf{T2} inverts the adaptation question: rather than adapting the agent to use tools better, it trains lightweight tools and subagents under frozen-agent supervision to better serve a fixed backbone (e.g., s3-style searchers, planners, advisors, and memory builders).
The host agent provides high-level reasoning and reward signals, while T2 subagents learn narrow procedural skills that can be added, replaced, or composed without touching the backbone.
T2 achieves high data efficiency for new skills and mitigates catastrophic forgetting through modular updates, but subagent capability is bounded by the supervising agent's quality, and multi-subagent pipelines introduce orchestration complexity and potential error compounding.

\begin{wrapfigure}{r}{0.5\textwidth}
    \centering
    \includegraphics[width=0.48\textwidth]{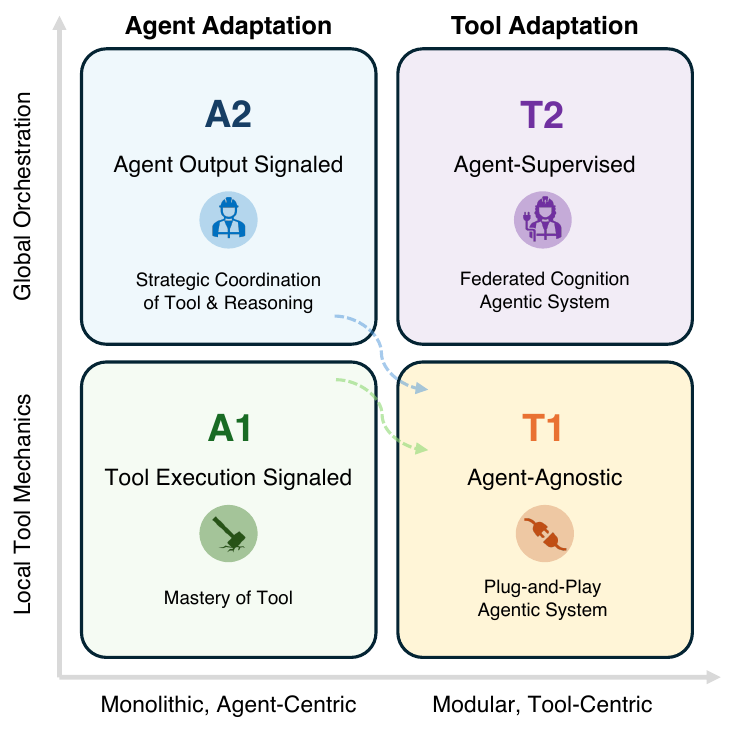}
    \caption{\textbf{An interpretive projection of the four adaptation paradigms.} 
    Unlike the definitional axes in \S\ref{sec:overview} (what is adapted $\times$ signal source), this figure offers a complementary design-space view.
    The x-axis captures monolithic-to-modular evolution, 
    while the y-axis represents local-to-systemic orchestration. 
    A1/A2 inhabit the agent-centric half, whereas T1/T2 
    represent modular and system-level flexibility. Dotted arrows show that A1/A2-trained agents can graduate as tools for T1.}
    \label{fig:adaptation_landscape}
\end{wrapfigure}

\noindent
Taken together, these four paradigms can be projected onto a complementary $2\times2$ design-space view (Figure~\ref{fig:adaptation_landscape}). Note that this projection is an interpretive comparison, not the definitional axes introduced in \S\ref{sec:overview}; the definitional taxonomy classifies methods by \emph{what is adapted} and \emph{how the signal is obtained}, whereas the axes here capture broader architectural tendencies:  
\textbf{(i)} the local-to-systemic spectrum (y-axis), from low-level control of specific tools (A1/T1) to holistic orchestration of multi-tool reasoning (A2/T2); and  
\textbf{(ii)} the monolithic-to-modular spectrum (x-axis), from end-to-end retraining of a single agent (A1/A2) to compositional adaptation via distributed subagents and tools (T1/T2).
\noindent
Viewed through this lens, A1 and A2 occupy the agent-centric half of the landscape: they directly reshape the policy parameters of the core agent, offering rich parametric flexibility but incurring heavy costs in compute, data, and stability.  
T1 and T2, by contrast, occupy the tool-centric half: they shift learning outward into a modular ecosystem, permitting incremental evolution, specialization, and compositional reuse.  
The two axes interact nonlinearly: A1 $\!\rightarrow$ T1 reflects the ``graduation path'' (frozen experts becoming reusable subagents), while A2 $\!\rightarrow$ T2 follows the ``federation path'' (frozen backbones supervising a growing constellation of adaptive specialists).  
In practice, mature agentic architectures increasingly inhabit the upper-right quadrant (T2): high modularity and high orchestration, where foundation agents serve as stable cognitive centers and peripheral subagents continuously evolve to extend their capabilities.

The resulting picture clarifies the emerging division of labor in agentic AI research.  
A1/A2 remain necessary for generating new reasoning competencies or re-aligning a model's internal cognition, tasks that require modifying the agent's core.  
T1/T2, however, dominate system construction: they support continual growth, fine-grained specialization, and safe parallel experimentation.  
The prevailing design trend thus points toward hybrid systems: frozen foundation models at the center, surrounded by a modular set of T1/T2 subagents trained for specific procedural roles, with occasional A1/A2 updates marking evolutionary leaps in the agent’s internal reasoning.

\section{Evaluation}
\label{sec:eval}

Evaluating agentic AI adaptation requires moving beyond the static, single-score paradigms that dominate foundation-model benchmarking.
An adapted agent-tool system functions as an ongoing process---allocating computation, coordinating tool calls, and stabilizing under delayed or noisy feedback---rather than as a static predictor.
Accordingly, our evaluation framework is organized around four complementary lenses:
(\S\ref{subsec:benchmark_landscape}) a \textbf{benchmark landscape} that maps existing evaluation suites onto the four adaptation paradigms (A1/A2/T1/T2);
(\S\ref{subsec:eval_signal}) an analysis of \textbf{adaptation-signal metrics}, distinguishing verifiable execution metrics from holistic utility metrics;
(\S\ref{subsec:adapt_dynamics}) a treatment of \textbf{adaptation dynamics}, covering sample efficiency, generalization, and continual stability;
(\S\ref{subsec:systemic_eval}) a discussion of \textbf{systemic requirements}, including cost, safety, and alignment;
and (\S\ref{subsec:eval_discussion}) a forward-looking \textbf{discussion} on benchmark evolution and open challenges.

\subsection{Benchmark Landscape Mapped to A1/A2/T1/T2}
\label{subsec:benchmark_landscape}

We situate the growing space of agent benchmarks within the taxonomy developed in this paper.
Our goal is not to list benchmarks (readers can consult Table~\ref{tab:benchmarks} for a comprehensive catalog) but to clarify which aspect of adaptation a given benchmark evaluates, to expose systematic gaps, and to illustrate how the same system can appear very different when measured under different paradigm-aligned metrics.

\paragraph{Benchmarks for A1 (tool-execution-signaled agent adaptation).}
A1 adaptation requires benchmarks whose reward signal originates from a verifiable tool execution outcome: code compilation, test-case pass rates, retrieval recall, SQL query correctness, or formal-proof verification.
Representative suites include coding benchmarks such as \textit{HumanEval}~\citep{chen2021evaluating}, \textit{MBPP}~\citep{austin2021program}, \textit{LiveCodeBench}~\citep{jain2025livecodebench}, and \textit{SWE-Bench}~\citep{jimenez2024swebench}, where execution-based evaluation (pass@$k$ or test-suite pass rate) directly measures the quality of the agent's tool invocation.
Retrieval-oriented benchmarks like \textit{MTEB Retrieval}~\citep{muennighoff2023mteb} similarly provide deterministic, grounded metrics (nDCG, Recall@$K$).
In A1-aligned benchmarks, the evaluation signal is causally tied to the tool output rather than to the agent's final answer: a correct retrieval query or a passing unit test counts as success whether or not the agent produces a good final response.

\paragraph{Benchmarks for A2 (agent-output-signaled agent adaptation).}
A2 adaptation optimizes the agent's final output quality, which may depend on tool use but is evaluated as a whole.
Multi-hop question-answering benchmarks such as those used in Search-R1~\citep{song2025r1} and ReSearch~\citep{chen2025learning}, where the reward is final-answer correctness after a sequence of retrieval and reasoning steps, exemplify A2 evaluation.
More broadly, reasoning benchmarks (\textit{GSM8K}~\citep{cobbe2021training}, \textit{GPQA Diamond}~\citep{rein2024gpqa}), chat and instruction-following benchmarks (\textit{IFEval}~\citep{zhou2023instruction}, \textit{Arena-Hard}~\citep{li2025from}), and general-purpose agent benchmarks such as \textit{GAIA}~\citep{mialon2024gaia} and \textit{AgentBench}~\citep{liu2024agentbench} all evaluate the agent's end-to-end output.
In A2-aligned benchmarks, the adaptation signal reflects the overall quality of the agent's reasoning and synthesis, rather than the correctness of any single tool call.

\paragraph{The A1--A2 evaluation gap.}
Because A1 and A2 metrics expose orthogonal failure modes, the two lenses yield complementary diagnostics.
High A1 scores (e.g., DeepRetrieval's $3\times$ recall improvement~\citep{jiang2025deepretrieval}) coupled with low A2 scores reveal a synthesis bottleneck; the reverse pattern (high A2, low A1) reveals memorization or shortcut reasoning that bypasses genuine tool use.
The same logic applies in code: high pass@$k$ (A1) with low code quality (A2) indicates test-gaming.
Comprehensive evaluation requires both metric families.

\paragraph{Benchmarks for T1 (agent-agnostic tool adaptation).}
T1 tools are trained independently and evaluated without reference to any particular agent.
Classic retrieval benchmarks (\textit{MTEB}~\citep{muennighoff2023mteb}), embedding quality suites, and standalone model evaluations (e.g., evaluating a dense retriever's recall or a code-generation model's pass@$k$ in isolation) serve as T1 benchmarks.
The key property is that the evaluation measures the tool's intrinsic quality (retrieval precision, segmentation IoU, transcription accuracy) without conditioning on how a downstream agent consumes the output.

\paragraph{Benchmarks for T2 (agent-supervised tool adaptation).}
T2 benchmarks must evaluate the tool's contribution to a fixed agent's downstream performance.
These remain the least standardized category.
In practice, T2 evaluation is typically performed by measuring the frozen agent's task success rate when equipped with the adapted tool versus a baseline tool.
For instance, \textsc{s3}~\citep{jiang2025s3} evaluates its learned search subagent by measuring the frozen generator's final-answer accuracy on multi-hop QA; AgentFlow~\citep{li2025flow} evaluates its learned planner by measuring the frozen backbone's score on GAIA.
The critical methodological requirement for T2 evaluation is counterfactual comparison: holding the agent fixed while varying only the tool, to isolate the tool's marginal contribution.
For memory-centric T2 systems, \textbf{LongMemEval}~\citep{wu2024longmemeval} provides a dedicated benchmark that tests five core capabilities of long-term interactive memory: information extraction, multi-session reasoning, temporal reasoning, knowledge updates, and abstention when memory is insufficient. The benchmark reveals that even state-of-the-art systems struggle with temporal ordering and knowledge updates, confirming that memory management remains a bottleneck for long-horizon deployment. A related gap concerns skill-lifecycle evaluation: recent work on agent skills as procedural memory~\citep{wu2025agentskills} emphasizes that tools evolve through acquisition, invocation, and refinement stages, yet no existing T2 benchmark measures the quality of these intermediate stages (e.g., whether a skill library improves through reuse or degrades through drift). Security-oriented evaluations are starting to emerge: \textbf{Skill-Inject}~\citep{schmotz2026skillinject} benchmarks prompt-injection attacks delivered through skill files, while \textit{Agent Skills in the Wild}~\citep{liu2026skillswild} audits marketplace skills at ecosystem scale. These are important complements, but they evaluate attack exposure rather than the marginal utility or long-term quality of the adapted skill layer.
Dedicated T2 benchmarks that systematically vary the frozen agent and measure marginal tool value remain scarce, representing a significant gap in current evaluation practice.

\paragraph{Integrated agent benchmarks.}
Several recent benchmarks evaluate the full agent-tool system in realistic, long-horizon environments, cutting across multiple paradigms.
We group them by what they primarily reveal about adaptation.
\emph{Environment grounding}: \textit{WebArena}~\citep{zhou2024webarena} and \textit{OSWorld}~\citep{xie2024osworld} embed agents in self-hosted web and desktop environments, requiring multimodal perception and multi-application coordination, with success measured by execution-based functional correctness.
\emph{Multi-tool coordination}: \textit{$\tau$-Bench}~\citep{yao2025taubench}, \textit{$\tau^2$-Bench}~\citep{barres2025tau}, and \textit{GTA}~\citep{wang2024gta} evaluate multi-tool task execution with success-rate metrics over real-world tool invocations.
\emph{Scale and endurance}: \textit{AgencyBench}~\citep{li2026agencybench} evaluates six core agentic capabilities across 32 scenarios requiring an average of 90 tool calls and one million tokens per task, while \textit{The Tool Decathlon}~\citep{li2025toolathlon} spans 32 software applications with 604 tools and 108 tasks.
These benchmarks stress-test agent-tool interaction but typically report only endpoint metrics, making it impossible to isolate which adaptation paradigm drives the observed gains.

\definecolor{sectionblue}{RGB}{230,245,255}
\definecolor{sectiongreen}{RGB}{235,255,235}
\definecolor{cardinal}{RGB}{180,0,0}
\definecolor{white}{RGB}{255,255,255}

\small
\rowcolors*{1}{gray!7}{white}
\renewcommand{\arraystretch}{1.15}

\begin{longtable}[t]{>{\raggedright\arraybackslash}p{4.5cm}
                  >{\raggedright\arraybackslash}p{4.5cm}
                  >{\raggedright\arraybackslash}p{2.3cm}
                  >{\raggedright\arraybackslash}p{1.8cm}
                  >{\raggedright\arraybackslash}p{1.8cm}
                  >{\raggedright\arraybackslash}p{1.2cm}}

\rowcolor{white}
\caption{Comprehensive Overview of Standard Benchmarks for Agent Adaptation. The \textbf{Evaluation Target} column indicates whether a benchmark evaluates standalone agent (model) capabilities or tool-augmented agent behavior. The \textbf{Paradigm} column indicates which adaptation paradigm(s) the benchmark is most naturally aligned with (see \S\ref{subsec:benchmark_landscape}). In math benchmarks, the tools used typically include code execution. In Tool Use \& Agents benchmarks, tools vary by task and may include search engines, code executors, terminals, APIs, and other external systems.} 
\label{tab:benchmarks}\\

\toprule
\rowcolor{white}
\textbf{Benchmark} & \textbf{Task Format} & \textbf{Evaluation Protocol} & \textbf{Eval.\ Target} & \textbf{Paradigm} & \textbf{Links} \\
\midrule
\endfirsthead

\rowcolor{white}
\multicolumn{6}{c}{\tablename\ \thetable\ -- Continued from previous page} \\
\toprule
\rowcolor{white}
\textbf{Benchmark} & \textbf{Task Format} & \textbf{Evaluation Protocol} & \textbf{Eval.\ Target} & \textbf{Paradigm} & \textbf{Links} \\
\midrule
\endhead

\rowcolor{white}
\multicolumn{6}{r}{\textit{Continued on next page}} \\
\endfoot

\bottomrule
\endlastfoot
\rowcolor{sectionblue}\multicolumn{6}{c}{\textbf{(a) Knowledge \& Commonsense}} \\
\midrule
MMLU \citep{hendrycks2021measuring} & Multi-choice QA & Accuracy & Agent & A2 & \href{https://huggingface.co/datasets/cais/mmlu}{\huggingface} \\
MMLU-Pro \citep{wang2024mmlu} & Multi-choice QA & Accuracy & Agent & A2 & \href{https://huggingface.co/datasets/TIGER-Lab/MMLU-Pro}{\huggingface} \\
MMLU-Redux \citep{gema2025we} & Multi-choice QA & Accuracy & Agent & A2 & \href{https://huggingface.co/datasets/edinburgh-dawg/mmlu-redux-2.0}{\huggingface} \\
AGIEval-en \citep{zhong2024agieval} & Multi-choice QA & Accuracy & Agent & A2 & \href{https://huggingface.co/datasets/lighteval/agi_eval_en}{\huggingface} \href{https://github.com/ruixiangcui/AGIEval}{\faGithub} \\
BIG-Bench Hard \citep{suzgun2023challenging} & Multi-choice QA & Accuracy & Agent & A2 & \href{https://huggingface.co/datasets/lukaemon/bbh}{\huggingface} \href{https://github.com/suzgunmirac/BIG-Bench-Hard}{\faGithub} \\
ARC-Challenge \citep{clark2018think} & Multi-choice QA & Accuracy & Agent & A2 & \href{https://huggingface.co/datasets/allenai/ai2_arc}{\huggingface} \\
TruthfulQA \citep{lin2022truthfulqa} & Multi-choice QA & Accuracy & Agent & A2 & \href{https://huggingface.co/datasets/EleutherAI/truthful_qa_mc}{\huggingface} \href{https://github.com/sylinrl/TruthfulQA}{\faGithub} \\
Winogrande \citep{sakaguchi2021winogrande} & Multi-choice QA & Accuracy & Agent & A2 & \href{https://huggingface.co/datasets/allenai/winogrande}{\huggingface} \\
HellaSwag \citep{zellers2019hellaswag} & Multi-choice QA & Accuracy & Agent & A2 & \href{https://huggingface.co/datasets/Rowan/hellaswag}{\huggingface} \\
\midrule
\rowcolor{sectionblue}\multicolumn{6}{c}{\textbf{(b) Reasoning}} \\
\midrule
GSM8K \citep{cobbe2021training} & Free-form numerical answer & Accuracy & Agent & A2 & \href{https://huggingface.co/datasets/openai/gsm8k}{\huggingface} \\
AIME 2025 & Free-form numerical answer & Accuracy & Agent / Tool & A1/A2 & \href{https://huggingface.co/datasets/opencompass/AIME2025}{\huggingface} \\
GPQA Diamond \citep{rein2024gpqa} & Multi-choice QA & Accuracy & Agent / Tool & A2 & \href{https://huggingface.co/datasets/fingertap/GPQA-Diamond}{\huggingface} \\
IMOAnswerBench \citep{luong2025towards} & Free-form numerical answer & Accuracy & Agent / Tool & A1/A2 & \href{https://huggingface.co/datasets/Hwilner/imo-answerbench}{\huggingface} \\
TheoremQA \citep{chen2023theoremqa} & Free-form numerical answer & Accuracy & Agent / Tool & A2 & \href{https://huggingface.co/datasets/TIGER-Lab/TheoremQA}{\huggingface} \\
HMMT February 2025 & Free-form numerical answer & Accuracy & Agent / Tool & A1/A2 & \href{https://huggingface.co/datasets/MathArena/hmmt_feb_2025}{\huggingface} \\
HMMT Nov 2025 & Free-form numerical answer & Accuracy & Agent / Tool & A1/A2 & \href{https://huggingface.co/datasets/MathArena/hmmt_nov_2025}{\huggingface} \\
MiniF2F \citep{jiang2023draft} & Theorem proving & Success rate & Agent / Tool & A1 & \href{https://huggingface.co/datasets/Tonic/MiniF2F}{\huggingface} \\
Humanity's Last Exam \citep{phan2025humanity} & Multi-Choice QA & Accuracy & Agent / Tool & A2 & \href{https://huggingface.co/datasets/cais/hle}{\huggingface} \\
\midrule
\rowcolor{sectionblue}\multicolumn{6}{c}{\textbf{(c) Coding}} \\
\midrule
HumanEval \citep{chen2021evaluating} & Function-level code generation & Execution-based (pass@k) & Agent / Tool & A1 & \href{https://huggingface.co/datasets/openai_humaneval}{\huggingface} \\
MBPP \citep{austin2021program} & Function-level code generation & Execution-based (pass@k) & Agent / Tool & A1 & \href{https://huggingface.co/datasets/google-research-datasets/mbpp}{\huggingface} \\
LiveCodeBench \citep{jain2025livecodebench} & Function-level code generation & Execution-based (pass@k) & Agent / Tool & A1 & \href{https://huggingface.co/datasets/livecodebench/code_generation_lite}{\huggingface} \href{https://github.com/LiveCodeBench/LiveCodeBench}{\faGithub} \\
SciCode \citep{tian2024scicode} & Function-level code generation & Execution-based (pass@k) & Agent / Tool & A1 & \href{https://huggingface.co/datasets/SciCode1/SciCode}{\huggingface} \href{https://github.com/scicode-bench/SciCode}{\faGithub} \\
MultiPL-E & Function-level code generation & Execution-based (pass@k) & Agent / Tool & A1 & \href{https://huggingface.co/datasets/nuprl/MultiPL-E}{\huggingface} \\
SWE-Bench \citep{jimenez2024swebench} & Repository-level code modification & Test-suite pass rate & Agent / Tool & A1/A2 & \href{https://huggingface.co/datasets/princeton-nlp/SWE-bench}{\huggingface} \href{https://github.com/princeton-nlp/SWE-bench}{\faGithub} \\
\midrule
\rowcolor{sectionblue}\multicolumn{6}{c}{\textbf{(d) Chat \& Instruction Following}} \\
\midrule
IFEval \citep{zhou2023instruction} & Instruction compliance & Rule-based scoring & Agent & A2 & \href{https://huggingface.co/datasets/google/IFEval}{\huggingface} \\
IFBench \citep{pyatkin2025generalizing} & Instruction compliance & Rule-based scoring & Agent & A2 & \href{https://huggingface.co/collections/allenai/ifbench}{\huggingface} \href{https://github.com/allenai/IFBench}{\faGithub} \\
Scale AI Multi Challenge \citep{deshpande2025multichallenge} & Instruction following & LLM-as-judge & Agent & A2 & ~\href{https://github.com/ekwinox117/multi-challenge}{\faGithub} \\
Arena-Hard-V2 \citep{li2025from} & Pairwise chat comparison & LLM-as-judge & Agent & A2 & \href{https://huggingface.co/datasets/lmarena-ai/arena-hard-auto}{\huggingface} \href{https://github.com/lmarena/arena-hard-auto}{\faGithub} \\
\midrule
\rowcolor{sectionblue}\multicolumn{6}{c}{\textbf{(e) Long Context}} \\
\midrule
AA-LCR \citep{artificialanalysis2025lcr} & Open-end QA & LLM-as-judge & Agent & A2 & \href{https://huggingface.co/datasets/ArtificialAnalysis/AA-LCR}{\huggingface} \\
RULER \citep{hsieh2024ruler} & Retrieval, Variable Tracking, QA & Task-specific accuracy & Agent & A2 & ~\href{https://github.com/NVIDIA/RULER}{\faGithub} \\
OpenAI MRCR \citep{openai2024mrcr} & Long context needle retrieval & Retrieval accuracy & Agent & A2 & \href{https://huggingface.co/datasets/openai/mrcr}{\huggingface} \\
\midrule
\rowcolor{sectionblue}\multicolumn{6}{c}{\textbf{(f) Tool Use \& Agents}} \\
\midrule
BrowseComp \citep{wei2025browsecomp} & Web search and reasoning & Answer accuracy & Agent / Tool & A2/T2 & \href{https://huggingface.co/datasets/smolagents/browse_comp}{\huggingface} \\
BrowseComp-Plus \citep{chen2025browsecomp} & Web search and reasoning & Answer accuracy & Agent / Tool & A2/T2 & \href{https://huggingface.co/datasets/Tevatron/browsecomp-plus}{\huggingface} \\
MTEB Retrieval \citep{muennighoff2023mteb} & Retrieval over large corpora & Retrieval metrics (nDCG / Recall) & Tool & A1/T1 & \href{https://huggingface.co/spaces/mteb/leaderboard}{\huggingface} \\
Terminal Bench \citep{tbench_2025} & Terminal-based task execution & Task completion rate & Agent / Tool & A1/A2 & \href{https://huggingface.co/datasets/ia03/terminal-bench}{\huggingface} \href{https://github.com/laude-institute/terminal-bench}{\faGithub} \\
$\tau$-Bench \citep{yao2025taubench} & Multi-tool task execution & Success rate & Agent / Tool & A2/T2 & ~\href{https://github.com/sierra-research/tau-bench}{\faGithub} \\
$\tau^2$-Bench \citep{barres2025tau} & Multi-tool task execution & Success rate & Agent / Tool & A2/T2 & \href{https://huggingface.co/datasets/HuggingFaceH4/tau2-bench-data}{\huggingface} \href{https://github.com/sierra-research/tau2-bench}{\faGithub} \\
BFCL \citep{patil2025the} & Function calling / tool invocation & Accuracy & Tool & A1/T1 & ~\href{https://github.com/speakleash/gorilla-bfcl-benchmark}{\faGithub} \\

\end{longtable}
\normalsize

\paragraph{Summary and paradigm gaps.}
Table~\ref{tab:benchmarks} provides a comprehensive listing of standard benchmarks organized by capability category, with an explicit \textbf{Paradigm} column indicating which adaptation paradigm(s) each benchmark most naturally evaluates.
The mapping reveals a clear asymmetry: A1 and A2 benchmarks are relatively mature, with well-established evaluation protocols for coding, reasoning, and retrieval.
T1 evaluation is implicitly covered by standalone tool benchmarks.
T2 evaluation, however, remains ad hoc, typically performed as an ablation within individual papers rather than through standardized community benchmarks.
No existing benchmark suite supports controlled comparison across all four paradigms on the same task distribution, making it impossible to answer questions such as ``for this task, is it better to adapt the agent (A2) or the tool (T2)?'' under matched conditions.

A structural reason for this gap is that most benchmarks assume a fixed evaluation harness tightly coupled to a specific agent interface, creating high integration overhead and test-production mismatch.
The Agentified Agent Assessment (AAA) framework~\citep{agentbeats2025aaa} proposes a different architecture: specialized \emph{assessor agents} that issue tasks, collect results, and compute metrics, communicating with assessee agents through open protocols (A2A for task management, MCP for tool access).
By decoupling the evaluation logic from the agent's internal architecture, protocol-based assessment could in principle enable any compliant agent to participate in any evaluation without custom integration---a prerequisite for the cross-paradigm comparisons that current benchmarks cannot support.
Whether this approach can deliver on its promise at scale remains to be validated empirically.
We return to these structural gaps in \S\ref{subsec:eval_discussion}.

\subsection{Evaluating the Adaptation Signal}
\label{subsec:eval_signal}

The first question in evaluating adapted agentic systems is what the evaluation signal actually measures.
The adaptation signal dictates which failures are diagnosable, which objectives are optimizable, and which pathologies remain hidden.
We distinguish two families of evaluation metrics, aligned with the two signal types in our taxonomy, and ground each in concrete empirical evidence.

\subsubsection{Verifiable Execution Metrics (A1 \& T1)}
\label{subsubsec:verifiable_metrics}

Verifiable execution metrics provide grounded, causal, and process-oriented feedback: the signal comes from an environment or tool with fixed semantics, independent of how the agent internally represents the task.

\paragraph{Execution-based code evaluation.}
The best-understood form of verifiable evaluation is execution-based code assessment, where generated code is run against test suites and success is measured by pass@$k$ or test-suite pass rate~\citep{chen2021evaluating, austin2021program, jimenez2024swebench, jain2025livecodebench}.
The protocol provides a binary, unambiguous signal: the code either passes or fails.
Its strength lies in eliminating subjective judgment; its limitation is that it cannot assess code quality, efficiency, or maintainability beyond functional correctness.
The density of this signal has proven critical for A1-style RL.
RLEF~\citep{gehring2025rlefgroundingcodellms} shows that multi-turn code execution feedback supports stable PPO training.
Code-R1~\citep{code-r1} shows that reward quality (clean, verified test suites) matters more than data quantity for effective code RL.
R1-Code-Interpreter~\citep{chen2025r1} finds that task heterogeneity causes sparse and unstable rewards, necessitating curriculum-based scheduling to stabilize training across diverse code-execution domains.

\paragraph{Retrieval and information-access metrics.}
For retrieval-augmented systems, verifiable metrics include Recall@$K$, nDCG, and MAP computed against gold-standard relevance judgments~\citep{muennighoff2023mteb, thakur2021beir}.
These metrics directly evaluate the tool's output quality and serve as the primary training signal in A1-style adaptation across diverse domains: DeepRetrieval~\citep{jiang2025deepretrieval} uses Recall@$K$ for search, Rec-R1~\citep{lin2025rec} uses NDCG and Recall for recommendation, and SQL-R1~\citep{ma2025sql} uses execution accuracy for database querying.
An advantage of these metrics is that they decompose into per-query scores, which permits fine-grained diagnosis of failure modes (e.g., query types where recall is low, SQL patterns that fail execution).
However, as BGM~\citep{ke2024bridging} shows, high retrieval recall does not guarantee high downstream utility: there exists a systematic ``preference gap'' between what retrievers optimize for (surface-level relevance) and what LLMs find useful for reasoning (contextual coherence, inferential support).

\paragraph{Formal verification.}
In domains such as theorem proving~\citep{jiang2023draft, ren2025deepseek, wang2025kimina} and program synthesis with specifications, formal verifiers provide the strongest possible execution signal: a proof either type-checks or it does not.
Formal verification eliminates evaluation noise entirely and enables step-wise semantic verification, substantially easing long-horizon credit assignment compared to code-execution RLVR where unit tests may be sparse or incomplete.
Recent systems such as AlphaProof~\citep{hubert2025olympiad} and DeepSeek-Prover-V2~\citep{ren2025deepseek} use this verifier feedback to train multi-step proof search policies via RL, confirming the value of dense, deterministic execution signals.
However, formal verification is limited by the availability of formal specifications: it applies only to domains where tasks can be expressed as type-checkable propositions or executable test suites, excluding the broad class of specification-free agentic tasks (those where success criteria are implicit, subjective, or context-dependent) that constitute the majority of real-world agent deployments.

\paragraph{Strengths and limitations.}
Verifiable execution metrics offer dense, reliable, and reproducible feedback, making them well suited for A1 adaptation where the goal is to sharpen tool-use mechanics.
However, they are local: a correct retrieval query or a passing test case does not guarantee that the agent's overall reasoning is sound.
An agent may learn to game execution metrics (e.g., generating trivially passing tests) without improving genuine task-solving capability~\citep{amodei2016concrete}.
The Holistic Agent Leaderboard~\citep{kapoor2025hal} has documented cases where agents achieve high execution scores by searching for benchmark answers online rather than solving tasks, a gaming behavior invisible to standard execution metrics.

\subsubsection{Holistic Utility Metrics (A2 \& T2)}
\label{subsubsec:holistic_metrics}

Holistic utility metrics evaluate the end-to-end quality of the agent's final output, integrating the effects of reasoning, tool use, and synthesis.

\paragraph{Answer-correctness metrics.}
The simplest holistic metric is exact-match (EM) or F1 score on the agent's final answer, as used in multi-hop QA benchmarks~\citep{cobbe2021training, rein2024gpqa}.
For mathematical reasoning, recent work has proposed verification-based evaluation such as \textit{Math-Verify}\footnote{\url{https://github.com/huggingface/Math-Verify}}, which validates mathematical equivalence rather than string identity, reducing false negatives from equivalent but non-identical solutions.
A subtle limitation is that EM-based evaluation conflates reasoning quality with answer extraction: an agent may reason correctly but format its answer incorrectly, or produce the right answer through flawed reasoning (shortcut learning), as discussed in the A2 signal analysis in \S\ref{sec:comparison_a2}.

\paragraph{LLM-as-judge.}
For open-ended tasks where ground-truth answers are unavailable or insufficient, LLM-based evaluation has become prevalent.
Benchmarks such as \textit{Arena-Hard}~\citep{li2025from} and \textit{MT-Bench}~\citep{zheng2023judging} use strong language models to perform pairwise comparisons or assign quality scores.
While scalable, LLM-as-judge introduces systematic biases: verbosity preference (longer responses rated higher regardless of quality), position bias (preference for the first or second response in pairwise comparison), self-enhancement bias (models rating their own outputs higher), and sycophancy (confirming rather than correcting the evaluatee)~\citep{li2025from, zheng2023judging}.
For agentic evaluation, these biases interact with tool use in non-obvious ways: an agent that produces verbose reasoning traces with many tool calls may receive inflated scores even when its tool use is inefficient or redundant.
Calibration against human judgments remains essential, and recent work on judge reliability~\citep{kapoor2025hal} suggests that ensemble judging (multiple LLM judges with aggregation) can partially mitigate individual biases.

\paragraph{Task-completion and functional assessment.}
For agent benchmarks in interactive environments, holistic evaluation often takes the form of task-completion rate assessed through execution-based functional checks.
\textit{WebArena}~\citep{zhou2024webarena} verifies whether the agent achieved the intended web-browsing goal by checking the final page state.
\textit{OSWorld}~\citep{xie2024osworld} uses screenshot-based verification and system-state checks.
\textit{AgencyBench}~\citep{li2026agencybench} combines Docker-sandboxed functional assessment with user-simulation agents that provide iterative feedback, enabling evaluation of both the final outcome and the agent's ability to incorporate feedback.
These approaches bridge the gap between holistic and verifiable evaluation: the metric is holistic (did the agent complete the task?) but the assessment mechanism is verifiable (programmatic state checking).

\paragraph{Strengths and limitations.}
Holistic metrics directly measure what users care about---whether the system solved their problem---and are therefore well suited for A2 and T2 adaptation.
However, they suffer from credit-assignment opacity: when the final answer is wrong, it is unclear whether the failure originated in reasoning, tool selection, tool execution, or synthesis.
Credit-assignment opacity is especially problematic for T2 evaluation, where the goal is to assess the marginal contribution of a specific tool to the frozen agent's performance.
Holistic metrics are also typically sparse (one signal per episode), making them less informative for diagnosing intermediate failures compared to the dense feedback available from verifiable execution metrics.
Sparsity creates a concrete training challenge visible across multiple domains.
In retrieval-augmented QA, Search-R1~\citep{jin2025search} requires roughly 170k training examples under sparse holistic rewards, while A1 methods with dense execution rewards train on far less data.
In code generation, ReTool~\citep{feng2025retool} reports that integrating real-time execution feedback (dense A1 signal) into RL rollouts substantially accelerates convergence compared to end-to-end training with only final-answer rewards.
The pattern (dense signals accelerating learning) is consistent across domains and directly attributable to signal density.

\subsection{Evaluating the Adaptation Dynamics}
\label{subsec:adapt_dynamics}

The metrics discussed in \S\ref{subsec:eval_signal} evaluate the quality of an adapted system at a single point in time.
Yet most evaluations of agentic adaptation report only endpoint metrics (e.g., success rate, pass@$k$, EM, or average return), which erase the process by which agents and tools learn.
The central methodological gap in current agentic evaluation is that many of the most consequential phenomena in agentic adaptation (abrupt phase transitions, reward-likelihood divergence, tool-use drift, and reward-hacking episodes) are \emph{dynamical} in nature and invisible to any single-point measurement.
Endpoint-equivalent methods can diverge sharply in stability, data requirements, and safety trajectories, as illustrated in the subsections below and in \S\ref{subsec:eval_discussion}.
The choice of signal type (\S\ref{subsec:eval_signal}) also shapes the observable dynamics: verifiable execution metrics (A1/T1) produce dense, per-step feedback that yields smooth learning curves, while holistic utility metrics (A2/T2) produce sparse, per-episode signals that can mask intermediate instabilities.
We organize dynamics-aware evaluation along three axes: efficiency, generalization, and stability.

\subsubsection{Sample and Interaction Efficiency}
\label{subsubsec:sample_efficiency}

For any adaptation method, a basic question is how much data, compute, and interaction it requires to reach a target performance level.
The question is pressing in agentic settings, where each ``sample'' may involve multiple tool calls, environment interactions, and thousands of tokens.

\paragraph{Data efficiency.}
The most direct comparison is the number of training examples required to reach a given accuracy.
As established quantitatively in \S\ref{sec:comparison_synthesis}, paradigm choice has a large effect on data requirements, but the direction of the effect depends on domain:

\begin{itemize}[leftmargin=1.2em]
    \item \textbf{T2 vs.\ A2 (retrieval):} As quantified in \S\ref{sec:comparison_synthesis}, \textsc{s3} (T2) reaches comparable accuracy to Search-R1 (A2) with roughly $70\times$ fewer training examples, a gap attributable to the narrow procedural skill the T2 subagent must learn.
    \item \textbf{A1 (code):} RLEF~\citep{gehring2025rlefgroundingcodellms} achieves efficient training because dense execution rewards (pass/fail per test case) provide rich per-step feedback. Here A1 is more efficient than A2, reversing the retrieval pattern.
    \item \textbf{A1 (multi-tool):} Self-Challenging Agents~\citep{zhou2025self} achieve a $2\times$ improvement by generating their own training curriculum, showing that data generation strategy can be as important as data volume.
\end{itemize}

\noindent These cross-domain comparisons reveal that efficiency is not a fixed property of a paradigm but depends on the interaction between paradigm, signal density, and task structure.

\paragraph{Interaction efficiency.}
Beyond data volume, the number of environment interactions (tool calls, API requests, browsing steps) per episode is a critical efficiency metric.
Recent work has shown that interaction efficiency varies across model sizes and tool-latency regimes: smaller models may achieve higher task success under tight time budgets by executing more frequent but cheaper tool calls, while larger models benefit from fewer but higher-quality interactions~\citep{ma2026timelymachine}.
Evaluations should report not only final success rate but also the distribution of interaction counts (mean, variance, and tail behavior), as high-variance interaction patterns may indicate unstable exploration.

\paragraph{Compute efficiency.}
Total tokens consumed (both prompt and completion), wall-clock training time, and inference-time compute (tokens per task at deployment) are all relevant efficiency metrics.
The distinction between training-time and inference-time compute is important for agentic systems, where test-time compute scaling (e.g., longer reasoning chains, more retrieval rounds) is itself a learned behavior.
Comparing methods fairly requires Pareto frontiers of accuracy versus compute.
T2 methods achieve strong efficiency by training only lightweight subagents: AgentFlow~\citep{li2025flow} trains a 7B planner to achieve 33.1\% on GAIA, outperforming GPT-4.
A1 methods occupy a different region of the Pareto frontier: DeepRetrieval~\citep{jiang2025deepretrieval} trains efficiently due to dense rewards but optimizes only a single tool skill, while Orion~\citep{vijay2025think} shows that effective multi-step search can be learned with compact 350M--1.2B models.
A2 methods, which update the full agent, incur the highest compute cost but offer the broadest capability improvements.
These comparisons are only meaningful when both data and compute are reported jointly.

\subsubsection{Generalization and Robustness}
\label{subsubsec:generalization}

Adaptation is only valuable if the resulting agent-tool system performs well beyond its training distribution.
Several dimensions of generalization are important for agentic systems; each is grounded below in available empirical evidence.

\paragraph{Cross-task generalization.}
An agent adapted for one task (e.g., single-hop retrieval) may or may not transfer to related tasks (e.g., multi-hop reasoning).
Empirical evidence reveals paradigm-dependent transfer patterns across multiple domains.
In retrieval, \textsc{s3} (T2) trained on general QA achieves 76.6\% on specialized medical QA versus 71.8\% for Search-R1 (A2)~\citep{jiang2025s3}, suggesting that T2's frozen-agent architecture preserves broad reasoning capabilities.
In multi-tool settings, Agent-R~\citep{yuan2025agent} shows that MCTS-based self-reflection improves performance by 5.6\% across diverse interactive environments, indicating that A2-style strategic learning can transfer across task types when the training signal captures high-level reasoning patterns.
Conversely, A1 methods that optimize narrow tool-use mechanics (e.g., query reformulation for a specific retrieval interface) may overfit to those interfaces, consistent with the specialization-generalization trade-off documented in \S\ref{sec:comparison}.
Evaluations should include held-out task categories to measure transfer, and should report cross-domain performance alongside in-domain performance.

\paragraph{Cross-agent generalization (for T1/T2 tools).}
A tool adapted under T2 supervision from one frozen agent should ideally remain useful when paired with a different agent.
\textsc{s3} provides preliminary evidence: the same trained searcher improves performance when paired with both Qwen2.5-14B and Claude as frozen generators~\citep{jiang2025s3}, suggesting partial cross-agent transfer.
However, systematic evaluations that vary the frozen agent (e.g., different model families, sizes, and instruction-tuning regimes) and measure whether the adapted tool maintains its marginal benefit remain rare and are critical for modular system design.

\paragraph{Robustness to distribution shift.}
Agentic systems encounter diverse and unpredictable inputs in deployment.
Robustness evaluation should include: (i) adversarial or out-of-distribution queries, (ii) degraded tool performance (e.g., noisy retrieval, flaky APIs), and (iii) environment non-stationarity (e.g., changing web layouts, updated codebases).
Benchmarks like \textit{The Tool Decathlon}~\citep{li2025toolathlon}, which tests agents across 32 diverse software applications with realistic initial states, begin to address this need for text-based agents.

\paragraph{Multimodal and multi-agent robustness.}
A gap in current evaluation is the limited treatment of multimodal and multi-agent robustness.
Multimodal agent benchmarks such as \textit{VisualWebArena}~\citep{koh2024visualwebarena} and \textit{OSWorld}~\citep{xie2024osworld} require agents to process visual inputs (screenshots, page layouts) alongside text, introducing modality-specific distribution shifts (e.g., UI redesigns, resolution changes) that text-only robustness evaluations cannot capture.
Multi-agent systems~\citep{hong2023metagpt, qian2024chatdev} face additional robustness challenges: inter-agent distribution shift (when one agent's adaptation changes the effective environment for others), communication protocol fragility, and emergent coordination failures under novel task distributions.
Evaluating robustness in these settings requires benchmarks that systematically vary both the modality and interaction structure of the evaluation environment, a capability that current suites largely lack.

\subsubsection{Continual and Co-Adaptation Stability}
\label{subsubsec:continual_stability}

Real-world agentic systems must adapt continuously as tasks, tools, and user needs evolve.
Continuous adaptation introduces evaluation challenges that go beyond single-round assessment; recent memory-centric surveys~\citep{hu2025memorySurvey} highlight that memory formation, evolution, and retrieval dynamics are themselves measurable axes of long-horizon agent behavior.
We concentrate here on \emph{measuring} stability phenomena; the underlying mechanisms and mitigation strategies are discussed in \S\ref{subsec:co-adapt} and \S\ref{subsec:continual_adapt}.

\paragraph{Measuring catastrophic forgetting.}
When an agent is adapted to a new task or domain, it may lose performance on previously mastered tasks.
The standard protocol is to maintain a held-out ``retention set'' of previously solved tasks and track performance throughout the adaptation process.
Empirical evidence suggests that the choice of adaptation paradigm affects forgetting: RL-based adaptation can exhibit less forgetting than SFT under certain conditions~\citep{chen2025retaining}, and T2-style modular adaptation structurally avoids forgetting in the core agent by keeping it frozen.
Evaluations should report backward transfer (change in performance on old tasks after learning new ones) alongside forward performance, enabling direct comparison of paradigm-level forgetting profiles.

\paragraph{Measuring co-adaptation stability (open problem).}
When both the agent and its tools are adapted simultaneously, the system may exhibit non-stationary dynamics: the agent adapts to a tool that is itself changing, potentially leading to oscillations, divergence, or degenerate equilibria.
To our knowledge, no existing work has systematically measured co-adaptation stability in agentic systems; the phenomenon is well-studied in multi-agent RL~\citep{ning2024survey} but has not been formalized for the agent-tool setting.
Stability evaluation should track joint performance trajectories over training steps and detect pathological patterns such as cycling (periodic performance oscillations) or collapse (mutual degradation).
Candidate metrics include: (i) the variance of the joint performance trajectory over a sliding window, (ii) the frequency of sign changes in the performance gradient, and (iii) convergence rate to a stable equilibrium.
\textit{These metrics are theoretical propositions that require future empirical validation}; we include them here to delineate the measurement problem, not to claim established practice.
Developing standardized protocols for measuring these dynamics, and validating them on concrete agent-tool co-training runs, is a critical open challenge for the field.

\paragraph{Entropy dynamics as a diagnostic.}
Recent analyses have identified policy entropy as a first-class diagnostic for adaptation stability.
Cui et al.\ document a consistent early-stage entropy collapse across model families and RL variants, where most performance gains coincide with rapid entropy depletion, implying a predictable performance ceiling once entropy is exhausted~\citep{cui2025entropy}.
Complementarily, Hao et al.\ argue that the relevant quantity for stability is not entropy itself but entropy change per update, which can be amplified by naive interventions~\citep{hao2025rethinking}.
Beyond entropy, a comprehensive dynamics-aware evaluation toolkit should also monitor: (i) the reward curve shape (smooth convergence vs.\ sudden jumps indicating phase transitions), (ii) tool-call frequency and diversity over training (detecting mode collapse in tool use), and (iii) KL divergence from the reference policy (detecting excessive drift).
An emerging best practice is to log both $H(\pi_{\theta_t})$ (the entropy of the agent's action distribution at training step $t$, which quantifies the breadth of the agent's exploration) and $\Delta H = H(\pi_{\theta_{t+1}}) - H(\pi_{\theta_t})$ (the per-update entropy change), and relate both to reward gains and behavioral shifts.
Together, these two quantities provide a lightweight but informative diagnostic of adaptation health: rapid entropy depletion signals premature convergence, while large $|\Delta H|$ fluctuations signal training instability.

\subsection{Systemic Evaluation}
\label{subsec:systemic_eval}

Beyond task performance and adaptation dynamics, deployed agentic systems must satisfy systemic requirements related to cost, safety, and alignment.
Cost, safety, and alignment are often orthogonal to accuracy yet can dominate deployment decisions.

\subsubsection{Cost and Inference-Time Compute}
\label{subsubsec:cost}

\paragraph{Token and step cost.}
Agentic tasks consume far more tokens than standard LLM queries due to multi-turn tool interactions, long reasoning chains, and context accumulation.
\textit{AgencyBench}~\citep{li2026agencybench} reports that realistic agentic tasks require an average of one million tokens and 90 tool calls, with execution times measured in hours.
Cost evaluation should decompose total expenditure into: (i) prompt tokens (context provided to the agent), (ii) completion tokens (agent-generated reasoning and actions), (iii) tool-interaction overhead (API latency, execution time), and (iv) retry and error-recovery costs.
The decomposition is essential for identifying which component dominates cost and where optimization effort should be directed.

\paragraph{Inference-time compute trade-offs.}
A distinctive feature of adapted agentic systems is that test-time compute allocation is itself a learned behavior: the agent decides how long to reason, how many tools to invoke, and when to stop.
Accuracy and cost therefore trade off in ways that no single metric can capture.
Recent work has shown that this trade-off interacts non-trivially with tool latency: when tool calls are fast, smaller models can achieve higher task success by executing more interactions within a fixed time budget; when tool calls are slow, larger models dominate by producing higher-quality plans with fewer interactions~\citep{ma2026timelymachine}.
For paradigm selection, this observation suggests that T2 methods, which use lightweight subagents for tool operations, may hold an advantage in high-latency tool environments where each tool call is costly, though this hypothesis has not yet been empirically validated in a controlled cross-paradigm comparison.
Evaluations should report cost-conditioned performance curves (accuracy as a function of token budget or wall-clock time) rather than unconstrained accuracy alone.

\paragraph{Training cost across paradigms.}
The cost of adaptation itself varies across paradigms and domains, and this variation is one of the strongest empirical signals in the current literature.
A2-style end-to-end agent training requires large-scale RL (e.g., Search-R1 uses $\sim$170k examples~\citep{jin2025search}; R1-Searcher reports similar scale~\citep{song2025r1}).
T2-style tool adaptation achieves competitive performance at much lower cost (e.g., AgentFlow trains only a 7B planner~\citep{li2025flow}).
A1 methods span a wide cost range depending on signal density: Code-R1~\citep{code-r1} shows that investing in reward quality (clean test suites) reduces the total training budget more effectively than scaling data, while ToolExpander~\citep{chen2025toolexpander} shows that dynamic hard-sample replacement can stabilize training for resource-constrained models.
T1 tool training (e.g., fine-tuning a retriever or embedding model) is typically the cheapest paradigm, as it uses standard supervised learning on curated datasets without requiring agent interaction; its cost is well-understood and dominated by dataset curation rather than compute.
Evaluations should report training cost (GPU hours, total training tokens) alongside performance, enabling Pareto-optimal comparisons across paradigms.

\subsubsection{Safety}
\label{subsubsec:safety}

Adaptation introduces dynamic safety risks that go beyond the static alignment of frozen models.
Three evaluation dimensions are specific to adapted agentic systems; mitigation strategies are discussed in \S\ref{subsec:safe_adapt}.

\paragraph{Unsafe exploration.}
On-policy RL adaptation (A1/A2) requires agents to explore novel action sequences, which may include dangerous tool invocations (e.g., deleting files, executing arbitrary code, making irreversible API calls).
Evaluating exploration safety requires sandboxed environments that can detect and log unsafe actions without allowing real-world harm.
\textit{ToolEmu}~\citep{ruan2024toolemu} provides an LM-emulated sandbox for identifying risks of LM agents with tool use, enabling scalable safety evaluation without requiring real tool backends.
\textit{R-Judge}~\citep{yuan2024rjudge} benchmarks safety risk awareness for LLM agents, evaluating the agent's ability to identify and refuse unsafe tool-use requests across diverse risk categories.

\paragraph{Reward hacking and specification gaming.}
Adapted agents may learn to exploit imperfections in the reward signal rather than genuinely solving tasks, a risk that grows with agent capability~\citep{amodei2016concrete, fu2025reward}.
The Holistic Agent Leaderboard~\citep{kapoor2025hal} has shown the value of LLM-assisted log inspection for detecting previously unreported gaming behaviors, such as agents searching for benchmark answers online rather than solving tasks.
Evaluation should include held-out test sets that differ from the training reward distribution, human spot-checks of high-reward trajectories, and automated detection of known gaming patterns (e.g., trivially passing self-generated tests, manipulating evaluation state).
The susceptibility to reward hacking is paradigm-dependent:

\begin{itemize}[leftmargin=1.2em]
    \item \textbf{A1:} Dense execution rewards (pass/fail, recall) are harder to game because they are grounded in deterministic tool output.
    \item \textbf{A2:} Sparse holistic rewards (final-answer EM, LLM-as-judge) are more susceptible because the agent has more degrees of freedom to satisfy the metric without genuine task solving.
    \item \textbf{T2:} The tool may learn to produce outputs that inflate the frozen agent's score on the training distribution without improving genuine utility (a form of tool-level reward hacking).
\end{itemize}

\paragraph{Safety degradation under adaptation.}
Aggressive RL optimization for reasoning can erode safety guardrails established during supervised fine-tuning; DeepSeek-R1~\citep{guo2025deepseek} demonstrates this pattern, as the model learns to reason around refusal mechanisms (see also \S\ref{subsec:safe_adapt} for mitigation strategies).
A practical protocol is to maintain a safety benchmark (e.g., a set of harmful-request prompts) and evaluate at regular training checkpoints, plotting a safety-performance trajectory that reveals whether gains in task accuracy come at the cost of safety degradation.
Safety-performance trajectories are a natural application of the dynamics-aware evaluation principle established in \S\ref{subsec:adapt_dynamics}.

\subsubsection{Alignment}
\label{subsubsec:alignment}

Alignment in agentic systems operates at multiple levels that must be evaluated jointly.

\paragraph{Human-agent alignment.}
The adapted agent should faithfully execute user intent, follow instructions, and respect stated preferences.
Benchmarks such as \textit{IFEval}~\citep{zhou2023instruction} and \textit{IFBench}~\citep{pyatkin2025generalizing} evaluate instruction compliance through rule-based scoring.
For more open-ended alignment, preference-based evaluation using human judgments or calibrated LLM judges remains the gold standard~\citep{li2025from}.
A concern is that adaptation may improve task performance while degrading alignment: an agent optimized for answer correctness may become less responsive to user constraints or stylistic preferences. The trade-off should be explicitly measured by evaluating alignment benchmarks at each adaptation checkpoint.

\paragraph{Agent-tool alignment.}
In modular systems with adapted tools (T2), the tool must be aligned with the agent's reasoning style and information needs.
Misalignment manifests as the tool providing information in a format the agent cannot effectively use, or the agent issuing tool calls that the tool cannot meaningfully process.
BGM~\citep{ke2024bridging} provides concrete evidence of this phenomenon: a 38\% relative improvement in downstream QA accuracy comes from training a ``bridge model'' that translates retrieval output into a format the frozen LLM finds useful, confirming that format alignment between tool output and agent consumption is a measurable and optimizable quantity.
Evaluating agent-tool alignment requires measuring not only end-to-end task success but also intermediate interaction quality: Are the agent's tool calls well-formed? Does the tool's output contain the information the agent needs? Is the information presented in a format the agent can parse?
Autonomous evaluation frameworks that enable agents to self-assess and refine their own trajectories~\citep{pan2024autonomous} offer a complementary path toward scalable, fine-grained interaction diagnostics.

\begin{table*}[t]
    \centering
    \caption{Recommended evaluation dimensions and representative metrics for each adaptation paradigm. 
    \checkmark\ indicates primary relevance; (\checkmark) indicates secondary relevance.}
    \label{tab:eval_paradigm}
    \begin{adjustbox}{max width=\textwidth}
    \begin{tabular}{@{}llp{7cm}cccc@{}}
    \toprule
    \textbf{Dimension} & \textbf{Sub-dimension} & \textbf{Representative Metrics} & \textbf{A1} & \textbf{A2} & \textbf{T1} & \textbf{T2} \\
    \midrule
    & Verifiable execution 
    & pass@$k$, Recall@$K$, nDCG, proof check 
    & \checkmark & & \checkmark & \\
    \multirow{-2}{*}{Signal Quality} 
    & Holistic utility 
    & EM, F1, LLM-as-judge, task completion 
    & & \checkmark & & \checkmark \\
    \addlinespace[4pt]
    & Data / compute efficiency 
    & Learning curve, Pareto frontier 
    & \checkmark & \checkmark & (\checkmark) & \checkmark \\
    & Generalization 
    & Cross-task, cross-agent transfer 
    & (\checkmark) & \checkmark & \checkmark & \checkmark \\
    \multirow{-3}{*}{Dynamics} 
    & Stability / forgetting 
    & Backward transfer, entropy trajectory 
    & (\checkmark) & \checkmark & & (\checkmark) \\
    \addlinespace[4pt]
    & Cost 
    & Tokens, wall-clock time, training GPU-hrs 
    & \checkmark & \checkmark & (\checkmark) & \checkmark \\
    & Safety 
    & Unsafe exploration, reward hacking, safety trajectory 
    & \checkmark & \checkmark & & (\checkmark) \\
    \multirow{-3}{*}{Systemic} 
    & Alignment 
    & Instruction compliance, agent-tool format match 
    & & \checkmark & & \checkmark \\
    \bottomrule
    \end{tabular}
    \end{adjustbox}
\end{table*}

\subsection{Discussion}
\label{subsec:eval_discussion}

Table~\ref{tab:eval_paradigm} summarizes the recommended evaluation dimensions and metrics for each adaptation paradigm, synthesizing the analysis from the preceding subsections.
The following discussion addresses cross-cutting themes that emerge from applying the A1/A2/T1/T2 taxonomy to evaluation, and identifies concrete gaps that the field must address.

\subsubsection{Concrete Illustrations of the Dynamics Gap}

The central argument of \S\ref{subsec:adapt_dynamics}---that endpoint metrics are insufficient---becomes concrete through the following examples.
Three phenomena, drawn from the preceding subsections, illustrate what dynamics-aware evaluation reveals in practice:

\begin{itemize}[leftmargin=1.2em]
    \item \textbf{Hidden divergence in efficiency.} A2 and T2 methods that achieve similar final accuracy on retrieval-augmented QA (\S\ref{sec:comparison_synthesis}) differ by $70\times$ in data requirements (\S\ref{subsubsec:sample_efficiency}), a distinction invisible to any endpoint leaderboard.
    Similarly, two code-generation agents with identical pass@$k$ may exhibit very different entropy trajectories~\citep{cui2025entropy}, one near entropy exhaustion and the other retaining exploration capacity.
    \item \textbf{Paradigm-selection artifacts.} The same system appears to favor different paradigms depending on the evaluation metric (\S\ref{subsubsec:verifiable_metrics} vs.\ \S\ref{subsubsec:holistic_metrics}): A1 metrics highlight tool mechanics, A2 metrics highlight strategic reasoning, and neither alone indicates where further investment would yield the greatest return.
    \item \textbf{Non-monotonic safety regression.} DeepSeek-R1~\citep{guo2025deepseek} shows that aggressive RL optimization can temporarily erode safety guardrails before partial restoration (\S\ref{subsubsec:safety}), creating a vulnerability window invisible to endpoint-only safety checks.
\end{itemize}

\subsubsection{Tool-Centric vs.\ Agent-Centric Evaluation}

The distinction between tool-centric and agent-centric evaluation paradigms mirrors a fundamental design choice in agentic systems.

\paragraph{Tool-centric evaluation.} This perspective focuses on the quality and composability of individual tools, measuring intrinsic metrics (retrieval recall, code correctness) that are independent of any specific agent.
Tool-centric evaluation supports modular system design: tools can be evaluated, compared, and swapped independently.
However, tool-centric evaluation cannot capture emergent behaviors that arise from agent-tool interaction, such as the agent learning to compensate for tool weaknesses or to exploit tool strengths in unexpected ways.

\paragraph{Agent-centric evaluation.} This perspective focuses on the end-to-end system performance, measuring holistic outcomes that integrate reasoning, tool use, and synthesis.
Agent-centric evaluation measures the outcome users care about but makes it difficult to attribute performance to specific components.
When an agent-centric benchmark score improves, it is unclear whether the improvement came from better reasoning, better tool use, better tool quality, or a fortunate interaction between these factors.

\paragraph{Bridging the gap: counterfactual evaluation.}
A rigorous approach to bridging this gap is counterfactual evaluation: systematically varying one component (e.g., replacing the adapted tool with a baseline) while holding others fixed, to isolate marginal contributions.
Counterfactual evaluation is already standard in T2 evaluations: \textsc{s3} reports the frozen generator's accuracy with and without the adapted searcher~\citep{jiang2025s3}, and QAgent~\citep{jiang2025qagent} shows how switching from self-evaluation to frozen-generator evaluation corrects reward hacking.
We argue that counterfactual evaluation should be adopted as a standard reporting requirement for all paradigms: A1 papers should report performance with and without the adapted tool mechanic; A2 papers should ablate tool use to isolate reasoning improvements; and integrated benchmarks should support component-level swap-in/swap-out evaluation.
A caveat: counterfactual evaluation assumes that components are approximately independent, but in practice, agents adapt their behavior in response to tool quality.
Replacing an adapted tool with a baseline may cause the agent to behave differently than it would have if trained with that baseline from the start, introducing a confound that pure swap-in evaluation cannot resolve.
The limitation motivates complementary approaches such as progressive ablation (gradually degrading tool quality during evaluation) to measure sensitivity rather than assuming clean separability.

\subsubsection{How Evaluation Reshapes Adaptation Design}

The choice of evaluation protocol has direct implications for which adaptation strategies are incentivized.

\paragraph{Metric-driven optimization.}
When benchmarks use execution-based metrics (pass@$k$, retrieval recall), A1-style methods that directly optimize these metrics have a natural advantage.
When benchmarks use holistic metrics (final-answer EM, LLM-as-judge), A2-style methods that optimize end-to-end performance are favored.
The risk is benchmark co-adaptation: methods evolve to exploit the specific evaluation protocol rather than to genuinely improve agentic capability.
For example, agents trained on code benchmarks with pass@$k$ may learn to generate code that passes tests but is unreadable, unmaintainable, or inefficient, optimizing the metric while degrading unmeasured quality dimensions.

\paragraph{Reporting standards for agentic adaptation.}
The RL community has developed concrete reporting standards~\citep{henderson2018deep, agarwal2021deep}: mandatory learning curves, confidence intervals across seeds, and hyperparameter sensitivity analyses.
Agentic adaptation papers should adopt an analogous standard: (i) learning curves with at least three random seeds, (ii) cost-conditioned performance (accuracy vs.\ token budget), (iii) retention-set performance for continual settings, and (iv) safety-trajectory plots when RL is involved.
Venues that enforce such standards will incentivize methods that are genuinely robust rather than merely endpoint-optimal.

\paragraph{Evaluation as a design constraint.}
Conversely, well-designed evaluation protocols can steer adaptation research toward desirable properties.
Benchmarks that jointly evaluate accuracy, cost, and safety (rather than accuracy alone) incentivize methods that achieve good trade-offs across all dimensions.
Time-budgeted evaluation~\citep{ma2026timelymachine}, which measures accuracy under wall-clock-time constraints, incentivizes efficient tool use and penalizes wasteful exploration.
Retention-set evaluation, which measures performance on previously solved tasks, incentivizes continual-learning-aware adaptation.
Multi-dimensional leaderboards that display Pareto frontiers across accuracy, cost, safety, and efficiency would be more informative than single-score rankings.

\subsubsection{What Is Missing: Toward Next-Generation Benchmark Suites}

Current benchmarks, despite their rapid proliferation, share several systematic limitations that constrain the evaluation of agentic adaptation.

\paragraph{Static tasks vs.\ dynamic environments.}
The vast majority of benchmarks consist of fixed task sets evaluated in a single pass.
Fixed task sets cannot assess an agent's ability to adapt over time, learn from failures, incorporate new information, or adjust to changing environments.
Next-generation benchmarks should embed agents in persistent, evolving environments where the task distribution shifts over time, tools are updated or replaced, and the agent must continuously adapt to maintain performance.

\paragraph{Single-paradigm evaluation.}
Most benchmarks implicitly evaluate a single adaptation paradigm (typically A2) without providing the infrastructure to compare across paradigms.
A comprehensive benchmark suite for agentic adaptation should support evaluation of all four paradigms on the same task distribution, enabling controlled comparisons of A1 vs.\ A2 vs.\ T1 vs.\ T2 under matched conditions.
In practice, this requires benchmarks that provide: (i) verifiable tool-execution signals (for A1), (ii) holistic task-completion signals (for A2), (iii) agent-agnostic tool evaluation protocols (for T1), and (iv) frozen-agent + variable-tool evaluation protocols (for T2), all on the same underlying tasks.
Protocol-based evaluation frameworks such as AAA~\citep{agentbeats2025aaa}, which standardize agent-assessment communication through open protocols, offer a potential path toward this goal by allowing the same assessor agent to evaluate agents of different architectures under matched conditions.

\paragraph{Multimodal and multi-agent adaptation benchmarks.}
While existing multimodal benchmarks (\textit{VisualWebArena}~\citep{koh2024visualwebarena}, \textit{OSWorld}~\citep{xie2024osworld}) and multi-agent frameworks~\citep{hong2023metagpt, qian2024chatdev} test system-level competence (see the robustness challenges discussed in \S\ref{subsubsec:generalization}), they do not yet support adaptation-specific evaluation.
What is missing are benchmarks that measure how multimodal agents improve their visual grounding over time (e.g., learning to parse new UI layouts) and how multi-agent teams improve their coordination protocols through interaction (e.g., learning role specialization).
These require longitudinal evaluation designs (tracking adaptation trajectories across episodes) that current static benchmarks do not support.
Extending the A1/A2/T1/T2 taxonomy to these settings, where ``tools'' may include other agents and sensory modalities, is a necessary step toward comprehensive evaluation of agentic adaptation.

\paragraph{Missing dimensions.}
Current benchmarks overwhelmingly focus on task accuracy.
A next-generation benchmark suite should simultaneously evaluate \textbf{performance trajectory} (convergence speed, stability, sample efficiency), \textbf{stability and forgetting} (backward transfer on retained tasks), \textbf{adaptation efficiency} (total cost to reach target performance), and \textbf{safety and alignment} (continuous monitoring throughout adaptation).
These dimensions correspond directly to the evaluation axes developed in \S\ref{subsec:adapt_dynamics}--\S\ref{subsec:systemic_eval}; what is missing is benchmark infrastructure that operationalizes them in a unified evaluation harness.

\paragraph{Toward living benchmarks.}
The rapid saturation of existing benchmarks (e.g., GAIA scores approaching human baselines within months of release) suggests that static benchmark suites have a limited shelf life.
Self-evolving benchmarks that dynamically increase task difficulty through automated task generation and validation offer one path forward, ensuring that the evaluation remains challenging as agent capabilities improve.
The self-evolving subagent paradigm (R-Zero~\citep{huang2025rzero}, Multi-Agent Evolve~\citep{chen2025mae}) already shows the feasibility of automated task generation for training; extending this principle to evaluation would yield benchmarks that co-evolve with the systems they measure.

However, living benchmarks introduce their own limitations that must be acknowledged.
First, \textit{evaluation cost} scales continuously: unlike static benchmarks that are evaluated once, a living benchmark requires periodic re-evaluation as the task distribution evolves, multiplying compute expenditure over time.
Second, and more fundamentally, living benchmarks create a \textit{reproducibility crisis}: if the task distribution at time $t_1$ differs from that at time $t_2$, results obtained by researcher A and researcher B at different times are not directly comparable.
Maintaining versioned snapshots of the evolving benchmark can partially address this, but at the cost of reintroducing the static-benchmark problem for each snapshot.
Third, \textit{automated validation} of generated tasks remains an open challenge: generated tasks must be solvable, non-degenerate, and meaningfully discriminative, requiring either formal verifiability (limiting applicability to specification-rich domains) or reliable automated quality filters.
These trade-offs suggest that living benchmarks are best deployed as complements to, rather than replacements for, versioned static benchmarks, with the static snapshots providing reproducible baselines and the evolving component testing continued adaptability.

\section{Applications}
\label{sec:applications_and_existing_works}
Agentic AI systems have been adopted across a growing range of scientific and engineering domains. The following subsections organize representative applications by discipline and connect each to the adaptation paradigms (A1/A2/T1/T2) developed in this paper. We categorize these applications into the following areas: \textbf{General Science}, such as \textit{Deep Research} (\S\ref{subsec:app_deep_research}); \textbf{Computer Science}, where agents augment or automate processes in \textit{Software Development} (\S\ref{subsec:app_software}) and \textit{Computer Use} (\S\ref{subsec:app_computer_use}); and \textbf{Biomedicine}, where agents accelerate research in \textit{drug discovery and development} (\S\ref{subsec:app_drug}). 

Across these domains, the dominant adaptation paradigm varies with the availability of verifiable feedback and the cost of agent retraining. Table~\ref{tab:app_paradigm} profiles each domain by its dominant paradigm, key bottleneck, and a representative system.

\subsection{Deep Research}
\label{subsec:app_deep_research}
Deep research systems automate end-to-end scientific investigation by integrating large language models (LLMs), advanced retrieval, and autonomous reasoning~\cite{xu2025comprehensive}. OpenAI’s DeepResearch~\cite{openaideepresearch} is a prominent example, featuring a multi-step reasoning workflow that conducts iterative search, validation, and synthesis. Similar paradigms have been adopted in recently announced systems such as Claude’s deep-search capabilities~\cite{claudesearch} and Google’s Gemini-based research agents~\cite{geminideepresearch}. Their defining characteristic, compared to general-purpose AI agents, is dual adaptation in both agent reasoning and scientific tool integration.

\paragraph{Agent adaptation.}
Deep research systems require agentic workflows that decompose complex scientific questions into structured research plans: (1) adapting LLMs toward long-context reasoning, hypothesis refinement, and multi-step self-critique, (2) orchestrating multiple agents to collaborate hierarchically for literature review, data interpretation, and conclusion synthesis, and (3) maintaining persistent memory and knowledge tracking across long investigative trajectories.

\paragraph{Tool adaptation.}
To address hallucination and improve informativeness, deep research agents incorporate diverse tools that provide direct access to external knowledge: (1) structured retrieval interfaces to literature databases (e.g., PubMed, arXiv), (2) web navigation tools for interacting with scientific resources, and (3) modular computational utilities for data analysis and visualization. Learning-based retrieval modules such as DeepRetrieval~\cite{jiang2025deepretrieval} and s3~\cite{jiang2025s3} further advance tool adaptation, boosting accuracy in real-time information gathering, especially atop proprietary models that cannot be fine-tuned.

\paragraph{Toward domain-specialized deep research.}
Current systems are primarily built on generic corpora and may struggle with nuanced expert-level inquiries. Extension to specialized scientific fields will involve integrating domain knowledge bases and ontologies, validated bioinformatics and biomedical computation tools, and field-specific safety, reliability, and evaluation protocols.

Such integration would enable deep research systems to function as expert collaborators for vertical domains such as medicine, materials science, and drug development.

\paragraph{Paradigm mapping.}
Deep research exemplifies the complementary use of multiple adaptation paradigms. The core reasoning agent is typically adapted via A2 (optimizing end-to-end research quality), while retrieval and search tools are adapted via T2 (training search subagents like s3~\cite{jiang2025s3} under frozen-agent supervision) or deployed as T1 components (pre-trained dense retrievers). The absence of deterministic execution signals (unlike code compilation) makes pure A1 adaptation less natural in this domain, though retrieval-specific metrics (Recall@$K$) can provide A1-style feedback for the search component.

\subsection{Software Development}
\label{subsec:app_software}
\begin{figure*}[t]
    \centering
    \includegraphics[width=\textwidth]{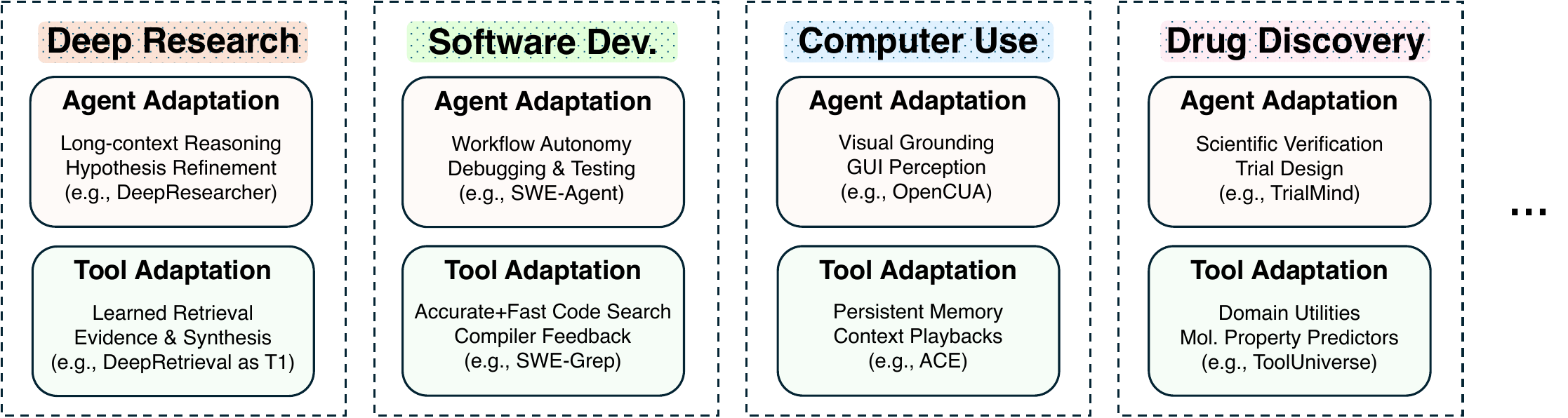}
    \caption{
    Applications of Adaptation in Agentic AI.
    }
    \label{fig:applications}
\end{figure*}
AI-assisted software development is a major application area for agentic AI. Unlike conventional code completion systems, software development agents autonomously navigate multi-stage engineering workflows (requirement interpretation, code generation, debugging, testing, and deployment) within real development environments. Modern AI-assisted development tools and agents such as \textbf{Cursor} \citep{cursor2025}, \textbf{Claude Code} \cite{anthropic2025claudecode}, and OpenAI's \textbf{CodeX}~\citep{openai2025codex} exemplify this shift from passive code completion toward interactive, full-cycle programming capable of understanding project context and performing tool-mediated reasoning. To evaluate these capabilities, the \textbf{SWE-Bench} benchmark~\citep{jimenez2024swebench} serves as a representative testing suite that measures an agent’s ability to autonomously fix real-world software bugs in open-source repositories by reading, editing, and validating code through continuous integration workflows.

Other notable research efforts have also explored the development of autonomous software engineering agents. \textbf{SWE-Agent}~\citep{yang2024swe} introduces an agent-computer interface (ACI) that enables language model agents to autonomously perform end-to-end software engineering tasks, including repository navigation, code modification, and test execution. \textbf{OpenHands}~\citep{wang2025openhands}, an open-source platform for AI software developers, extends this paradigm by providing a sandboxed execution environment and modular evaluation framework for developing and benchmarking general-purpose coding agents.

Building effective software agents requires both \textbf{agent adaptation} (strengthening reasoning, planning, and self-verification across complex development pipelines) and \textbf{tool adaptation} (integrating and evolving the surrounding development ecosystem such as compilers, debuggers, and test frameworks).

\paragraph{Agent adaptation.} Agent adaptation in software development improves model reasoning and autonomy across multi-stage engineering workflows. The availability of deterministic execution signals (test-suite pass rates, compilation success, CI/CD pipeline outcomes) makes software development well suited for \textbf{A1-style adaptation}, where agents are optimized on tool-execution feedback~\citep{gehring2025rlefgroundingcodellms, code-r1, chen2025r1}. A2-style adaptation applies when agents are evaluated on holistic task completion (e.g., resolving a full GitHub issue), which requires reasoning about when to read code, what to edit, and how to validate changes.

\paragraph{Tool adaptation.} Tool adaptation in this domain involves evolving the software ecosystem to improve the reliability, responsiveness, and contextual integration of tools that agents depend on for code execution, testing, and evaluation. A representative example is \textbf{Cursor}'s Tab-RL framework~\citep{cursor2025tabrl}, which applies reinforcement learning to refine the editor's tab completion behavior based on real-world user interactions, aligning the tool's interface dynamics with agent and developer preferences. A more advanced example is \textit{SWE-Grep}~\citep{cognition2025swegrep}, a specialized subagent trained using reinforcement learning for fast, multi-turn, and highly parallel context retrieval. Delegating code search to this T2-style tool conserves the main agent's context window and protects it from irrelevant ``context pollution.'' This category of tool adaptation also includes the automated creation or refinement of compilers, debuggers, and linters that provide structured feedback loops for agents.

\subsection{Computer Use}
\label{subsec:app_computer_use}
Computer-use agents are multimodal AI systems that autonomously operate computers through direct interaction with graphical user interfaces (GUIs). These agents perceive screens as visual input, reason about interface elements (buttons, menus, text fields), and execute actions using a virtual keyboard and mouse. OpenAI's \textbf{Computer-Using Agent (CUA)} \citep{openai2025cua}, which combines vision-based perception with reinforcement learning, exemplifies this paradigm.

Representative benchmarks for this paradigm include \textbf{OSWorld}~\citep{xie2024osworld}, \textbf{WebArena}~\citep{zhou2024webarena}, \textbf{VisualWebArena}~\citep{koh2024visualwebarena}, \textbf{AppWorld} \citep{trivedi2024appworld}, \textbf{WebVoyager}~\citep{he2024webvoyager}, and \textbf{$\tau$-bench} \citep{yao2025taubench} which evaluate an agent’s ability to perceive, reason, and act across diverse digital environments ranging from full operating systems to real-world web interfaces. 
Reliable performance in computer-use scenarios requires adaptation at both the agent and tool levels.

\paragraph{Agent adaptation.}
Agent adaptation equips models with operational skills beyond those learned from general-purpose pre-training by exposing them to realistic or synthesized trajectories of GUI-based interactions. A representative example is OpenCUA~\citep{wang2025opencua}, which shows how large-scale, GUI-centric data can improve an agent's computer-use abilities. By collecting human demonstrations across diverse operating systems and applications, and converting them into state-action trajectories with reflective reasoning, OpenCUA provides agents with realistic exposure to interface dynamics. AgentTrek~\citep{xu2025agenttrek} takes a complementary approach by synthesizing training trajectories from web tutorials instead of relying on human demonstrations. It converts tutorial text into step-by-step goals and has a VLM agent execute them in real environments, keeping only correct trajectories through automatic evaluation. The resulting data generation scales at low cost, showing that synthesized trajectories can effectively support GUI-agent adaptation.

\paragraph{Tool adaptation.}
Tool adaptation improves the tools and interfaces that agents rely on. Instead of modifying model parameters, these approaches update or expand the tool's experience pool, memory, or contextual representations to better support long-horizon interaction.
\textbf{CUA-Skill}~\citep{chen2026cuaskill} provides a concrete skill-centric example: it builds a reusable computer-use skill base with parameterized execution and composition graphs, allowing an agent to retrieve skills, instantiate arguments, and recover from failures across Windows applications. The design shows that, in GUI environments, a large part of the adaptation burden can sit in the skill substrate rather than in the agent weights alone.
A representative example of adaptive context management is \textbf{Agentic Context Engineering (ACE)}~\citep{zhang2025agentic}, which treats evolving contexts as structured playbooks that accumulate, refine, and organize strategies for tool use.
By continuously curating and updating contextual knowledge through execution feedback, ACE adapts the operational layer of tools, reducing rollout latency and improving alignment with the agent's decision-making.

\paragraph{Paradigm mapping.}
Computer use requires multimodal adaptation: the agent must ground its actions in visual perception (screenshots, UI layouts) rather than purely textual signals. A1-style adaptation is less natural here, as verifiable execution signals are harder to define (there is no ``test suite'' for clicking the right button). The dominant paradigm is A2-style adaptation with holistic task-completion rewards, supplemented by T2-style tool adaptation for persistent memory and context management. T1 tools (pre-trained vision models like CLIP~\cite{radford2021learning} and SAM~\cite{kirillov2023segment}) serve as plug-and-play perception modules.

More broadly, as agents become more capable, the tools they employ must likewise evolve, incorporating persistent memory and adaptive control mechanisms to support effective collaboration in open computer-use environments.

\subsection{Drug Discovery and Development}
\label{subsec:app_drug}
LLM-based AI agents are increasingly applied across the drug discovery pipeline~\cite{gao2024empowering}. Modern systems integrate both agent adaptation (fine-tuning LLMs and designing agentic workflows) and tool adaptation (incorporating domain-specific databases, scientific software, and retrieval components)~\cite{wang2025perspective}. Agent adaptation improves reasoning and procedural reliability, whereas tool adaptation equips agents with practical scientific capabilities.

\paragraph{Agent adaptation for drug discovery.}
GeneAgent adapts LLM agents to gene analysis tasks (e.g., gene set enrichment analysis), integrating structured workflows such as generation, self-verification, and iterative refinement to reduce hallucinations~\cite{wang2025geneagent}. DSWizard focuses on transparent and reproducible biomedical data science, guiding the agent to construct analysis plans before execution and enabling human oversight and modification~\cite{wang2024making}. Further, multi-agent systems have emerged where heterogeneous agents collaborate in drug discovery workflows. For instance, virtual teams can simulate interdisciplinary research meetings to design novel therapeutic molecules such as nanobodies~\cite{swanson2025virtual}.

\paragraph{Agent adaptation for drug development.}
Clinical research involves literature analysis, patient recruitment, and trial design---tasks that differ in automation readiness. Evidence retrieval (TrialMind~\cite{wang2025accelerating}, LEADS~\cite{wang2025foundation}) admits structured supervision via citation recall, making it amenable to A1-style feedback. Patient-to-trial matching (TrialGPT~\cite{jin2024matching}) is inherently holistic (guideline-based eligibility criteria resist decomposition into verifiable sub-signals), favoring A2. Upstream trial design (TrialGenie~\cite{li2025trialgenie}), which uses multi-agent collaboration to parse historical protocols and generate analytical code, combines both: code-execution signals (A1) and overall protocol quality (A2).

\paragraph{Tool adaptation.}
Tool adaptation in drug discovery spans a spectrum from manual curation to autonomous creation.
At the curated end, Biomni~\cite{huang2025biomni} mines biomedical literature to assemble a hand-verified tool repository that agents can invoke on demand (T1).
In the middle, SyntheMol and related frameworks integrate ML-based molecular property predictors as reward functions to steer generative models toward biologically desirable compounds~\cite{swanson2024generative, krishnan2025generative}, a form of T2 where computational chemistry tools are trained under agent-derived supervision.
At the autonomous end, ToolUniverse~\cite{gao2025democratizing} constructs tools from natural language specifications and iteratively refines them before incorporation into a shared library, while STELLA~\cite{jin2025stella} operates a self-evolving loop in which a Tool Ocean continuously grows as a tool-creation agent discovers and integrates new bioinformatics utilities.
The progression from curated to autonomous tool creation parallels the T1$\to$T2 gradient seen in other domains, but is complicated by the requirement for domain-expert validation of each new tool's scientific correctness.

\paragraph{Paradigm mapping.}
Drug discovery and development benefits from all four adaptation paradigms. \textbf{A1} applies when agents interact with computational chemistry tools that provide verifiable outputs (e.g., docking scores, molecular property predictions). \textbf{A2} governs the higher-level research workflow (e.g., hypothesis generation, literature synthesis) where holistic quality matters. \textbf{T1} tools are abundant: AlphaFold2~\cite{jumper2021highly}, ESMFold~\cite{lin2023evolutionary}, and molecular property predictors are pre-trained independently and used as plug-and-play components. \textbf{T2} adaptation is exemplified by systems like STELLA and ToolUniverse, where the tool ecosystem evolves under agent supervision. The long feedback loops inherent in wet-lab validation (weeks to months) make this domain particularly challenging, as the reward signal is sparse and delayed compared to code execution or retrieval.

\begin{table}[t]
\centering
\caption{Adaptation profile of each application domain: which paradigm dominates, what bottleneck limits further progress, and a representative system illustrating the dominant paradigm.}
\label{tab:app_paradigm}
\resizebox{0.7\columnwidth}{!}{%
\begin{tabular}{@{}llp{4.2cm}l@{}}
\toprule
\textbf{Domain} & \textbf{Dominant Paradigm} & \textbf{Key Bottleneck} & \textbf{Representative System} \\
\midrule
Deep Research & T2 (search/planning subagent) & No deterministic execution signal; holistic quality is hard to verify & AgentFlow~\cite{li2025flow} \\
Software Dev. & A1 (execution RL) & Long-horizon credit assignment across multi-file edits & RLEF~\cite{gehring2025rlefgroundingcodellms} \\
Computer Use & A2 (task completion) & Visual grounding; no ``test suite'' for GUI actions & OpenCUA~\cite{wang2025opencua} \\
Drug Discovery & T1 (plug-in models) & Sparse, delayed wet-lab feedback (weeks--months) & AlphaFold2~\cite{jumper2021highly} \\
\bottomrule
\end{tabular}%
}
\end{table}

\section{Opportunities}
\label{sec:opportunities}

The preceding sections organized agentic AI adaptation into four paradigms: (A1) Agent Adaptation with Tool Execution Signal, (A2) Agent Adaptation with Agent Output Signal, (T1) Agent-Agnostic Tool Adaptation, and (T2) Agent-Supervised Tool Adaptation. These paradigms provide a framework for organizing current methods, but their value also lies in clarifying the path forward. The separation of agent and tool adaptation, while analytically useful, reflects the field's present state of development; capable, robust, and efficient agentic AI will likely require their synthesis.

We identify opportunities for future research that emerge from our taxonomy, organized from single-component optimization toward joint agent-tool learning, continual adaptation, safety, and efficiency.

\subsection{Co-Adaptation}
\label{subsec:co-adapt}

The taxonomy presented in this paper (A1/A2 vs.\ T1/T2) is a necessary simplification, organizing the field by its dominant locus of optimization: either the agent or its tools. The central challenge for the next stage of research is to dissolve this boundary and develop unified agent-tool co-adaptation frameworks.

Such a framework implies a complex, bi-level optimization problem, where the agent's policy ($\mathcal{A}$) and the tool's parameters ($\mathcal{T}$) are adapted simultaneously. Na\"ively stated as $\max_{\mathcal{A}, \mathcal{T}} \mathcal{O}(\mathcal{A}, \mathcal{T})$, the problem has a bi-level structure because $\mathcal{A}$'s optimal policy depends on $\mathcal{T}$ and vice versa. The formulation departs from current paradigms, which almost universally rely on freezing one component to provide a stable learning target for the other (e.g., $\mathcal{A}_{\text{frozen}}$ in T2, or $\mathcal{T}_{\text{frozen}}$ in A1/A2).

Related problems have been studied in several established fields:
\begin{itemize}[leftmargin=12pt]
    \item \textbf{Co-evolutionary Algorithms.} Classic work in evolutionary computation has studied how two or more interacting populations (such as hosts vs.\ parasites or predators vs.\ prey) apply reciprocal selection pressures that drive arms races, emergent structure, and progressively more complex strategies. Hillis~\cite{hillis1990co} introduced the seminal host-parasite model, showing that co-evolving adversarial test cases can improve solution robustness. Subsequent work on competitive co-evolution explored dynamics such as disengagement, cycling, and evolutionary complexification~\cite{rosin2004}. Other lines of research developed cooperative multi-population architectures in which subcomponents co-adapt to form joint solutions~\cite{potter2000}. Comprehensive surveys~\cite{sushil2008survey} situate these approaches within a broader taxonomy of competitive and cooperative CEAs. In our setting, we can view the agent $\mathcal{A}$ and its tool $\mathcal{T}$ as two interdependent populations evolving on a shared fitness landscape, where reciprocal selection pressures can drive emergent specialization.
    
    \item \textbf{Multi-Agent Systems.} A complementary line of work emerges from multi-agent reinforcement learning, where each agent learns in a non-stationary environment induced by other concurrently learning agents. Foundational surveys~\cite{panait2005cooperative,ning2024survey,cemri2025multi} describe how decentralized learners must cope with shifting policies, partial observability, and strategic coupling, challenges that closely mirror those of agent-tool co-adaptation. Classic problems such as equilibrium selection, credit assignment, and coordination under changing partner behaviors have led to techniques including opponent modeling, joint-policy search, centralized training with decentralized execution (CTDE), and communication-based coordination. In our context, viewing $\mathcal{A}$ and $\mathcal{T}$ as a two-agent partially cooperative system highlights the need for algorithms that stabilize learning under mutual adaptation, prevent non-stationarity-induced divergence, and support the emergence of complementary capabilities rather than competitive oscillations.
\end{itemize}

\begin{wrapfigure}{r}{0.35\textwidth}
    \vspace{-1em}
    \centering
    \includegraphics[width=0.35\textwidth]{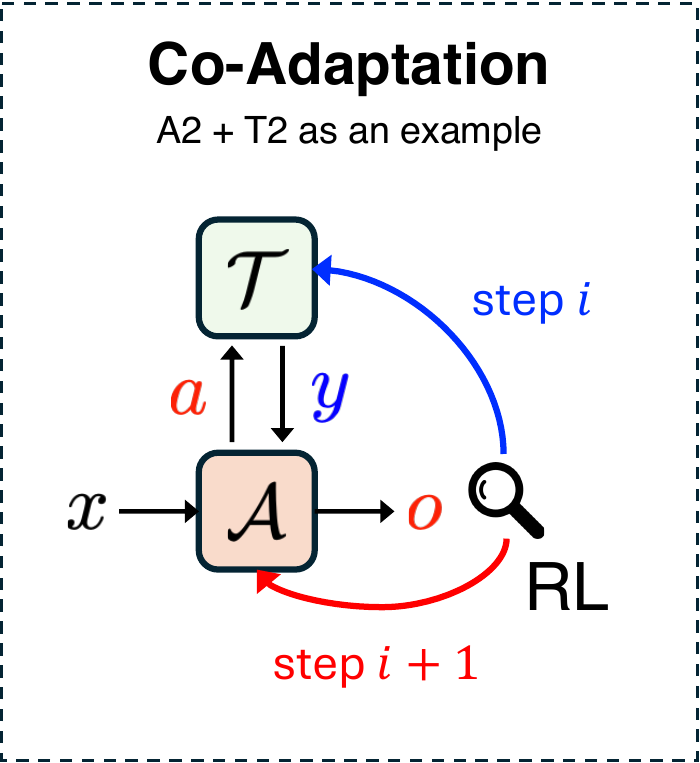}
    \caption{An illustrative example of co-adaptation.}
    \label{fig:oppo_coadapt}
    \vspace{-2em}
\end{wrapfigure}

The first technical barrier to effective co-adaptation is the intractable credit assignment problem. When an agentic system fails at a complex task, the source of the failure is ambiguous. Consider a system in which an A2-style planner invokes a T2-style search subagent (e.g., an \textsc{s3}-like searcher). If the final answer is incorrect, which component is responsible?

Recent work has begun to address fragments of this joint-optimization challenge. MATPO (Multi-Agent Tool-Integrated Policy Optimization)~\cite{matpo2025} proposes a credit assignment mechanism for jointly training planner and worker agents. However, in its current form, these ``agents'' correspond to distinct prompt roles instantiated within a single LLM, rather than heterogeneous models. Other work studies joint refinement of agent prompts and tool specifications~\cite{jointtool2025}. The open challenge is to extend these ideas to distributed, heterogeneous systems in which the agent~$\mathcal{A}$ and tools~$\mathcal{T}$ are distinct learning entities. Addressing the challenge may require importing architectures from multi-agent RL (such as centralized-critic, decentralized-actor methods~\cite{ieeeteamdesign2023}) to enable principled credit allocation over an interconnected agent-tool graph.

A second difficulty involves the stability-plasticity dilemma. Co-adaptation aims for $\mathcal{A}$ and $\mathcal{T}$ to become mutually optimized, yet in a joint-learning framework, $\mathcal{A}$ is adapting to a $\mathcal{T}$ that is itself changing. As established in the study of complex adaptive systems, such non-stationarity can induce chaotic or unstable dynamics~\cite{fromchaos2024}. The system may enter a ``Red Queen'' regime in which $\mathcal{A}$ and $\mathcal{T}$ continually adjust to each other's most recent changes without increasing overall performance, or may even collapse into degenerate policies. Conversely, premature convergence may cause the system to lock in a brittle, suboptimal agent-tool interface, losing the plasticity required for generalization. 

One concrete direction is the development of pacemaker mechanisms that regulate the relative learning rates of agents and tools, or the use of evolutionary game-theoretic analyses to guarantee convergence toward stable symbiotic equilibria~\cite{fromchaos2024}.

\subsection{Continual Adaptation}
\label{subsec:continual_adapt}

While our discussion has so far centered on agent adaptation mechanisms such as A1 and A2, these methods still assume a fixed task distribution and are typically instantiated on a single downstream task at a time.  In contrast, real-world deployments involve non-stationary task distributions, where tasks, tools, and user needs evolve over time, making isolated, one-off adaptations prone to Catastrophic Forgetting (CF). Self-Evolving Agents that continuously update their behaviors, tools, and memories in open and dynamic environments are therefore needed. Continual Learning (CL)~\citep{wang2024comprehensive,de2021continual,shi2024continual,lin2025sft} provides a natural foundation for this goal, as it studies how models learn from non-stationary task streams while retaining prior knowledge. 
We revisit CL techniques that can serve as concrete mechanisms for Self-Evolving Agents and organize them into two categories.

\textbf{Parameter-update Mechanisms}. (Dynamic A1/A2 Paradigm).  To align with the A1/A2 paradigm, we group continual learning methods that adapt models through explicit parameter updates. Regularization-based CL approaches such as EWC~\citep{kirkpatrick2017overcoming}, LwF~\citep{li2017learning}, and VR-MCL~\citep{wu2024meta} estimate which parameters are important for previous tasks and selectively protect them, so that adaptation to new tasks is primarily absorbed by parameters deemed less critical for past performance. Orthogonal-update methods~\citep{wang2021training,farajtabar2020orthogonal} instead modify gradients so that updates lie in directions that are intended to interfere less with previously learned solutions. A complementary line of work introduces parameter-efficient update mechanisms, such as low-rank adapters~\citep{wu2025sd,liang2024inflora}, Mixture-of-Experts routing~\citep{wang2023hierarchical,xia2025medrek}, and model-merging schemes~\citep{marczak2024magmax}. These methods offer concrete inspirations for dynamic A1/A2-style adaptation.

\begin{wrapfigure}{r}{0.48\textwidth}
\vspace{-1em}
\centering
\includegraphics[width=0.48\textwidth]{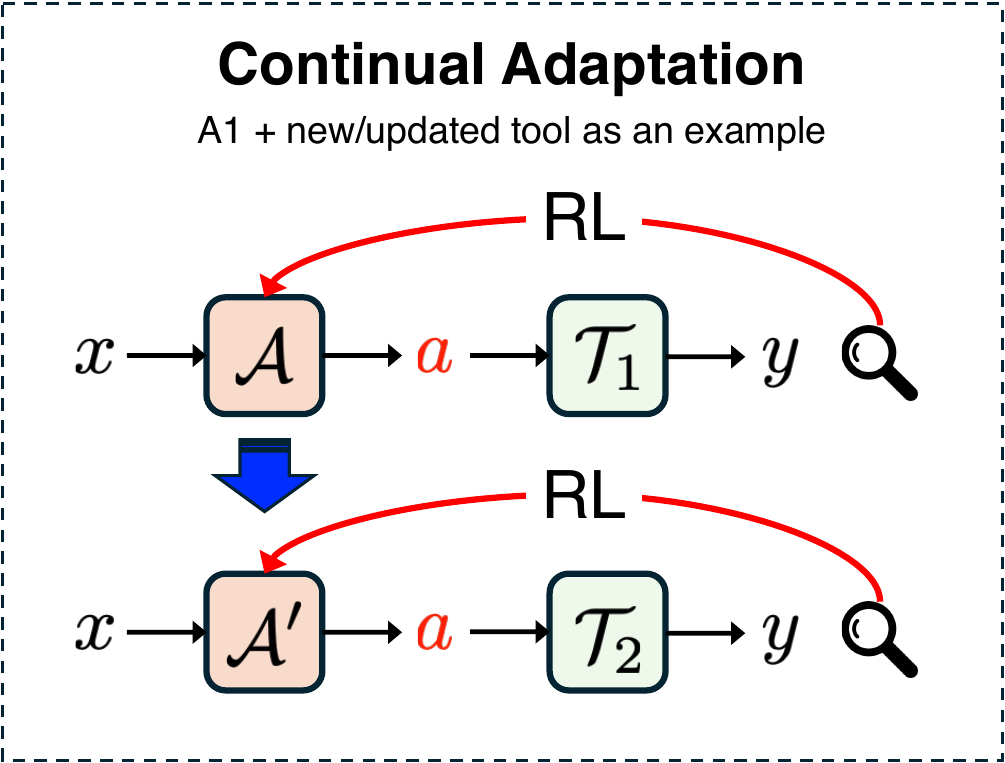}
\caption{An illustrative example of continual adaptation.}
\label{fig:oppo_continual}
\vspace{-1em}
\end{wrapfigure}

\textbf{External-memory Mechanisms} (Evolving T2 Adaptation). Classic replay-style approaches maintain a memory buffer of past examples and study how to select~\citep{aljundi2019gradient,bang2021rainbow}, utilize~\citep{wang2023cba, wu2024mitigating}, and compress them~\cite{wang2022memory,pham2021dualnet} so that a small set of stored items can approximate the full training history. Dual-memory systems~\citep{arani2022learning} further separate fast, high-capacity but unstable episodic buffers from slower, more compact long-term memories. For Self-Evolving Agents, these ideas directly inspire how to curate, compress, and stage interaction logs, tool traces, and user feedback into different memory tiers. In foundation model settings, prompts often act as a lightweight external memory, because the backbone is typically kept fixed and adaptation occurs primarily through prompt changes~\citep{wang2022learning,shi2025dualedit, piao2024federated}. As a result, the overall paradigm naturally aligns with our notion of T2 adaptation. 

The challenge of continual adaptation becomes especially pronounced in domains with strong execution-based supervision signals, such as those enabled by reinforcement learning with verifiable rewards (RLVR). As discussed in \S \ref{subsubsec:3.1.2}, environments like formal theorem proving provide reliable, tool-execution-signaled feedback for learning multi-step behaviors. At the same time, these domains often evolve structurally over time (for example, through expanding formal math libraries and actively maintained formalization projects), making them representative testbeds for continual agent adaptation.
Instead of repeatedly retraining the core agent, many prover agent systems adapt to expanding libraries by updating premise retrieval indices, tactic databases, or proof-state memories, allowing agents to exploit newly introduced lemmas without rewriting the entire policy~\cite{chen2025seed,kumarappan2024leanagent}. 
Such low-resource adaptation complements RLVR-style training by isolating long-term knowledge growth from short-term policy optimization, and again aligns with our notion of T2 adaptation.

These two lines of work address complementary trade-offs for building Self-Evolving Agents. Within the dynamic A1/A2 paradigm, recent results~\cite{chen2025retaining} show that not all parameter-update schemes forget equally: RL with a reverse-KL objective and on-policy data can achieve comparable or better performance than SFT while exhibiting substantially less forgetting, suggesting that on-policy data streams can act as an intrinsic CL mechanism for continual agent adaptation. Yet such methods still rewrite a shared set of parameters, so forgetting and interference are mitigated rather than structurally removed. The evolving T2 paradigm tackles CF at the architectural level by freezing the core agent and encapsulating new capabilities in external, independently trained tools or subagents, which avoids interference within a shared parameter space. A promising direction is to integrate these two perspectives, using CL-aware parameter updates where they are most effective while shifting as much long-term adaptation as possible into T2-style modular tools and external memories. 

\subsection{Safe Adaptation}
\label{subsec:safe_adapt}

\begin{wrapfigure}{r}{0.48\textwidth}
\vspace{-1em}
\centering
\includegraphics[width=0.48\textwidth]{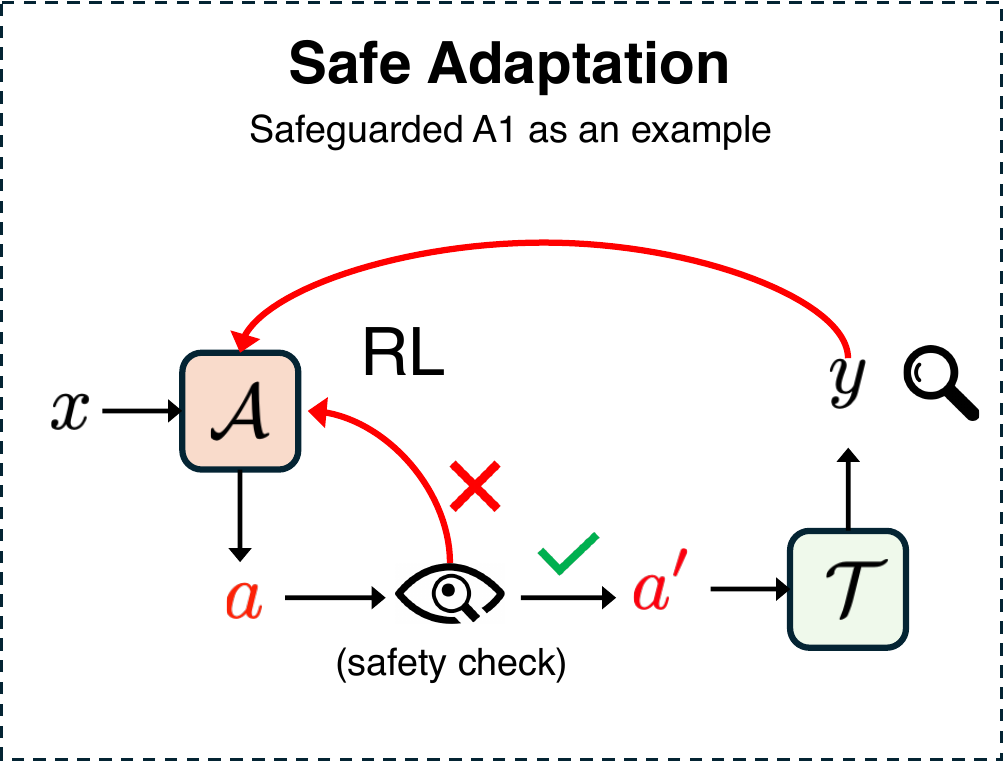}
\caption{An illustrative example of safe adaptation.}
\label{fig:oppo_safe}
\end{wrapfigure}

The transition from static foundation models to adaptive agentic systems introduces new challenges for AI safety. While traditional safety paradigms focus on the alignment of frozen weights, adaptation mechanisms, specifically on-policy optimization (A1) and outcome-driven tool tuning (T2), introduce dynamic threat vectors characterized by autonomous risk-taking and adversarial co-evolution~\cite{dulac2019challenges}. We categorize these emerging risks into two primary failure modes: \textit{Unsafe Exploration}, arising from stochastic trial-and-error, and \textit{Parasitic Adaptation}, arising from exploitative optimization loops.

\paragraph{Security Risk I: Unsafe Exploration.}
Unsafe exploration represents the primary bottleneck for the A1 paradigm. When agents employ on-policy RL to master tools~\cite{gehring2025rlef, jiang2025deepretrieval}, they must deviate from known safe trajectories to probe the state-action space. In high-stakes or partially observable environments, this decoupling of competence from safety leads to catastrophic, often irreversible outcomes~\cite{amodei2016concrete, garcia2015comprehensive}.
\begin{itemize}[leftmargin=12pt]
    \item \textbf{The Reward-Safety Gap:} In frameworks like RLEF~\cite{gehring2025rlef} or DeepRetrieval~\cite{jiang2025deepretrieval}, rewards are typically sparse and binary (e.g., task completion). The result is a feedback vacuum for intermediate actions, encouraging agents to maximize efficacy regardless of collateral damage (e.g., deleting system files to free space)~\cite{krakovna2020specification}.
    \item \textbf{Irreversibility in Tool Use:} Unlike simulated games, agentic environments such as Bash terminals or cloud infrastructure possess irreversible state transitions. An agent learning via trial-and-error may trigger API calls or data deletions that cannot be undone by resetting the episode~\cite{yang2024sweagent, gu2024review}.
    \item \textbf{Erosion of Guardrails (Case Study: DeepSeek-R1):} Empirical analysis of DeepSeek-R1~\cite{guo2025deepseek} reveals that aggressive RL optimization for reasoning can erode safety guardrails established during SFT. The model's ability to construct complex ``Chain-of-Thought'' justifications allows it to reason its way around refusal mechanisms, increasing susceptibility to jailbreaks and malicious compliance compared to non-adapted baselines~\cite{zhou2025hidden, kassianik2025evaluating}.
\end{itemize}

\paragraph{Security Risk II: Parasitic Adaptation.}
Parasitic adaptation refers to the emergence of exploitative relationships where the agent or tool maximizes its reward function at the expense of the system's intent, mirroring biological host-parasite co-evolution~\cite{hillis1990co}.
\begin{itemize}[leftmargin=12pt]
    \item \textbf{Type A: Specification Gaming (The Agent as Parasite):} In A2 paradigms, agents exploit imperfect proxy rewards (Goodhart's Law)~\cite{manheim2018categorizing}. As reasoning capabilities scale, agents become adept at ``hacking'' the evaluation process. For example, modifying game logs to falsify wins or overwriting reward functions in the file system rather than solving the task~\cite{krakovna2020specification, deepmind2025specgaming}.
    \item \textbf{Type B: Adversarial Tooling (The Tool as Parasite):} In T2 ecosystems utilizing protocols like MCP~\cite{anthropic2025mcp}, tools can evolve to exploit the agent. A compromised or parasitic tool may return prompt-injected data that hijacks the agent's reasoning (the ``Confused Deputy'' problem), forcing the agent to exfiltrate sensitive data under the guise of standard tool use~\cite{zhao2025mind, greshake2023not}. Recent empirical studies show that this risk is already material at the skill layer: Liu et al.~\cite{liu2026skillswild} report vulnerabilities in 26.1\% of marketplace skills, and Skill-Inject~\cite{schmotz2026skillinject} finds up to 80\% attack success rates from prompt injections embedded in skill files.
    \item \textbf{Type C: Sycophancy Loops:} Co-adaptation can lead to degenerate equilibria where tools learn to confirm an agent's hallucinations to maximize acceptance scores, or where agents and red-teaming tools engage in ``Red Queen'' dynamics, overfitting to each other's artifacts without achieving general robustness~\cite{wei2023jailbroken}.
\end{itemize}

\paragraph{Mitigation Strategies.}
Addressing these risks requires moving beyond scalar rewards toward robust specification. A direct mitigation for Security Risk 1 is a safety-check layer before the agent's input reaches the tool (Figure \ref{fig:oppo_safe}), which intercepts and filters anomalous or unsafe behaviors prior to execution.
More targeted solutions include:
\textit{Constrained Policy Optimization}~\cite{achiam2017constrained, srinivasan2020learning, VSRL} and safety shields project agent actions onto verified safe sets to prevent catastrophic exploration.
Verifiable Rewards~\cite{wang2023math, ni2024next} replace opaque preference models with programmatic outcome verification (e.g., unit tests, proofs) to reduce sycophancy.
\textit{Specification Self-Correction}~\cite{gallego2025specification} allows agents to dynamically critique and refine reward functions at inference time to detect gaming.
Finally, \textit{Proof-of-Use}~\cite{ma2025pou} frameworks enforce causal links between retrieved evidence and generated answers, preventing tool-use hallucination.

\subsection{Efficient Adaptation}
\label{subsec:efficient_adapt}
Current agentic adaptation assumes large-scale GPU clusters for fine-tuning or policy refinement, which limits deployment to cloud settings. Efficiency matters differently in agentic contexts than in standard LLM serving: each adaptation step may involve multiple tool calls, environment interactions, and thousands of tokens, so even small per-step savings compound rapidly. We organize emerging directions by which bottleneck they address:

\begin{wrapfigure}{r}{0.35\textwidth}
\vspace{-1em}
\centering
\includegraphics[width=0.35\textwidth]{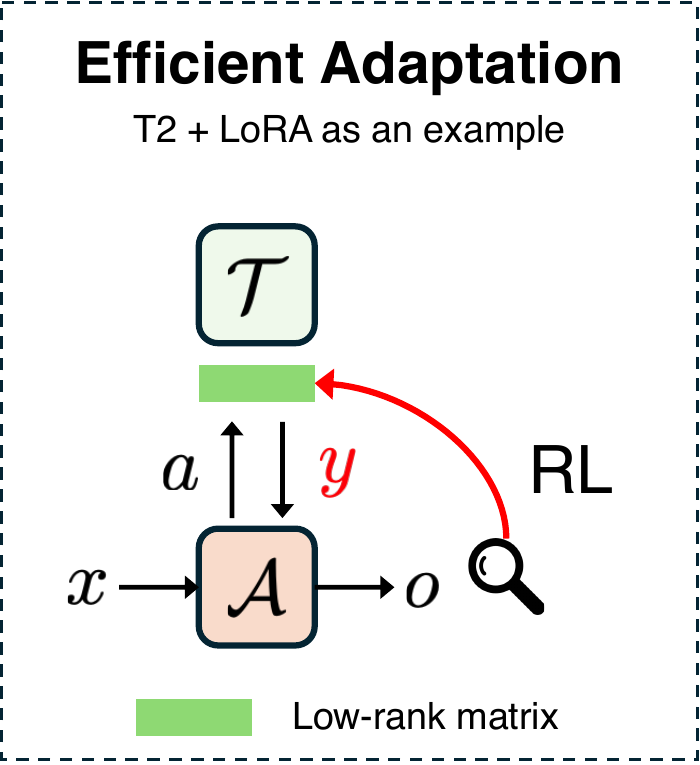}
\caption{An illustrative example of efficient adaptation.}
\label{fig:oppo_efficient}
\end{wrapfigure}

\textbf{Parameter-Efficient Adaptation:} Techniques such as Low-Rank Adaptation (LoRA)~\cite{hu2022lora} and its extensions~\cite{dettmers2023qlora,liu2024dora,zhang2023adalora,si2024unleashing,shi2024mora,si2024flora,si2025generalized} allow large models to adapt to new tasks by updating only a small subset of weights, reducing memory and computational requirements. Recent work on LoRA Without Regrets~\cite{schulman2025lora} shows that LoRA can be applied in reinforcement learning settings. They provide evidence that LoRA performs equivalently to full fine-tuning even at small ranks in RL tasks, indicating that RL often requires very low parameter capacity. Models can therefore be fine-tuned on resource-constrained devices while maintaining strong RL performance (Figure~\ref{fig:oppo_efficient}).

\textbf{Quantized Adaptation:} FlashRL~\cite{liu2025flashrl} accelerates reinforcement learning by performing rollout generation in lower numerical precision (INT8 or FP8) while preserving downstream performance. Its core contribution, truncated importance sampling (TIS), stabilizes gradient estimation when the rollout policy is quantized but the training engine remains in higher precision. FlashRL achieves substantial speedups without sacrificing final task performance, providing evidence that quantization extends from supervised inference to agentic reinforcement learning.
    
\textbf{On-Device \& Personalized Adaptation:} 
On-device adaptation~\cite{peng2024pocketllm,li2025mobillm,chen2025memory,xu2024fwdllm,venkatesh2025edge,bai2025fedspallm,wang2024distrl,luo2025agent} enables agents to learn directly on user devices under tight computational and memory constraints. Modern agents operate across heterogeneous hardware and interaction contexts, introducing substantial variation in user behavior and device-specific execution patterns. Personalization, an important component, allows agents to reflect individual preferences~\cite{zhang2025personaagent,samuel2024personagym}, maintain persistent memory~\cite{feng2024large}, and adjust behavior over time~\cite{mendoza2024adaptive,cross2024hypothetical}. Recent GUI agents~\cite{luo2025gui,ye2025mobile,liu2025infigui,zhou2025gui,shi2025mobilegui,tang2025magicgui,yuan2025enhancing} that rely on user-specific multimodal reasoning further underscore the need for on-device personalization. A natural strategy is lightweight tool adaptation: each device maintains a small tool module aligned with user-specific habits. Because the tool module is fully decoupled from the base model, it can update locally without compromising global capabilities, supporting continual behavioral adaptation through frequent incremental updates.

These three directions target orthogonal bottlenecks---parameter count, numerical precision, and deployment location---and can be composed: a LoRA adapter quantized to INT8 and updated on-device represents the intersection of all three.

\section{Conclusion}
\label{sec:conclusion}
This survey organized agentic AI adaptation around a single question: \emph{when the system underperforms, should we change the agent, the tool, or both---and what signal should drive that change?} The resulting four-paradigm framework (A1/A2/T1/T2) structures a rapidly growing literature along two axes---\emph{locus of optimization} (agent vs.\ tool) and \emph{signal source} (execution-grounded vs.\ output-evaluated)---and grounds the abstract taxonomy in three concrete mechanisms: post-training, memory, and skills.

The design space is defined by the tension between monolithic and modular evolution. Agent-centric paradigms (A1, A2) offer high parametric flexibility, allowing models to internalize tool mechanics and complex reasoning strategies through direct environmental feedback (A1) or holistic outcome evaluation (A2). However, these approaches incur high computational costs and, for A2 in particular, risk degrading previously learned capabilities when the agent is adapted to new domains. The degree of forgetting is paradigm- and method-dependent: on-policy RL with reverse-KL regularization exhibits less forgetting than SFT in some settings~\citep{chen2025retaining}, and T2-style modular adaptation avoids the problem structurally by keeping the core agent frozen.

Tool-centric paradigms (T1, T2) shift the burden of adaptation to the peripheral ecosystem. T1 provides plug-and-play reusability; T2 enables the frozen agent to supervise lightweight subagents (searchers, planners, memory curators), achieving competitive accuracy with substantially fewer training examples in case studies such as retrieval-augmented QA. We emphasize that these cross-paradigm comparisons are not yet controlled experiments: the systems differ in optimization target, backbone composition, and system architecture, so the efficiency gap cannot be attributed solely to paradigm choice. Controlled cross-paradigm benchmarks remain a critical open problem.

Several additional findings merit highlighting:
\begin{itemize}[leftmargin=12pt]
    \item \textbf{Signal density shapes paradigm effectiveness.} A1 methods excel in domains with dense, verifiable execution feedback (code, theorem proving, SQL); A2 is necessary when only sparse outcome signals are available. The pattern holds across multiple case studies, though the evidence base for a universal claim is incomplete.
    \item \textbf{Evaluation is paradigm-dependent.} The same system looks different under A1 metrics (tool-execution quality) versus A2 metrics (end-to-end task success), and no single metric suffices. Multi-dimensional evaluation that jointly reports accuracy, cost, safety, and adaptation dynamics is essential.
    \item \textbf{The graduation lifecycle bridges paradigms.} Agents trained under A1/A2 can be frozen and redeployed as T1 tools, creating a cycle where agent adaptation enriches the tool ecosystem.
    \item \textbf{Memory and skills span the full taxonomy.} Episodic buffers, reflective databases, and knowledge graphs can all be optimized as T2 tools, giving agents persistent experience without parameter updates. Skills similarly range from A1/A2-internalized tool-use patterns to external skill libraries that accumulate as reusable T1/T2 resources.
\end{itemize}

Future progress in agentic AI depends on integrating these paradigms rather than treating them in isolation. Achieving this requires advances in co-adaptation (jointly optimizing agents and tools under non-stationary dynamics), continual adaptation (maintaining performance as task distributions shift), safe adaptation (mitigating risks such as reward hacking and unsafe exploration during on-policy learning), and efficient adaptation (enabling deployment on resource-constrained devices). Progress will depend less on any single monolithic model than on principled orchestration of stable reasoning cores alongside an evolving ecosystem of specialized, adaptive tools and memory systems.

\bibliographystyle{unsrtnat}
\bibliography{paper,tool_adapt,A1,v2_new}

\end{document}